\documentclass{bmvc2k}

\usepackage{times}
\usepackage{epsfig}
\usepackage{graphicx}
\usepackage{amsmath}
\usepackage{amssymb}
\usepackage{color}
\usepackage{wrapfig,lipsum,booktabs}
\usepackage{xcolor}
\usepackage{multirow}
\usepackage{helvet}

\title{Noise-Aware Video Saliency Prediction} 
\runninghead{Prashnani et. al.}{Noise-Aware Video Saliency Prediction}
\addauthor{Ekta Prashnani\textsuperscript{1,}}{ekta@ece.ucsb.edu}{2}
\addauthor{Orazio Gallo}{ogallo@nvidia.com}{2}
\addauthor{Joohwan Kim}{sckim@nvidia.com}{2}
\addauthor{Josef Spjut}{jspjut@nvidia.com}{2}
\addauthor{Pradeep Sen}{psen@ece.ucsb.edu}{1}
\addauthor{Iuri Frosio}{ifrosio@nvidia.com}{2}
\addinstitution{
	University of California,\\
	Santa Barbara,\\
	California, USA
}
\addinstitution{
	NVIDIA Research,\\
	Santa Clara,\\
	California, USA
}

\def\eg{\emph{e.g}\bmvaOneDot,\xspace}

\def\ie{\emph{i.e}\bmvaOneDot,\xspace}

\DeclareMathOperator{\E}{\mathbb{E}}

\newcommand\nomarkerfootnote[1]{%
	\begingroup
	\renewcommand\thefootnote{}\footnote{#1}%
	\addtocounter{footnote}{-1}%
	\endgroup
}
\newcommand{\addtext} [1] {#1}
\definecolor{myblue}{RGB}{0, 25, 102}

\begin{document}

\maketitle
\begin{abstract}

We tackle the problem of predicting saliency maps for videos of dynamic scenes.
We note that the accuracy of the maps reconstructed from the gaze data of a fixed number of observers varies with the frame, as it depends on the content of the scene.
This issue is particularly pressing when a limited number of observers are available.
In such cases, directly minimizing the discrepancy between the predicted and measured saliency maps, as traditional deep-learning methods do, results in overfitting to the noisy data.
We propose a \emph{noise-aware training} (NAT) paradigm that quantifies and accounts for the uncertainty arising from frame-specific gaze data inaccuracy.
We show that NAT is especially advantageous when limited training data is available, with experiments across different models, loss functions, and datasets.
We also introduce a video game-based saliency dataset, with rich temporal semantics, and multiple gaze attractors per frame.
The dataset and source code are available at \href{https://github.com/NVlabs/NAT-saliency }{https://github.com/NVlabs/NAT-saliency}.

\end{abstract}
\nomarkerfootnote{\hspace{-0.2in}This work was done in part when E. Prashnani was interning at NVIDIA.}
\section{Introduction}
\label{sec:introduction}
Humans can perceive high-frequency details only within a small solid angle, and thus, analyze scenes by directing their gaze to the relevant parts~\cite{Gei98,Dez16}.
Predicting a distribution of gaze locations (\ie a \textit{saliency map}) for a visual stimulus has widespread applications such as image or video compression~\cite{Bae20} and foveated rendering~\cite{Kim19, Kap19}, among others.
This has inspired an active area of research -- visual saliency prediction. 
Early methods focused on low- or mid-level visual features~\cite{Itt98,Itt00,Koc87}, and recent methods leverage high-level priors through deep learning (DL) for saliency prediction \addtext{ and related tasks such as salient object detection}~\cite{Hua15,Kum17,Kum14,Pan17,Wan17,Jia20,Kru17,Liu18,Liu15,Gao20,Zha20,Zhang20,Liu20B}.

\begin{figure*}[h!]
	\centering
	\includegraphics[width=0.89\textwidth]{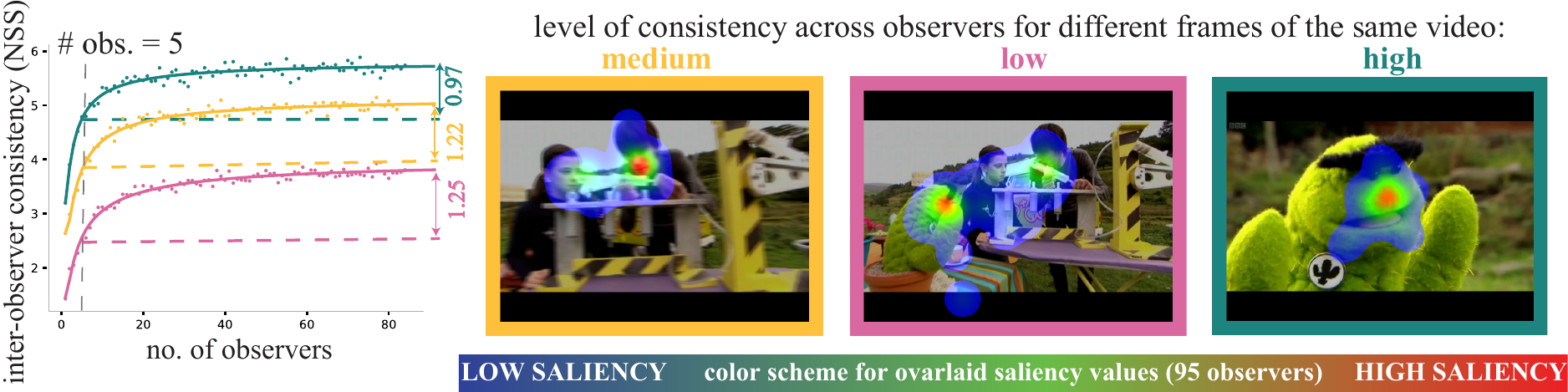}
	\caption{\addtext{\textbf{Motivation for noise-aware training (NAT).} Frames from a video (DIEM~\cite{Mit11} dataset) are shown with an overlay of the saliency maps reconstructed from the gaze data of $95$ observers.
	The level of gaze consistency across observers varies with frame content, leading to different asymptotic values and convergence rates of the per-frame inter-observer consistency (IOC) curves.
	Consequently, the accuracy of the saliency maps reconstructed from gaze data varies across frames -- especially when a limited number of observers (say, $5$ observers) are available.
	This impedes traditional training that directly minimizes the discrepancy between predicted and measured maps.
	We introduce NAT to address this.}}
	\label{fig:teaser}
	\vspace{-0.15in}
\end{figure*}

Given the improved accessibility of eye trackers~\cite{Val20}, datasets for saliency prediction are captured by recording gaze locations of observers viewing an image or a video.
These gaze locations are then used to estimate a per-frame/image saliency map.
Generally speaking, the quality of the reconstructed saliency maps increases with the number of gaze samples.
However, two factors make \addtext{it} particularly challenging \addtext{to reconstruct high-quality maps }for videos.
First, since a single observer contributes only a few (typically one~\cite{Mit11}) gaze locations per video frame, more  observers are needed to capture sufficient per-frame gaze data for videos as compared to images (\eg the CAT2000 dataset has on average ${\sim}333$ fixations per image from $24$ observers~\cite{Bor15}, while LEDOV has ${\sim}32$ fixations per video frame from $32$ observers~\cite{Jia21}).
Therefore, the cost associated with the creation of truly large-scale datasets with tens of thousands of videos can be prohibitively high.
Second, for videos of dynamic scenes, it is hard to guarantee high accuracy of the reconstructed saliency maps across all frames from the gaze data of a fixed number of observers.
This is because the gaze behavior consistency across observers depends on the scene content~\cite{Wil11}: scenes that elicit a high consistency would require fewer observers to reconstruct accurate saliency maps than those for which the inter-observer consistency (IOC) in gaze behavior is low.

Fig.~\ref{fig:teaser} shows $3$ frames from a DIEM video~\cite{Mit11} with 95-observer saliency map overlays and the per-frame IOC as a function of the number of observers used to reconstruct the saliency map~\cite{Kum15, Wil11, Jia21}.
A converged IOC curve indicates that additional observers do not add new information to the reconstructed saliency map and the captured number of observers (\eg $95$ in Fig.~\ref{fig:teaser}) are sufficient for accurate estimation of saliency~\cite{Jia21,Jud12}.
As is clear from these plots, when the number of available observers is small, the IOC curve differs from its asymptotic value by a varying amount for each frame.
This leads to varying per-frame accuracy of the saliency map reconstructed from few observers.
In such cases, traditional training methods, which minimize the discrepancy between the predicted and measured saliency, can lead to overfitting to the inaccurate saliency maps in the training dataset.

We address these issues by proposing a \emph{Noise-Aware Training} (NAT) paradigm:
we interpret the discrepancy $d$ between the measured and predicted saliency maps as a random variable, and train the saliency predictor through likelihood maximization.
We show that NAT avoids overfitting to incomplete or inaccurate saliency maps, weighs training frames based on their reliability, and yields consistent improvement over traditional training, for different datasets, deep neural networks (DNN), and training discrepancies, \emph{especially when few observers or frames are available for training}.
Therefore, NAT ushers in the possibility of designing larger-scale video-saliency datasets with fewer observers per video, since it learns high-quality models with less training data.

Although existing datasets have been vital to advance video saliency research~\cite{Jia21, Mit11}, a significant portion of these datasets consists of almost-static content, as observed recently by Tangemann et al.~\cite{Tan20}.
Using these datasets for training and evaluation therefore makes it difficult to assess how saliency prediction methods fare on aspects specific to \textit{videos}, such as predicting saliency on temporally-evolving content.
Consequently, even an image-based saliency predictor can provide good results for existing video saliency datasets~\cite{Tan20}.
As a step towards designing datasets with dynamic content, we introduce the Fortnite Gaze Estimation Dataset (ForGED), that contains clips from game-play videos of Fortnite, a third-person-shooter game amassing hundreds of million of players worldwide.
With ForGED, we contribute a novel dataset with unique characteristics such as: fast temporal dynamics, semantically-evolving content, multiple attractors of attention, and a new gaming context.
\vspace{-0.12in}

\section{Related work}
\label{sec:related_work}
\paragraph{Saliency prediction methods.}

In recent years, DL-based approaches have \addtext{remarkably advanced }video saliency prediction~\cite{Bor19}.
Existing works include (i) 3D CNN architectures that observe a short sub-sequence of frames~\cite{Jai21,Bel20,Min19}; (ii) architectures that parse one frame at a time but maintain information about past frames in feature maps (\eg simple  temporal accumulation or LSTMs~\cite{Lin19,Wan18,Gor2018,Wu20,Dro20}); or (iii) a combination of both~\cite{Baz16}. 
Some methods also decouple spatial and temporal saliency through specific features, such as ``object-ness'' and motion in a frame~\cite{Jia18,Jia21,Bak17}\addtext{, adopt vision transformers~\cite{Liu20B}, or predict a compact spatial representations such as a GMM~\cite{Red20}}.
Overall, existing works largely focus on improving model architectures, \addtext{output representations~\cite{Red20}}, and training procedures. 
In contrast, our NAT paradigm \addtext{is broadly applicable across all these categories} and it only modifies the loss function to account for the level of reliability of the measured saliency maps. 
We demonstrate the model-agnostic applicability of NAT through experiments on representative DNN architectures -- \addtext{we use ViNET~\cite{Jai21} (2021 state-of-the-art that uses 3D CNN)}, TASED-Net~\cite{Min19} (a 3D CNN-based model), and SalEMA~\cite{Lin19}. 

\vspace{-0.1in}
\paragraph{Metrics and measures of uncertainty for saliency.} 
Popular metrics for training and evaluating saliency models include density-based functions (Kullback-Leibler divergence, KLD, correlation coefficient, CC, similarity, SIM~\cite{Swa91}), and fixation-based functions (area under the ROC curve, AUC~\cite{Byl18,Byl15}, normalized scanpath saliency, NSS~\cite{Pet05, Ric13, Byl18}).
Fixation-based metrics evaluate  saliency at the captured gaze locations, \addtext{without} reconstructing the entire map. We observed that when few locations on a small training set are available, models that directly optimize either type of function show suboptimal performance.

The adoption of correction terms on a incomplete probability distributions has been explored in population satistics~\cite{Cha03,Hol98}.
Adapting these concepts to gaze data is possible at low spatial resolutions~\cite{Wil11}.
However, at full resolution, gaze data tends to be too sparse to collect sufficient statistics in each pixel.
IOC curves are also used to estimate the level of completeness of saliency maps~\cite{Jia21}, and the upper bounds on the performance of a saliency predictor ~\cite{Wil11,Jud12,Meu13,Kum15}.
Such approaches provide an insight on level of accuracy and uncertainty in saliency maps, but depend on the availability of sufficient observers to estimate the full curves. 
In contrast, NAT is designed specifically for limited-data setting.

\vspace{-0.1in}
\paragraph{Video saliency datasets.}
Some datasets capture video saliency for specific content (like sports~\cite{Rod08}, movies~\cite{Wig12}, faces~\cite{Liu17}), while others (like DHF1K~\cite{Wan18}, LEDOV~\cite{Jia18}, and DIEM~\cite{Mit11}) do for everyday scenes~\cite{Bor19}.
We perform our experimental analysis using two of the largest datasets, DIEM and LEDOV, which also provide high-quality gaze annotations, and, more importantly, access to per-observer gaze data -- a feature that is not available in the most popular DHF1K dataset, among other artifacts~\cite{Tan20}.
 
Videos with dynamic content are key to capturing and assessing \textit{video}-specific saliency. 
However, existing datasets contain mostly-static content, which can be explained by image-based models~\cite{Tan20}.
Existing datasets with videos of highly-dynamic content are either constrained in visual content variety and real-time gaze capture (\eg Atari-Head dataset~\cite{Zha19}), or capture gaze data from only a single subject (such as a game player~\cite{Bor12}, or a driver~\cite{All16}), limiting the accuracy of test-time evaluations.
We therefore turn to game-play videos of Fortnite, with its rich temporal dynamics, to further evaluate video-specific saliency.
ForGED features videos from Fortnite with gaze data from up to $21$ observers per video frame, enabling an effective benchmark for training and evaluating video-specific saliency.
\vspace{-0.1in}

\section{Noise-Aware Training (NAT)}
\label{sec:method_NAT}
\begin{wrapfigure}{r}{0.53\columnwidth} 
	\centering
	\includegraphics[width=0.53\columnwidth, trim=0cm 0cm 0cm 0cm, clip=true]{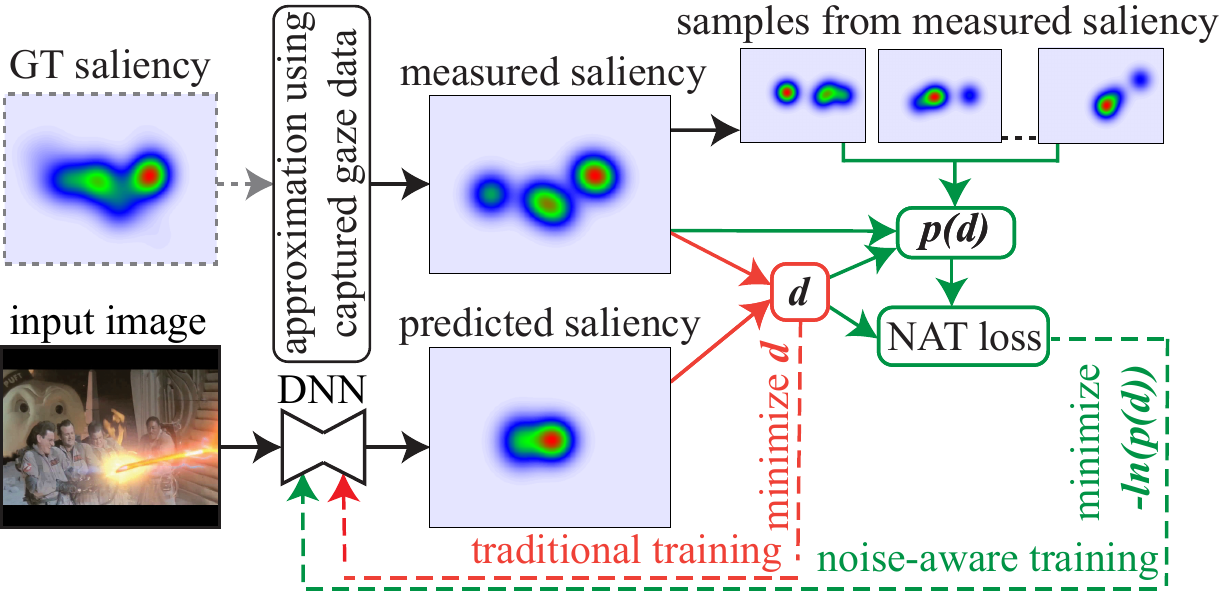}
	\vspace{-0.2cm}
	\caption{\textbf{Overview of NAT.}
		\addtext{For an input image, a saliency map is approximated from measured gaze data.
			This can result in a noisy/incomplete version of the GT saliency -- especially when limited gaze data is available.
			Instead of training a DNN by directly minimizing the discrepancy, $d$, between the measured and predicted saliency (traditional training), with NAT we first estimate a distribution for $d$, $p(d)$, that quantifies the uncertainty in $d$ due to the inaccuracies in the measured saliency maps.
			We then train the DNN to optimize the likelihood of $d$.}}
	\label{fig:NAT_overview}
\end{wrapfigure}
The accuracy of the saliency maps in videos varies with frame content, especially when limited gaze data is available.
The inaccuracy in the saliency maps can stem from errors in gaze measurements, such as inaccurate localization of Purkinje reflections or calibration issues in gaze tracker~\cite{Fei17} -- we term these \textit{measurement noise}. 
Using an insufficient number of observers to estimate the probabilities in different sub-regions of the saliency map is another source of noise, which we term \textit{incomplete sampling}.
While the measurement noise can be partially alleviated with techniques such as temporal filtering~\cite{Cha08}, the best way to overcome \textit{both} sources of noise is to capture sufficient data.
Since this can be impractical, we now discuss our proposed strategy to effectively train a DNN for saliency prediction, accounting for the noise level in a measured saliency map (Fig.~\ref{fig:NAT_overview}).

Let $x_i$ be the probability distribution of the \addtext{\textit{ground-truth}} saliency map for the $i^{th}$ frame, reconstructed from \textit{sufficient} gaze data (\eg when the IOC curve is close to its asymptotic value).
\addtext{The traditional approach to train a saliency predictor (abbreviated as TT : traditional training) optimizes}:
\begin{equation}
    J^{\text{ideal}} = \sum\nolimits_i d(\hat{x}_i,x_i),
    \label{eq:ideal_vanilla_cost_function}
\end{equation}
where $\hat{x}_i$ is the predicted saliency map, and $d(\cdot,\cdot)$ is a discrepancy measure such as KLD, CC, NSS, or a mix of these.
\addtext{Since reconstructing an accurate $x_i$ is challenging, the existing methods instead end up optimizing:}
\begin{equation}
    J^{\text{real}} = \sum\nolimits_i d(\hat{x}_i,\tilde{x}_i),
    \label{eq:vanilla_cost_function}
\end{equation}
where $\tilde{x}_i$ is an \addtext{\textit{approximation}} of the \addtext{unobservable} $x_i$.
\addtext{We adopt the standard practice to estimate $\tilde{x}_i$ from captured gaze data~\cite{Jia21,Ehi09,Wil11}: spatial locations are sampled from $x_i$ during gaze acquisition, followed by blurring with a Gaussian kernel and normalization to obtain the probability density function (pdf) $\tilde{x}_i$.
This can also be seen as a Gaussian Mixture Model with equal-variance components at measured gaze locations.
Let us denote this process of sampling spatial locations and reconstructing a pdf (``SR'') as: 
\begin{equation}
	\tilde{x}_i\ = \ SR(x_i\ ;\  N),
	\label{eq:xtilde}
\end{equation}
where $N$ is the number of spatial locations sampled from $x_i$ via gaze data capture.
For videos, $N$ is equivalently the number of observers.} 

\addtext{Given that $\tilde{x}_i$ can be prone to inaccuracies/noise, minimizing} $J^{\text{real}}$ during training can lead to noise overfitting and suboptimal convergence \addtext{(see Supplementary Sec.~8)}. 
Instead of directly minimizing $d(\hat{x}_i,\tilde{x}_i)$, \addtext{our approach} models the uncertainty in $d(x_i,\tilde{x}_i)$ due to the noise in $\tilde{x}_i$. \addtext{We first estimate a probability density function for $d(\hat{x}_i,\tilde{x}_i)$, denoted by $p[d(x_i,\tilde{x}_i)]$, and then} train the DNN for saliency prediction by \addtext{maximizing the likelihood of $d(x_i,\tilde{x}_i)$}.

We interpret $d(x_i,\tilde{x}_i)$ as Gaussian random variable with statistics $\E[d(x_i,\tilde{x}_i)]$, $\text{Var}[d(x_i,\tilde{x}_i)]$.
\addtext{We first consider an ideal case where $x_i$ is available and therefore we} can compute these statistics \addtext{by sampling and reconstructing} several realizations of $\tilde{x}_i$ from  $x_i$ (Eq.~\ref{eq:xtilde}; no gaze data acquisition needed), \addtext{and then computing sample mean $\E[d(x_i,\tilde{x}_i)]$  and variance $\text{Var}[d(x_i,\tilde{x}_i)]$.}
The value of these statistics depends on \addtext{the number of available gaze locations} $N$ used to reconstruct $\tilde{x}_i$ and on the complexity of $x_i$.
For example, when $x_i$ consists of a simple, unimodal distribution -- \eg when only one location in a frame catches the attention of all the observers -- a small $N$ is sufficient to bring $\tilde{x}_i$ close to $x_i$, which leads to low $\E[d(x_i,\tilde{x}_i)]$ and $\text{Var}[d(x_i,\tilde{x}_i)]$ values.
Alternatively, for a complex multimodal $x_i$, a larger $N$ is required for $\tilde{x}_i$ to converge to $x_i$ and consequently, $\E[d(x_i,\tilde{x}_i)]$ and $\text{Var}[d(x_i,\tilde{x}_i)]$ are large when $N$ is small (more discussion on this in Supplementary, Sec.~2).

Our NAT cost function is then defined as the following negative log likelihood:
\begin{equation}
    J_{\text{NAT}} = -\text{ln}\prod\nolimits_i p[d(\hat{x}_i,\tilde{x}_i)] = -\sum\nolimits_i \text{ln}\{p[d(\hat{x}_i,\tilde{x}_i)]\},
    \label{eq:nat_likelihood}
\end{equation}
that enables us to account for the presence of noise in the training data, for any choice of $d$.
\addtext{If $\E[d(x_i,\tilde{x}_i)]$ and $\text{Var}[d(x_i,\tilde{x}_i)]$ are known}, and assuming that $d(x_i,\tilde{x}_i)$ is a Gaussian random variable, we can simplify Eq.~\ref{eq:nat_likelihood} (see Supplementary\addtext{, Sec.~1}) to get:

\begin{equation}
    J_{\text{NAT}}^{\text{ideal}} = \sum\nolimits_i {\{d(\hat{x}_i, \tilde{x}_i) - \E[d(x_i,\tilde{x}_i)]\}^2} / {\text{Var}[d(x_i,\tilde{x}_i)]}.
    \label{eq:nat_cost_function_ideal}
\end{equation}

We note that $J_{\text{NAT}}^{\text{ideal}}$ penalizes $\hat{x}_i$ that are far from $\tilde{x}_i$, as in the traditional case.
However, it also ensures that $\hat{x}_i$ is not predicted \emph{too close} to the noisy $\tilde{x}_i$, which helps prevent noise overfitting (similar to discrepancy principles applied in image denoising~\cite{Ber10,Fro18}).
The penalization is inversely proportional to $\text{Var}[d(x_i,\tilde{x}_i)]$, \addtext{\ie it is strong for frames where $\tilde{x_i}$ is a good approximation of $x_i$.}
In contrast, $\E[d(x_i,\tilde{x}_i)]$ and $\text{Var}[d(x_i,\tilde{x}_i)]$ are large for multimodal, sparse \addtext{$\tilde{x}_i$ \textit{containing gaze data from only a few observers}, since in such cases, $\tilde{x_i}$ is not a good approximation of $x_i$.}
\addtext{This prevents the NAT formulation from overfitting to such uncertain $\tilde{x_i}$, by weakly penalizing the errors in $\hat{x}_i$ when compared to $\tilde{x}_i$.}

However, Eq.~\ref{eq:nat_cost_function_ideal} cannot be implemented in practice, as $x_i$ \addtext{(and consequently $\E[d(x_i,\tilde{x}_i)]$ and $\text{Var}[d(x_i,\tilde{x}_i)]$) is unknown.
We only have access to $\tilde{x}_i$, a noisy realization of $x_i$.
We therefore turn to approximating the statistics of $d(x_i,\tilde{x}_i)$ as:}
\begin{equation}
\E[d(x_i,\tilde{x}_i)] \approx \E[d(\tilde{x}_i,\tilde{\tilde{x}}_i)],
\;\text{Var}[d(x_i,\tilde{x}_i)] \approx \text{Var}[d(\tilde{x}_i,\tilde{\tilde{x}}_i)].
\label{eq:approx2}
\end{equation}

\addtext{Here, $\tilde{\tilde{x_i}}\ = \ SR(\tilde{x_i}\ ;\ N)$ is the pdf obtained by sampling N spatial locations  from $\tilde{x}_i$, followed by blurring ($N$ is also the number of gaze fixations sampled from $x_i$ by real observers).
The difference between how $\tilde{x_i}$ is reconstructed from $x_i$ and $\tilde{\tilde{x_i}}$ from $\tilde{x_i}$ is in the manner of obtaining the $N$ spatial locations: the $N$ spatial locations used to reconstruct $\tilde{x}$ are obtained from human gaze when viewing the $i^{th}$ frame; while for reconstructing $\tilde{\tilde{x_i}}$, $N$ spatial locations are sampled from the pdf $\tilde{x_i}$.
Multiple realizations of $\tilde{\tilde{x}}_i$ are then used to estimate $\E[d(\tilde{x}_i,\tilde{\tilde{x}}_i)]$ and $\text{Var}[d(\tilde{x}_i,\tilde{\tilde{x}}_i)]$.}
Intuitively, the approximation \addtext{in Eq.~\ref{eq:approx2}} holds because the level of consistency across multiple realizations of $\tilde{\tilde{x}}_i$ would be low when $\tilde{x}_i$ is complex (multimodal) \addtext{with small $N$} and indicates that the underlying \addtext{GT} saliency map $x_i$ must also be complex. 
Similarly, a high consistency across multiple realizations of $\tilde{\tilde{x}}_i$ points towards \addtext{a reliable} $\tilde{x}_i$. 
Therefore, the spatial noise introduced by sampling from $\tilde{x}_i$ serves as a proxy of the various noise introduced by the insufficient gaze-capturing process.
We observe empirically that these approximations hold with a mean absolute percentage error of $10-21\%$ on real cases (see Supplementary \addtext{Sec.~4}).

\addtext{Using Eq.~\ref{eq:approx2}, the NAT formulation from Eq.~\ref{eq:nat_cost_function_ideal} is modified to minimize}:
\begin{equation}
J_{\text{NAT}}^{\text{real}} = \sum\nolimits_i
{\{d(\hat{x}_i,\tilde{x}_i) - \E[d(\tilde{x}_i,\tilde{\tilde{x}}_i)]\}^2}/{\text{Var}[d(\tilde{x}_i, \tilde{\tilde{x}}_i)]},
\label{eq:vat_training_cost_function_real}
\end{equation}
where all the terms are now well-defined and a DNN can be trained using this cost function.
When implementing Eq.~\ref{eq:vat_training_cost_function_real}, for numerical stability, a small offset of $5e^{-5}$ is applied to the denominator, and $\E[d(\tilde{x}_i,\tilde{\tilde{x}}_i)]$ and $\text{Var}[d(\tilde{x}_i, \tilde{\tilde{x}}_i)]$ are computed using $10$ realization of $\tilde{\tilde{x}}_i$.

\addtext{Fig.~\ref{fig:datasets} shows the mean and standard deviation of $\text{KLD}(\tilde{x}_i||\tilde{\tilde{x}}_i)$ for some frames in ForGED, as estimated by Eq.~\ref{eq:approx2}.
Frames with high consistency across \addtext{several} observers are considered more reliable for training -- a feature that is exploited by NAT in Eq.\ref{eq:vat_training_cost_function_real}.}
\vspace{-0.1in}
\section{The ForGED dataset}
\label{sec:method_fortneyetd}

\begin{figure}[t]
	\centering
	\begingroup
	\setlength{\tabcolsep}{0pt}
	\renewcommand{\arraystretch}{0.1}
	\begin{tabular}{cccccc}
		\scriptsize{$\tilde{x}_i$ with 15 observers} & \scriptsize{$\tilde{x}_i$ with 5 observers} & \scriptsize{$\tilde{x}_i$ with 2 observers} \vspace{0.1cm} & \scriptsize{$\tilde{x}_i$ with 15 observers} & \scriptsize{$\tilde{x}_i$ with 5 observers} & \scriptsize{$\tilde{x}_i$ with 2 observers}\\ 
		\includegraphics[width=0.16\textwidth,trim={1.5cm 0.5cm 1.5cm 0.5cm},clip]{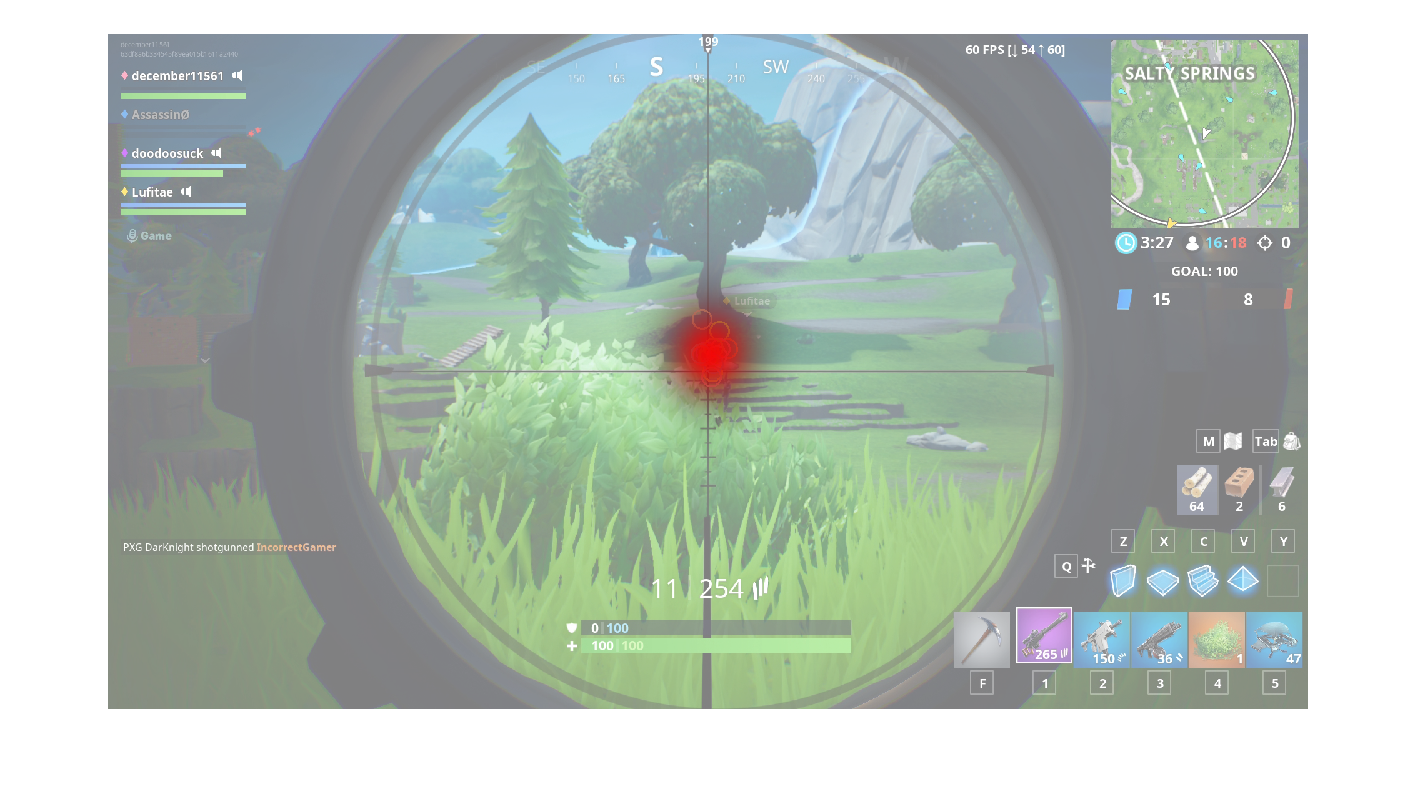} &  
		\includegraphics[width=0.16\textwidth,trim={1.5cm 0.5cm 1.5cm 0.5cm},clip]{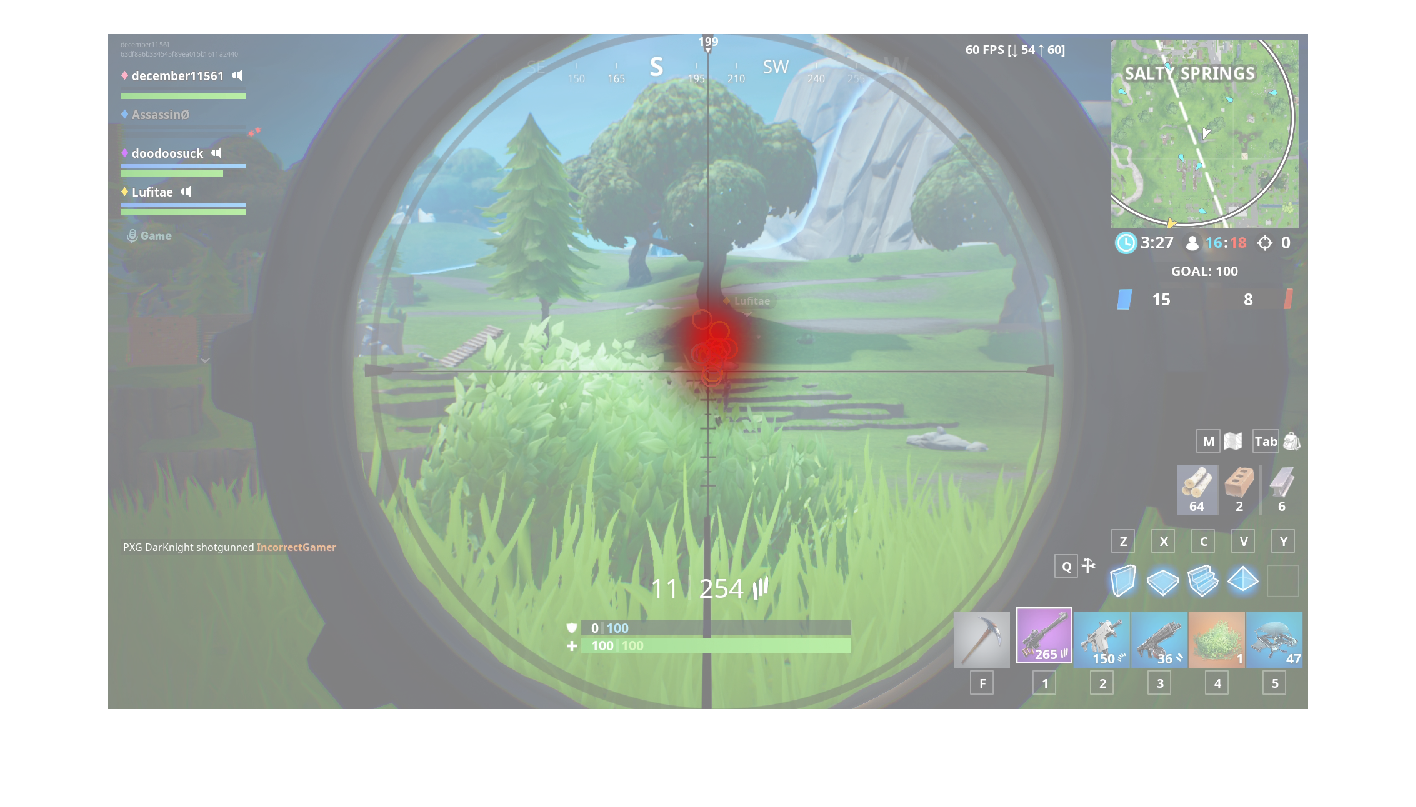} &
		\includegraphics[width=0.16\textwidth,trim={1.5cm 0.5cm 1.5cm 0.5cm},clip]{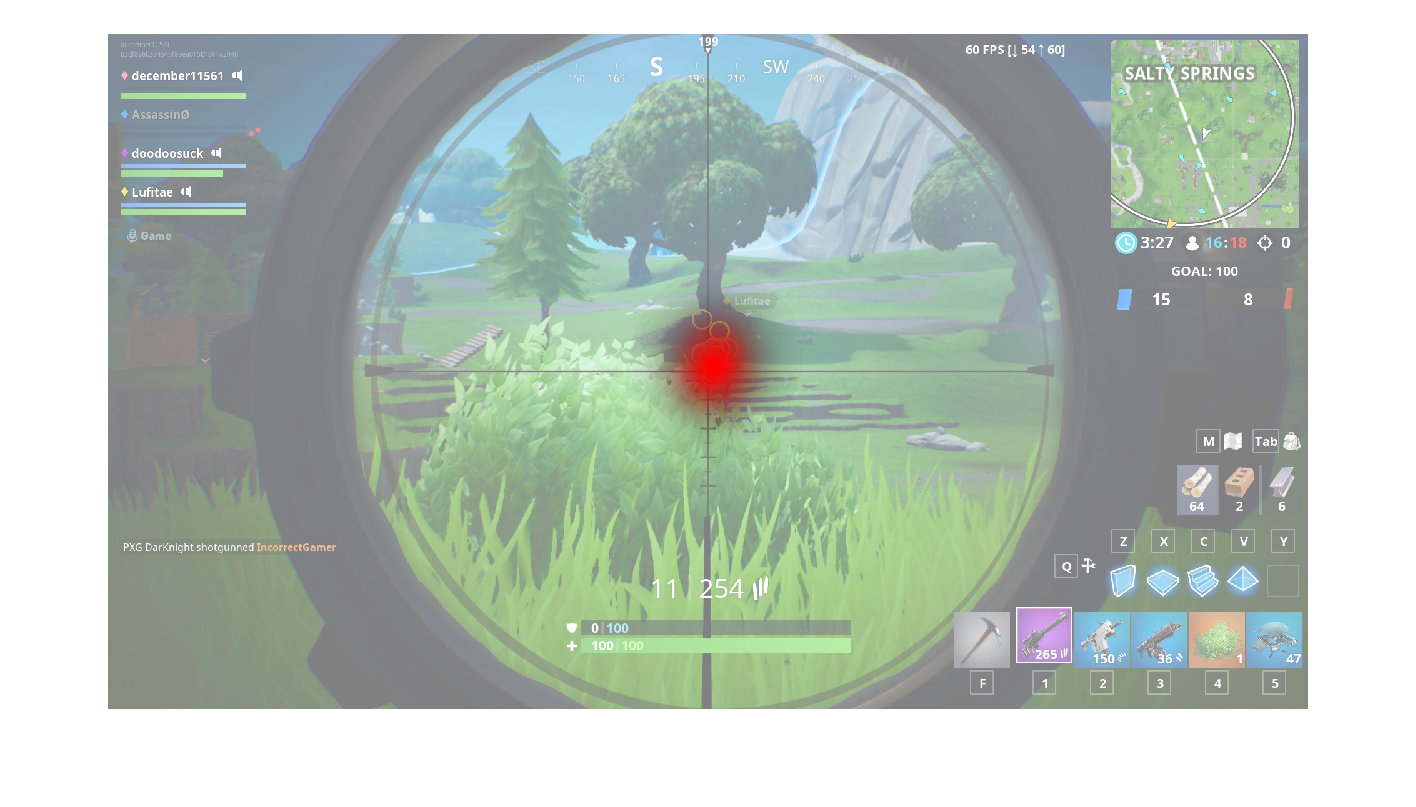} &
		\includegraphics[width=0.16\textwidth,trim={1.5cm 0.5cm 1.5cm 0.5cm},clip]{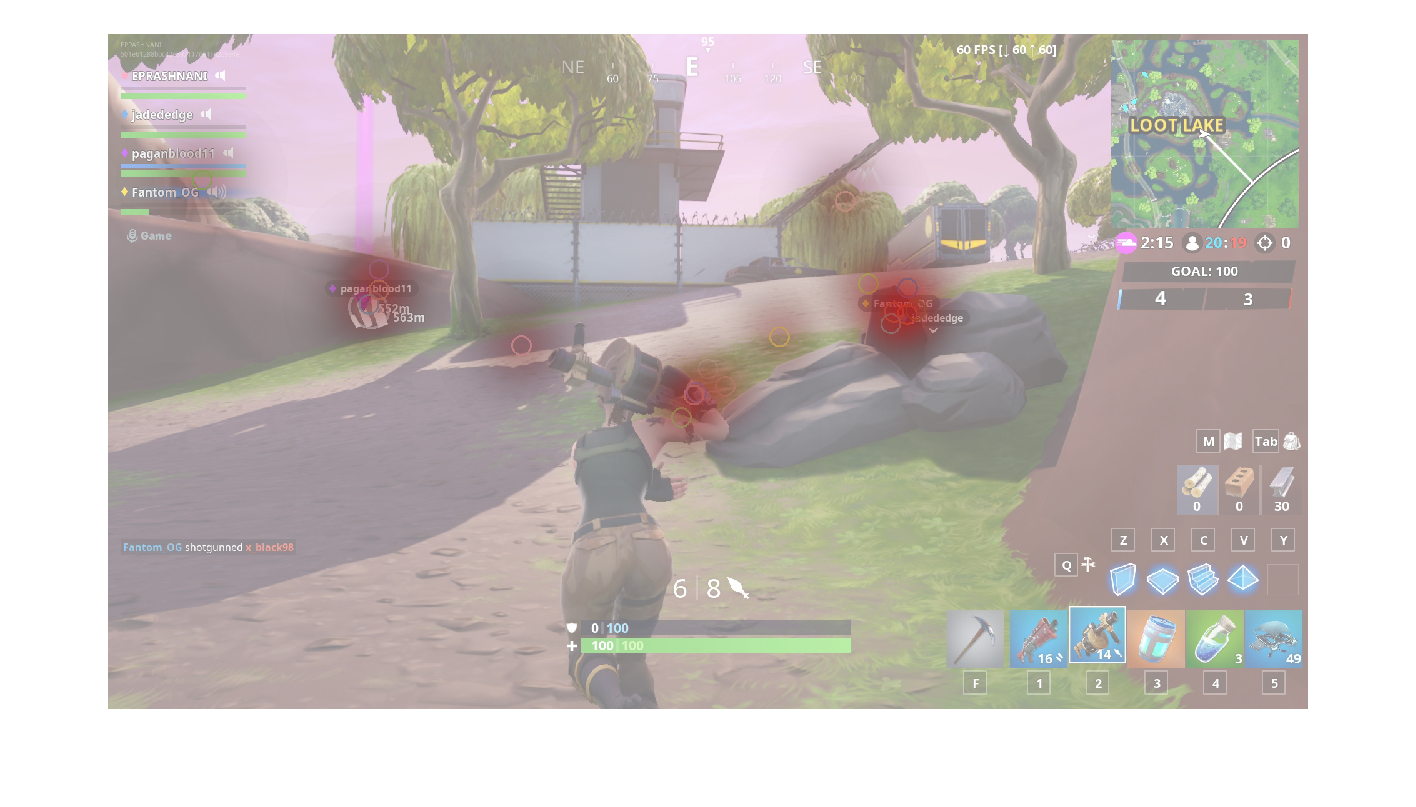} & 
		\includegraphics[width=0.16\textwidth,trim={1.5cm 0.5cm 1.5cm 0.5cm},clip]{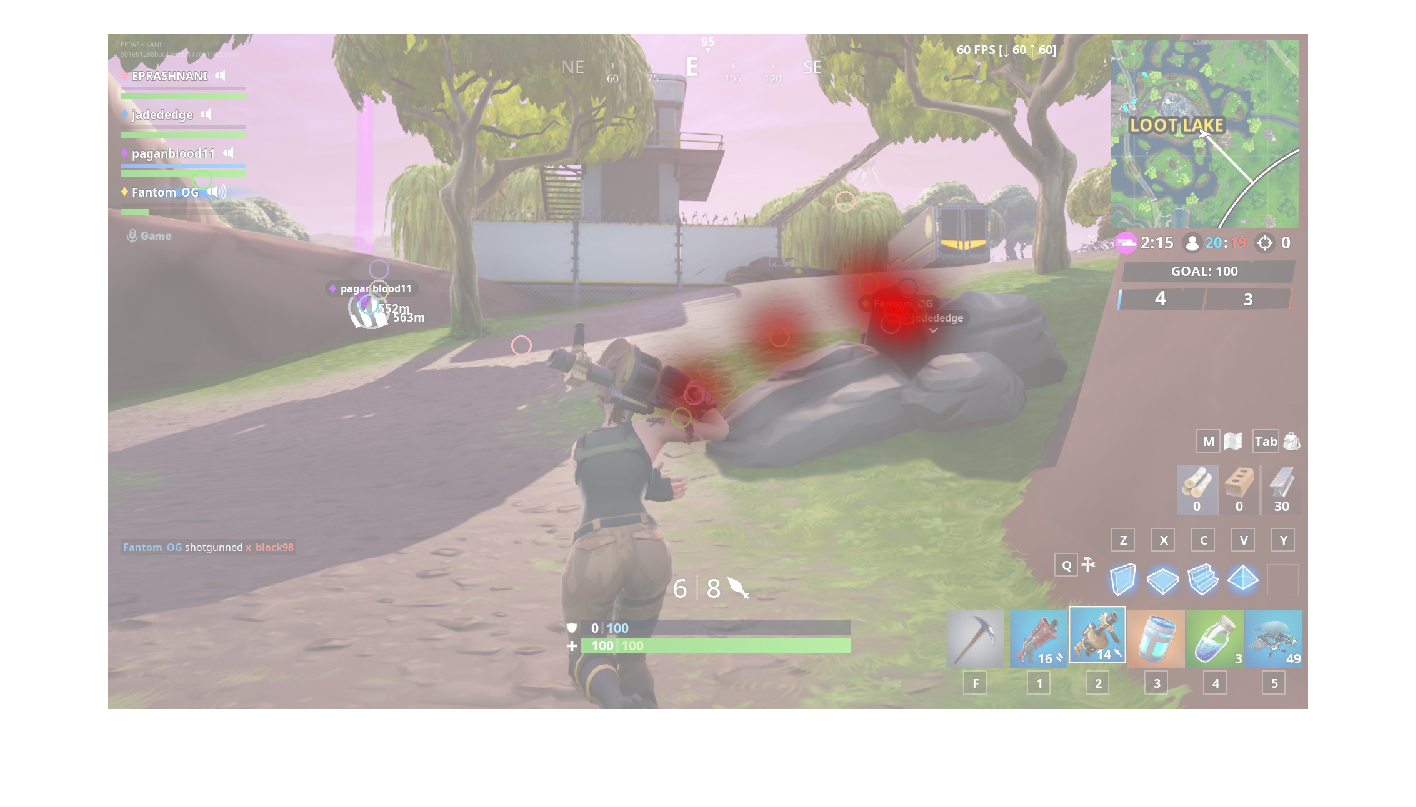} &
		\includegraphics[width=0.16\textwidth,trim={1.5cm 0.5cm 1.5cm 0.5cm},clip]{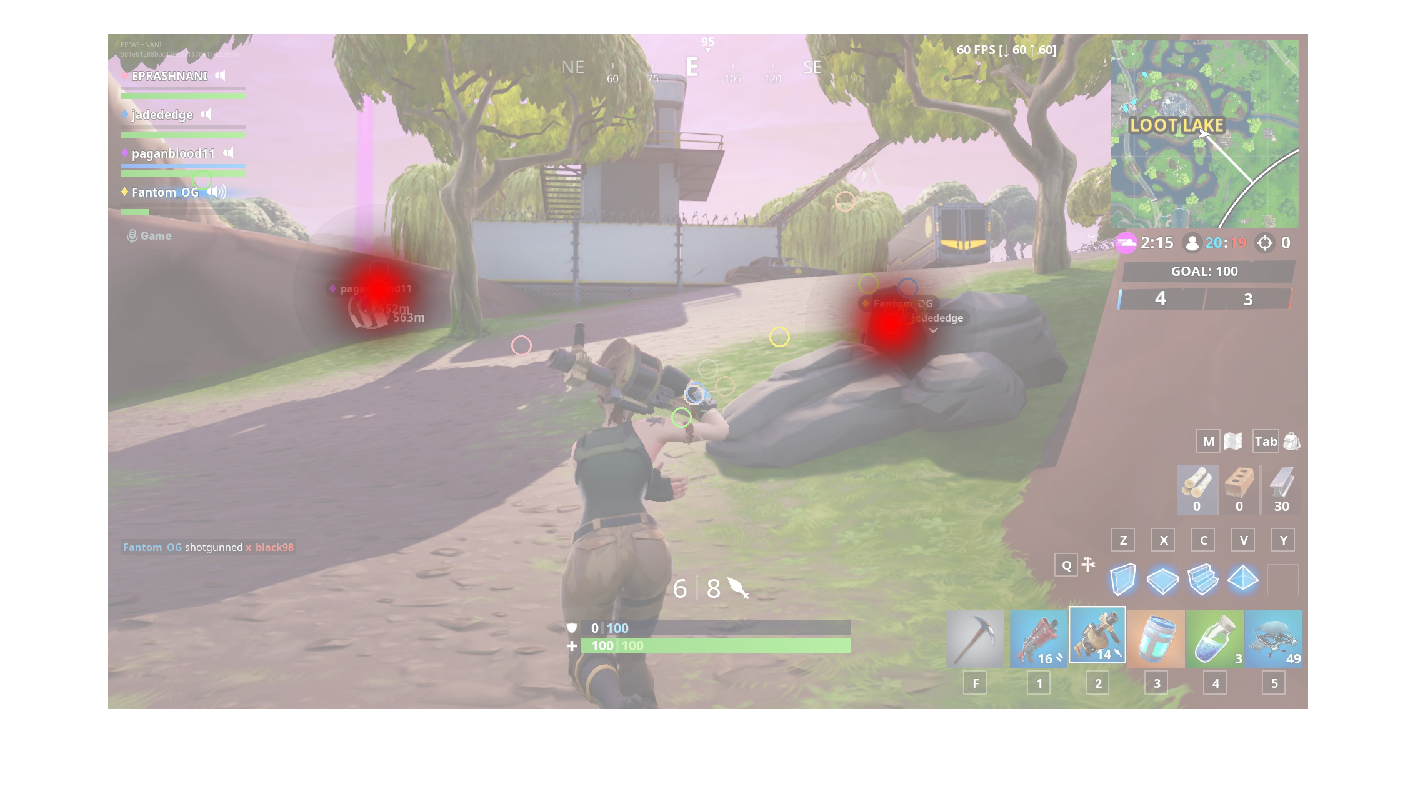} \vspace{-0.45cm}\\
		\textcolor{blue}{\small $0.197 \pm 0.077$} & \textcolor{blue}{\small $0.261 \pm 0.136$} & \textcolor{blue}{\small $0.478 \pm 0.418$} & \textcolor{blue}{\small $0.442 \pm 0.112$} & \textcolor{blue}{\small $0.585 \pm 0.277$} & \textcolor{blue}{\small $1.018 \pm 0.630$} \vspace{0.2cm} \\
		\includegraphics[width=0.16\textwidth,trim={1.5cm 0.5cm 1.5cm 0.5cm},clip]{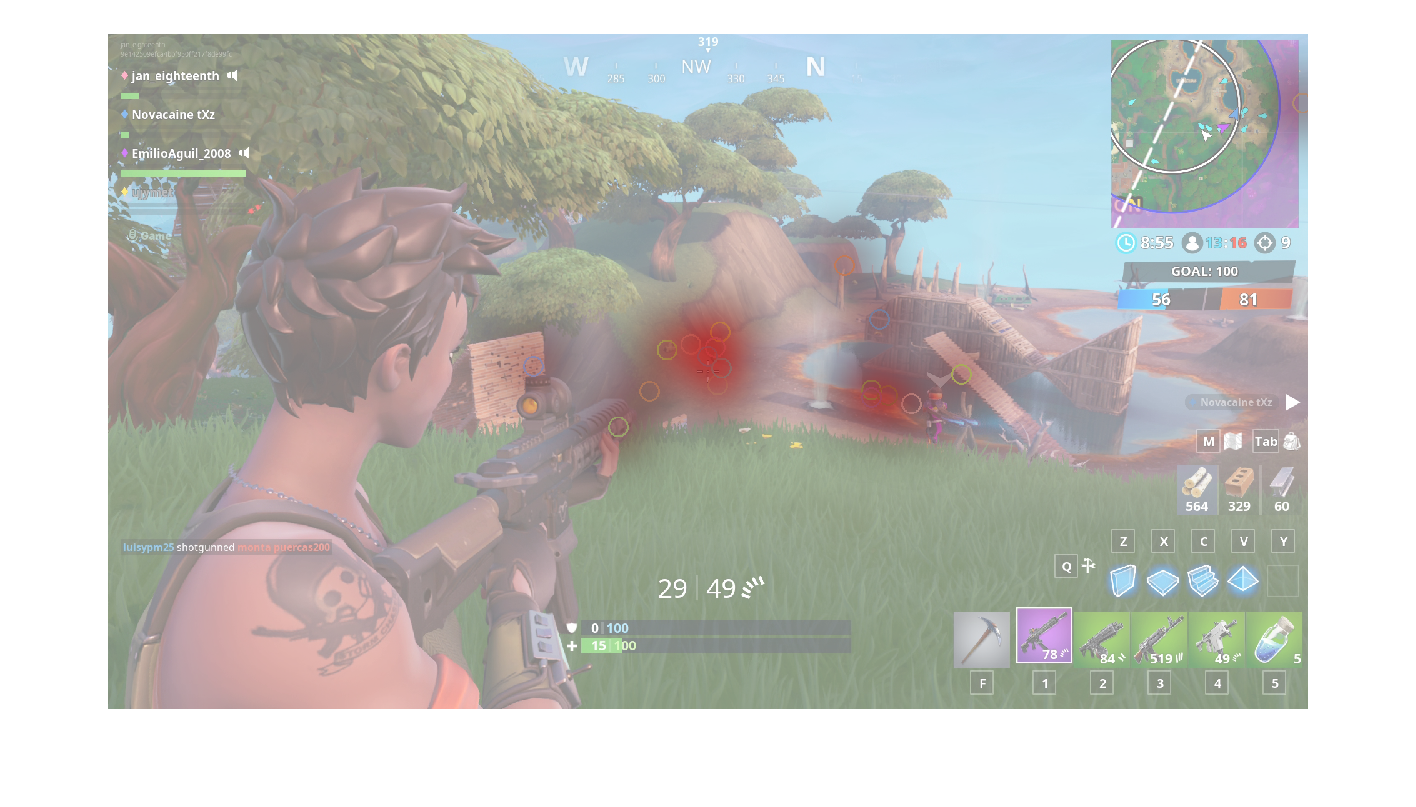} &     \includegraphics[width=0.16\textwidth,trim={1.5cm 0.5cm 1.5cm 0.5cm},clip]{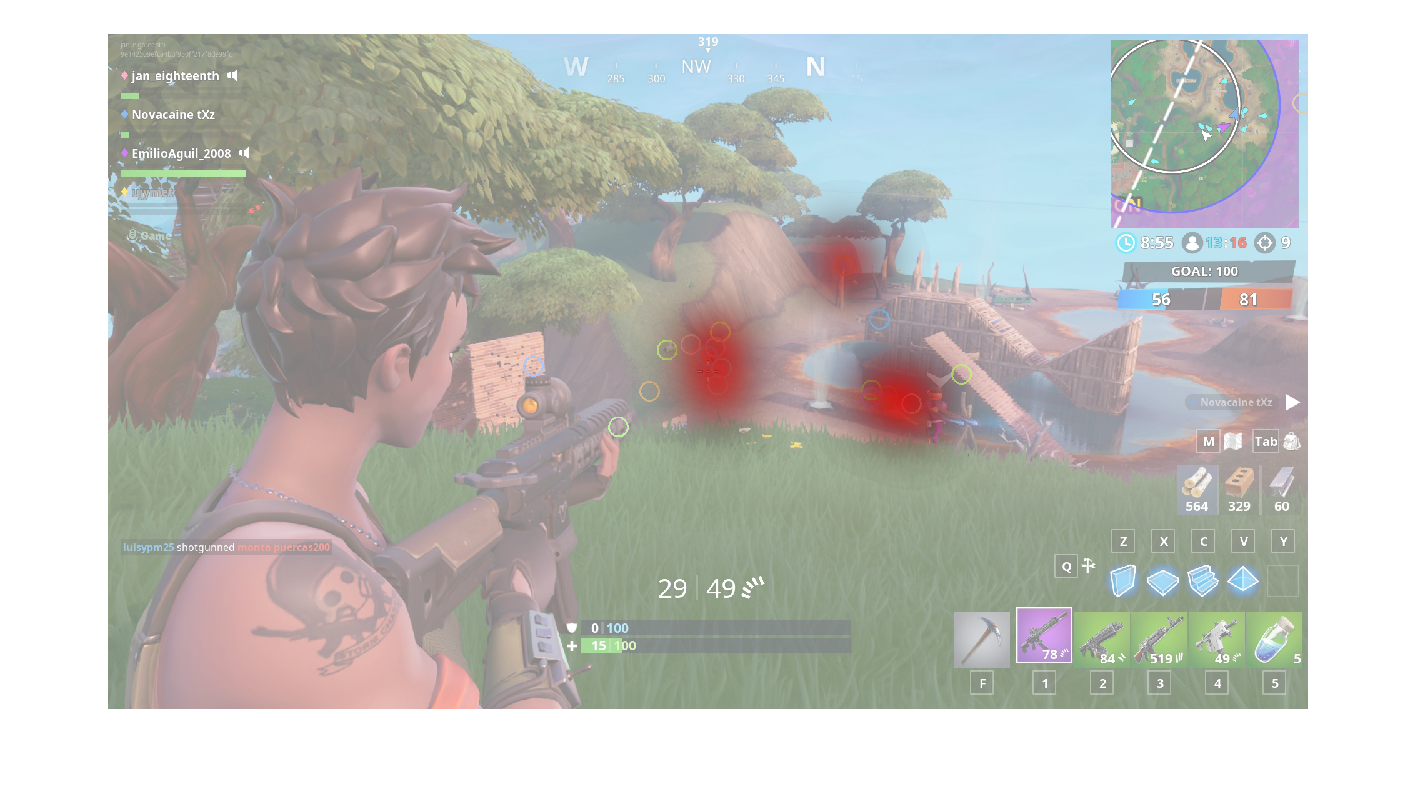} &
		\includegraphics[width=0.16\textwidth,trim={1.5cm 0.5cm 1.5cm 0.5cm},clip]{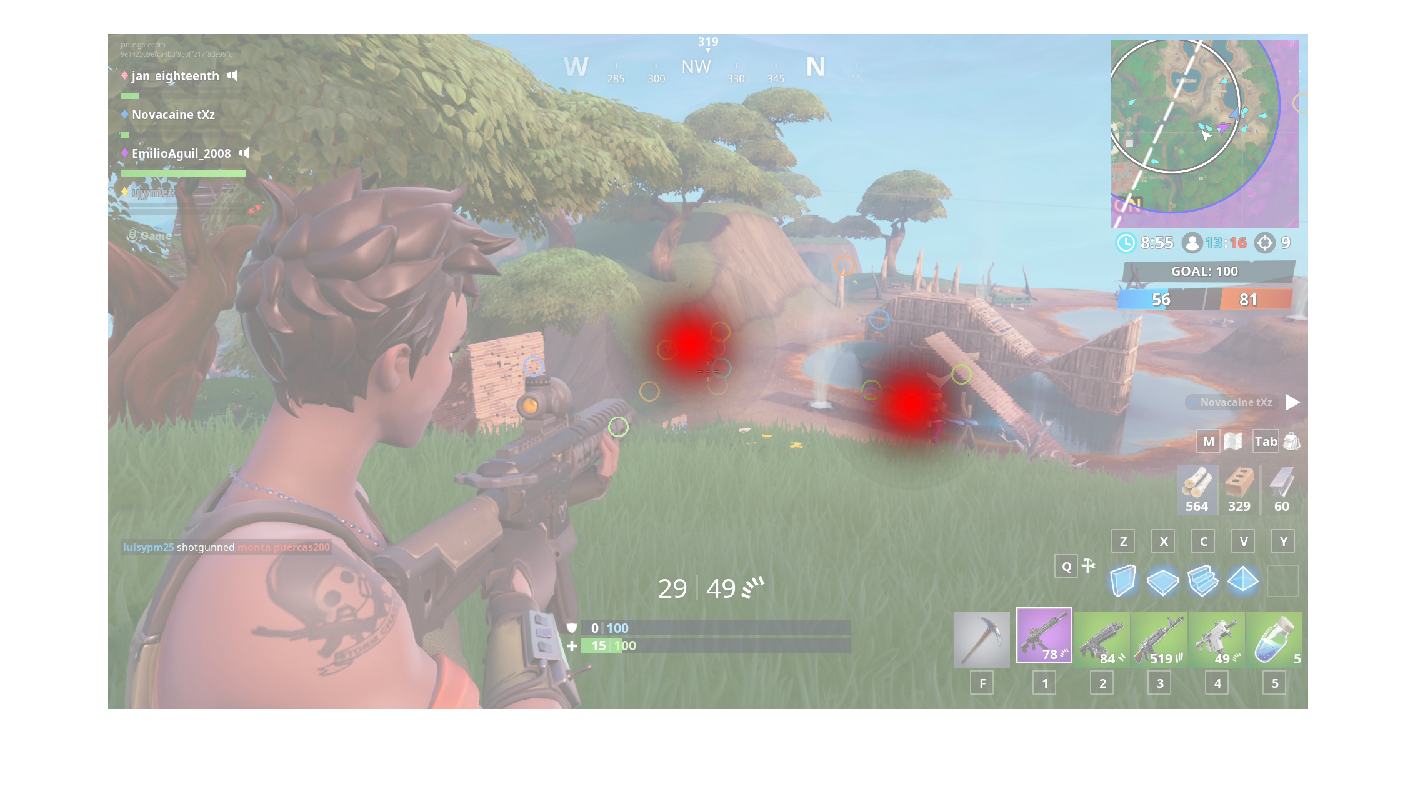} &		\includegraphics[width=0.16\textwidth,trim={1.5cm 0.5cm 1.5cm 0.5cm},clip]{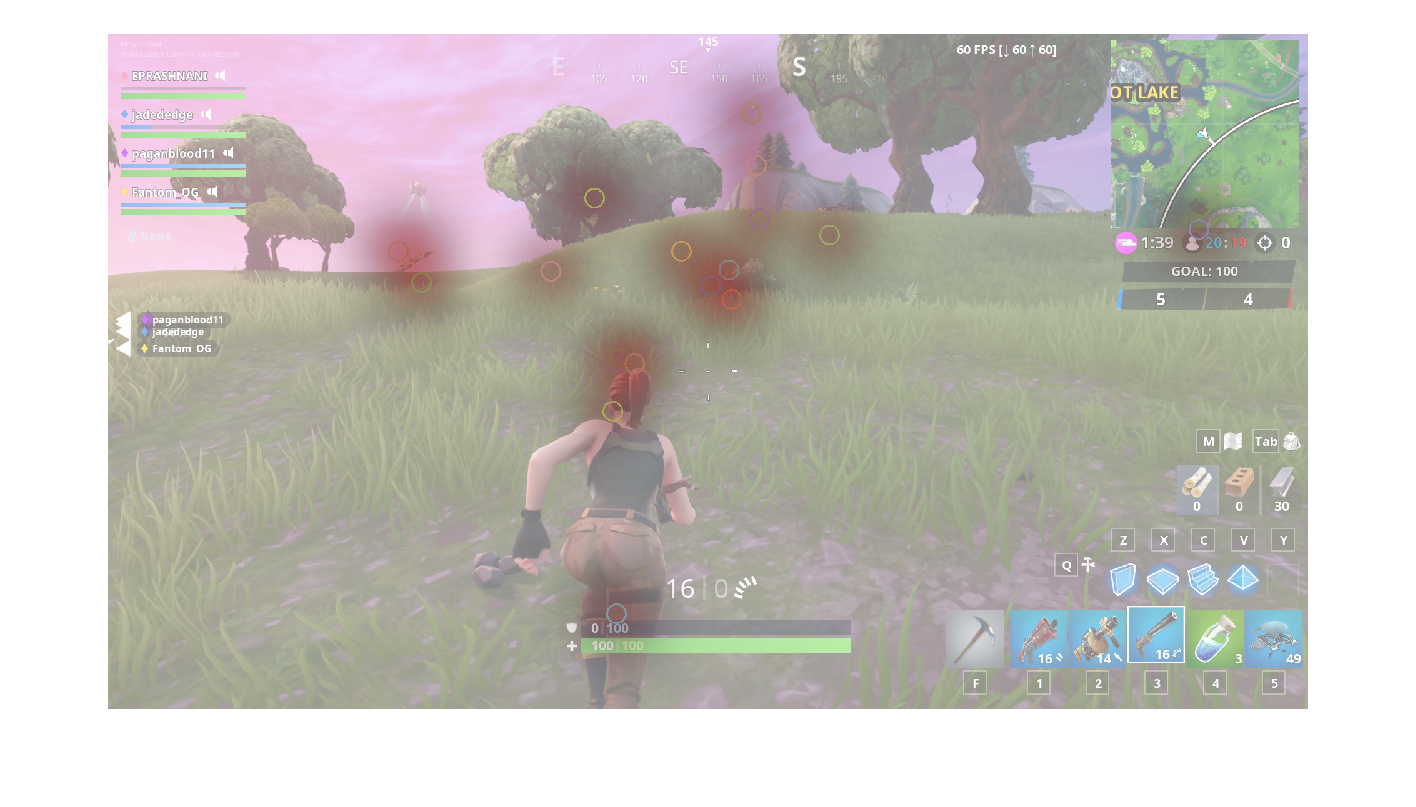} 
		&  \includegraphics[width=0.16\textwidth,trim={1.5cm 0.5cm 1.5cm 0.5cm},clip]{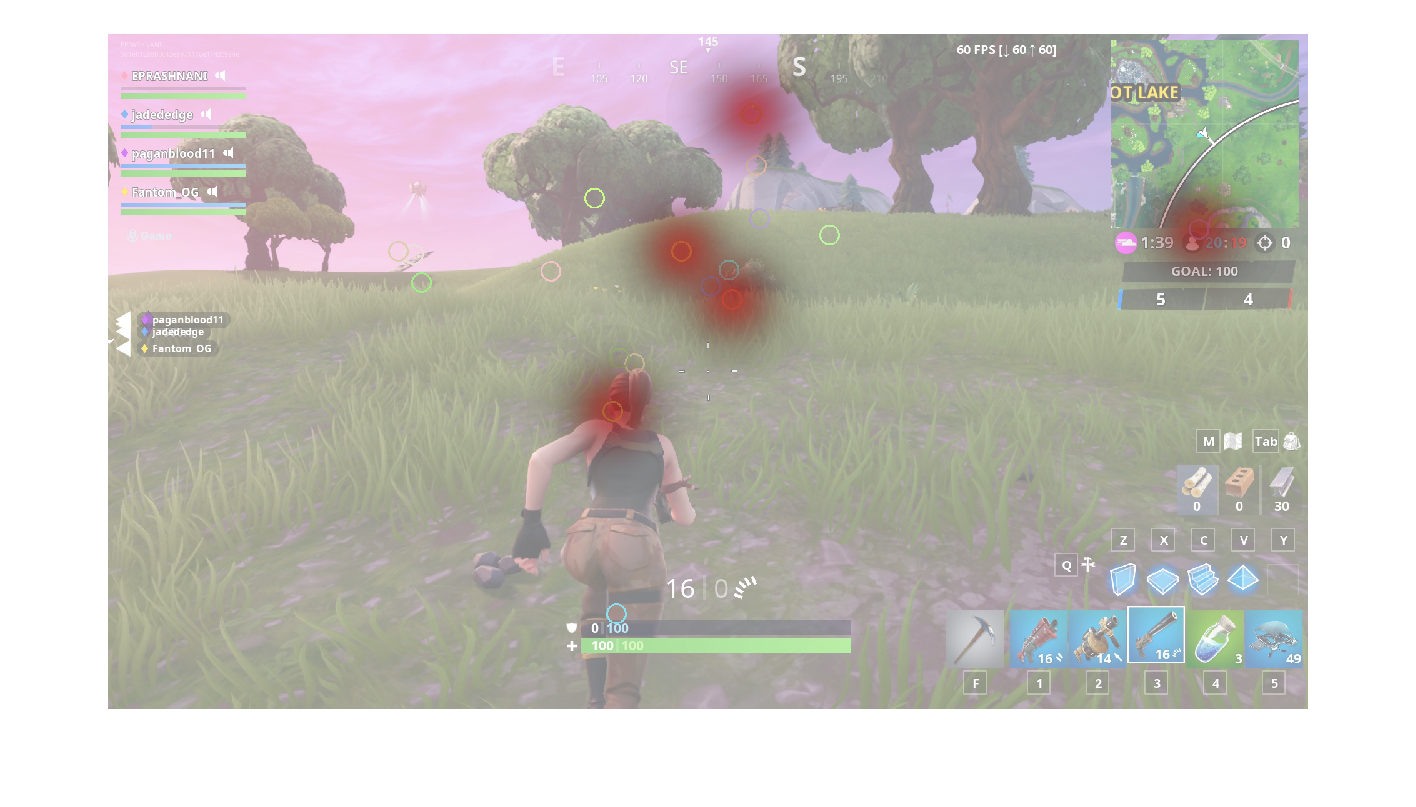} &
		\includegraphics[width=0.16\textwidth,trim={1.5cm 0.5cm 1.5cm 0.5cm},clip]{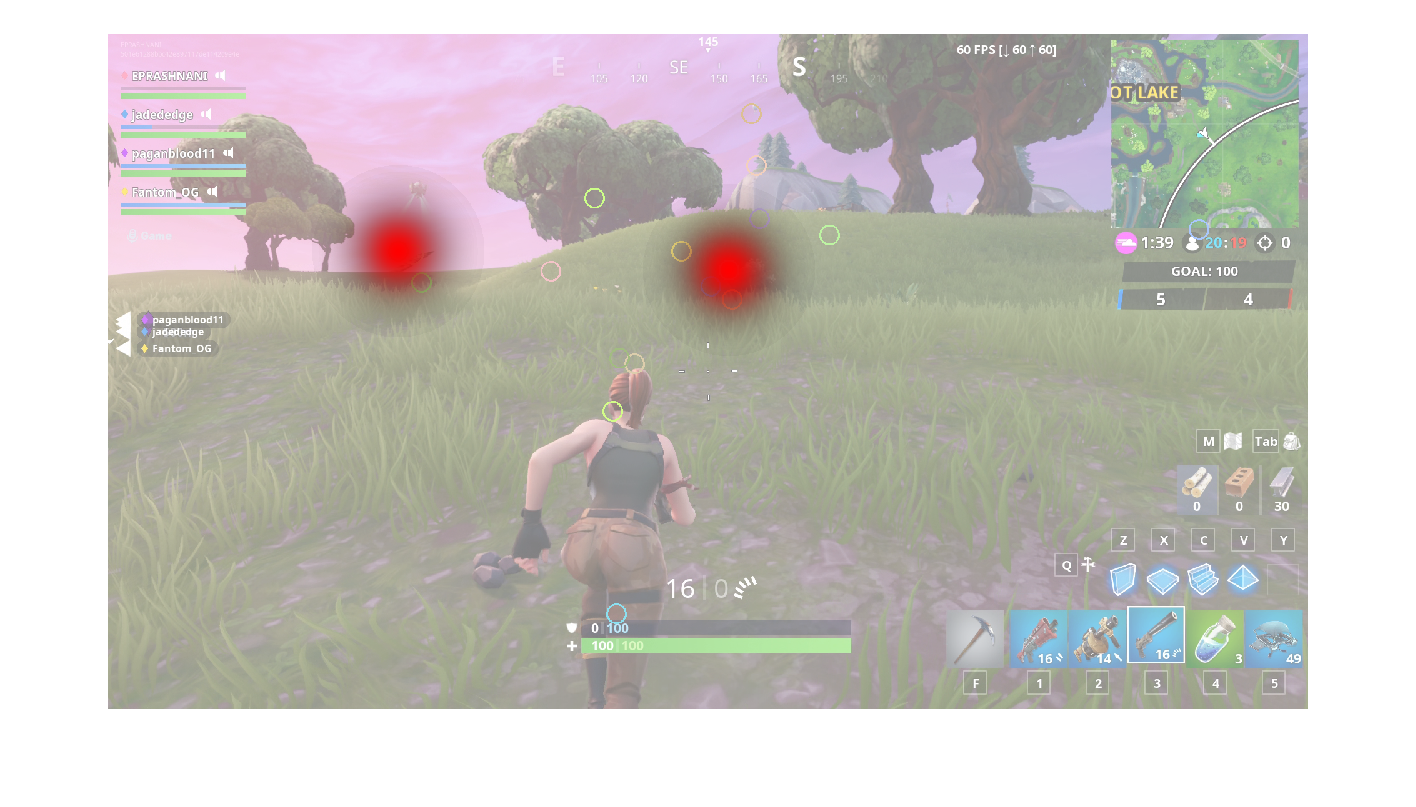} \vspace{-0.45cm}\\
		\textcolor{blue}{\small $0.372 \pm 0.106$} & \textcolor{blue}{\small $0.652 \pm 0.324$} & \textcolor{blue}{\small $1.079 \pm 0.691$} \vspace{0.1cm} & \textcolor{blue}{\small $0.517 \pm 0.154$} & \textcolor{blue}{\small $0.937 \pm 0.453$} & \textcolor{blue}{\small $0.947 \pm 0.722$} \vspace{0.1cm} \\

	\end{tabular}
	\endgroup
	\caption{Typical frames from ForGED with gaussian-blurred gaze locations of specified number of observers overlaid in red. For each image, we also show $\E[\text{KLD}(\tilde{x}_i||\tilde{\tilde{x}}_i)] \pm\text{Std}[\text{KLD}(\tilde{x}_i||\tilde{\tilde{x}}_i)]$. These quantities increase when the saliency map is sparse/multimodal and number of observers is small -- a setting that reduces the reliability of a frame for training. \small{\textit{ForGED images have been published with the permission from Epic Games.}}
	}
	\label{fig:datasets}
\end{figure}

Videogames present an interesting and challenging domain for saliency methods --  given their market value, dynamic content, multiple attractors of visual attention, and dependence of human gaze on temporal semantics.
We therefore introduce ForGED, a video-saliency dataset with $480$, $13$-second clips of Fortnite game play annotated with gaze data from up to $21$ observers per video. 
Compared to popular existing datasets such as LEDOV~\cite{Jia18} and DIEM~\cite{Mit11}, ForGED provides higher dynamism and a video-game context, with the highest number of frames at a consistent $1080$p resolution.
We summarize the characteristics of each of the datasets used in our experiments in Tab.~\ref{tab:datasets} and show typical ForGED frames in Fig.~\ref{fig:datasets}.

\begin{table*}[b!]
	\centering
	\resizebox{0.99\textwidth}{!}{
		\begin{tabular}{c|ccccccc}
			
			\multirow{2}{*}{Dataset}& \multirow{2}{*}{Videos}& \multirow{2}{*}{Frames}  & \multirow{2}{*}{Resolution} & Max. Observers & Total Obs. & \multirow{2}{*}{Content type}   & \addtext{mean $\pm$ std.dev.}\\
			&&&&used for training&for testing&&\addtext{of optical flow magnitude}\\
			\hline
			DIEM~\cite{Mit11}  & 84     & 240,015 & 720p@30Hz  & 31            & 51-219       & everyday videos (shows, advs, etc.) & \addtext{9.41$\pm$33.85}\\
			LEDOV~\cite{Jia18} & 538    & 179,336 & $\geq$720p@24Hz  & 32            & 32           & human/animal activities     & \addtext{4.09$\pm$8.65}          \\
			ForGED (ours)      & 480    & 374,400 & 1080p@60hz & 5 - 15            & 15-21        & videogame (Fortnite)       & \addtext{27.26$\pm$39.08}           \\
	\end{tabular}}
	\caption{Characteristics of video-saliency datasets, including the proposed ForGED dataset.}
	\label{tab:datasets}
	\vspace{-0.1in}
\end{table*}

\vspace{-0.1in}
\addtext{\paragraph{\textbf{Dynamic content in ForGED}.} To compare the dynamic content level of ForGED to those of LEDOV and DIEM, we use RAFT~\cite{Tee20} to compute the mean and standard deviation of the magnitude of the optical flow on a random subset of $100,000$ frames from the three datasets, at a uniform $1080$p resolution and $30$ fps framerate (Tab.~\ref{tab:datasets}).
This is in ForGED approximately $3\times$ that of DIEM and more than $6\times$ larger than LEDOV, suggesting that objects move faster (on average) in ForGED. It also has the largest standard deviation suggesting a larger variety of motion magnitudes in ForGED.}

\vspace{-0.1in}
\paragraph{Gaze data acquisition \addtext{and viewing behavior in ForGED}.}
To acquire ForGED, we first recorded 12 hours of Fortnite Battle Royale game-play videos from 8 players of varying expertise using OBS~\cite{OBD}. 
We then sampled $480$ $15$-second clips to show to a different set of $102$ participants  with varying degree of familiarity with Fortnite.
Each viewer was tasked with viewing a total of $48$ clips, randomly sampled from the pool of $480$, and interspersed with $3$-second ``intervals'' showing a central red dot on grey screen~\cite{Jia21} to ensure consistent gaze starting point for each clip (total 15-minute viewing time per viewer).
Each participant viewed the video clips on a $1080$p monitor situated approximately 80cm away, while their gaze was recorded with Tobii Tracker 4C at 90Hz.
After analyzing the gaze patterns, we discarded the initial $2$ seconds of each clip, when observers were mostly spending time to understand the context, to get a total of $374,400$ video frames annotated with gaze data. 
Accumulating the gaze of all frames, we observe that ForGED presents a bias towards the frame center and top-right corner.
This is because in Fortnite the main character and the crosshair lie at the screen center -- making it an important region -- and the mini-map on the top right corner attracts regular viewer attention to understand the terrain.
Such a bias is uniquely representative of the observer behavior not only in Fortnite, but also in other third person shooting games with similar scene layouts.
Further analysis of gaze data in ForGED, such as IOC curves, visual comparison of gaze data biases in ForGED, LEDOV and DIEM, are presented in the Supplementary, Sec.~3.

\vspace{-0.1in}
\section{Results}
\label{sec:results}
\addtext{We compare TT (Eq.~\ref{eq:vanilla_cost_function}) and NAT (Eq.~\ref{eq:vat_training_cost_function_real}) on three datasets (ForGED, LEDOV~\cite{Jia18}, and DIEM~\cite{Mit11}) and three DNNs (ViNet~\cite{Jai21}, the state-of-the-art on  DHF1K~\cite{Wan18}; TASED-Net, a 3D-CNN-based architecture~\cite{Min19}; and SalEMA, an RNN-based architecture~\cite{Lin19}). We further evaluate NAT against TT when density-based (\eg KLD) or fixation-based (\eg NSS) discrepancy functions are used as $d(\cdot,\cdot)$ in $J^{\text{real}}$ (Eq.~\ref{eq:vanilla_cost_function}) and $J_{\text{NAT}}^{\text{real}}$ (Eq.~\ref{eq:vat_training_cost_function_real}).}
\addtext{We first evaluated and improved the author-specified hyperparameters for ViNet, TASED-Net, and SalEMA, by performing TT (Eq.~\ref{eq:vanilla_cost_function}) on the entire LEDOV training set (see Tab.~\ref{tab:hyperparams}).
We use the improved settings for our experiments (see Supplementary Sec.~6).
We also verify that training on existing saliency datasets does not generalize to ForGED (Tab.~\ref{tab:vinet-ledov-forged}c), given its novel content.}
\begin{wraptable}{r}{0.6\columnwidth}
	\vspace{0.1in}
	\centering
	\resizebox{0.6\columnwidth}{!}{
		\renewcommand{\arraystretch}{1.05}
\scriptsize
\centering
\begin{tabular}[b]{c|c|c|c|c|c}
\textbf{method, hyperparameter settings} & \textbf{KLD}$\downarrow$ &\textbf{CC}$\uparrow$ &\textbf{SIM}$\uparrow$ & \textbf{NSS}$\uparrow$ &\textbf{AUC-J}$\uparrow$ \\
\hline
ViNET, Adam, $0.0001$, KLD (default)& $0.806$& $0.697$& $0.569$& $3.781$& $0.881$\\
ViNET, RMSprop, $0.0001$, KLD (improved)&$\mathbf{0.773}$& $\mathbf{0.710}$& $\mathbf{0.573}$& $\mathbf{3.969}$& $\mathbf{0.889}$\\
\hline
TASED-Net, SGD, learning rate schedule (default)& $1.104$& $0.554$& $0.452$& $2.536$& $0.828$\\
TASED-Net, RMSprop, $0.001$, KLD (improved)&$\mathbf{0.754}$& $\mathbf{0.724}$& $\mathbf{0.572}$& $\mathbf{4.227}$& $\mathbf{0.921}$\\

\hline
SalEMA, Adam, $1e^{-7}$, BCE (default)& $1.238$& $0.511$& $0.412$& $2.426$& $0.894$\\
SalEMA, RMSprop, $1e^{-5}$, KLD (improved)&$\mathbf{1.052}$& $\mathbf{0.612}$& $\mathbf{0.463}$& $\mathbf{3.237}$& $\mathbf{0.912}$\\
\end{tabular}
	}
	\caption{\addtext{LEDOV test-set performance when trained (traditionally) with default and improved settings for ViNet~\cite{Jai21}, TASED-Net ~\cite{Min19}, and SalEMA~\cite{Lin19}.}}
	\label{tab:hyperparams}
	\vspace{-0.07in}
\end{wraptable}
\vspace{-0.25in}
\paragraph{Experimental setup:} We want to compare TT and NAT when training with different amounts of data and varying levels of accuracy/gaze-data completeness.
We emulate small-size training datasets from LEDOV, DIEM, and ForGED by controlling the number of fixations, $N$, used to reconstruct $\tilde{x}$ in the training set and the number of training videos, $V$, used.
We report the performance evaluation of TT and NAT on test set for each $(V,N)$ value used for the training set.
\addtext{The values for $V$ and $N$ are chosen to gradually increase the training dataset size and accuracy until the maximum $V$ and/or $N$ is reached.}
\addtext{To reconstruct $\tilde{x}$, we choose a kernel of size $\sim 1^{\circ}$ viewing angle~\cite{Jia21,Ehi09,Wil11} and discuss alternative $\tilde{x}$ reconstruction strategies~\cite{Kum14, Kum17, Kum18} in Sec.~\ref{sec:discussion} (and Sec.~7 in the Supplementary)}.

\addtext{ForGED data are randomly split into 379 videos for training, 26 for validation, and 75 for testing.
	For LEDOV, we adopt the train / val / test split specified by the authors.
	DIEM contains gaze data from many observers on a few videos: we use $60$ videos with fewest observers for training and evaluate on the remaining videos with $51-219$ observers.}
\addtext{Evaluation is performed on test-set  maps reconstructed from the set of \textit{all} the available observers, that is sufficiently large to lead to converged IOC curves even for multimodal maps (see supplementary video for ForGED test set multimodality); consequently, we  also assume a negligible noise level in evaluation.}
\addtext{We omit experimenting with DHF1K in favor of LEDOV which is similar in scope to DHF1K~\cite{Tan20}, but contains a larger number of observers (converged IOC curves), while DHF1K lacks accurate per-observer gaze data.}

\vspace{-0.05in}
\paragraph{Dataset type and size:}
\label{subsec:results:datasize}
\addtext{We compare NAT and TT on different dataset types, by training ViNet and TASED-Net on ForGED, LEDOV, and DIEM, and changing $V$ and $N$ to assess the performance gain of NAT as a function of the level of accuracy and completeness of the training dataset.}
\addtext{Tab.~\ref{tab:vinet-ledov-forged}a and \ref{tab:vinet-ledov-forged}b show the results for ViNet trained on ForGED and LEDOV,}
whereas Tab.~\ref{tab:tased-discrep}a and \ref{tab:kld-arch}a show the results for TASED-Net.
\addtext{With ViNet, we observe a consistent performance gain of NAT over TT.
	Although NAT is particularly advantageous when $N$ and $V$ are small, training on the \textit{entire} LEDOV dataset (last row in Tab.~\ref{tab:vinet-ledov-forged}a) also shows a significant improvement for NAT since, depending on their content, some frames can still have insufficient fixation data.}
With TASED-Net trained on ForGED, NAT consistently outperforms TT when the number of training videos is $\leq 100$, \ie when noise overfitting may occur. 
Notably, NAT on 30 videos / 15 observers and 100 videos / 5 observers is comparable or superior to TT with 379 videos / 5 observers, which corresponds to $\geq3\times$ saving factor in terms of the data required for training.
Similar conclusions can be drawn for LEDOV (Tab.~\ref{tab:kld-arch}a) and DIEM (see Supplementary).
We also test the case of practical importance of an \textit{unbalanced} LEDOV dataset, with an uneven number of observers in the training videos.
Since NAT, by design, accounts for the varying reliability of the gaze data in training frames, it significantly outperforms TT (last two rows of Tab.~\ref{tab:kld-arch}a).

\begin{table}[h!]
	\begin{minipage}{0.49\columnwidth}
		\centering
		\resizebox{1.0\columnwidth}{!}{\scriptsize
\centering
\begin{tabular}[b]{c|c|c|c|c|c|c|c}
\textbf{train videos $V$} & \textbf{train obs. $N$} & \textbf{loss} & \textbf{KLD}$\downarrow$&\textbf{CC}$\uparrow$ &\textbf{SIM}$\uparrow$ & \textbf{NSS}$\uparrow$ &\textbf{AUC-J}$\uparrow$ \\
\hline
\multirow{8}{*}{$30$} & \multirow{2}{*}{$5$} & TT &$2.636$& $0.266$& $0.250$& $1.344$& $0.528$\\
& & NAT &$\mathbf{2.054}$& $\mathbf{0.406}$& $\mathbf{0.353}$& $\mathbf{1.979}$& $\mathbf{0.624}$\\
\cline{2-8}
& \multirow{2}{*}{$15$} & TT &$1.475$& $0.467$& $0.414$& $2.320$& $0.779$\\
& & NAT &$\mathbf{1.320}$& $\mathbf{0.502}$& $\mathbf{0.427}$& $\mathbf{2.467}$& $\mathbf{0.813}$\\
\cline{2-8}
& \multirow{2}{*}{$25$} & TT &$1.717$& $0.446$& $0.395$& $2.286$& $0.708$\\
& & NAT &$\mathbf{1.441}$& $\mathbf{0.482}$& $\mathbf{0.419}$& $\mathbf{2.450}$& $\mathbf{0.786}$\\
\cline{2-8}
& \multirow{2}{*}{$30-32$ (all)} & TT &$1.828$ & $0.448$ & $0.392$ & $2.281$ & $0.663$\\
& & NAT &$\mathbf{1.446}$&$\mathbf{0.491}$&$\mathbf{0.424}$&$\mathbf{2.462}$&$\mathbf{0.770}$\\
\hline
\multirow{2}{*}{$100$} & \multirow{2}{*}{$30-32$ (all)}& TT &$1.303$& $0.539$& $0.453$& $2.676$& $\mathbf{0.798}$\\
& & NAT &$\mathbf{1.275}$& $\mathbf{0.562}$& $\mathbf{0.471}$& $\mathbf{2.848}$& $0.784$\\
\hline
\multirow{2}{*}{$200$} &\multirow{2}{*}{$30-32$ (all)} & TT &$1.066$& $\mathbf{0.611}$& $\mathbf{0.511}$& $\mathbf{3.104}$& $0.840$\\
& & NAT &$\mathbf{1.020}$& $0.598$& $0.503$& $3.025$& $\mathbf{0.869}$\\
\hline
\multirow{2}{*}{$300$} & \multirow{2}{*}{$30-32$ (all)}& TT &$0.959$& $0.655$& $0.535$& $3.456$& $0.847$\\
& & NAT &$\mathbf{0.897}$& $\mathbf{0.669}$& $\mathbf{0.546}$& $\mathbf{3.517}$& $\mathbf{0.863}$\\
\hline
\multirow{2}{*}{$461$ (all)} & \multirow{2}{*}{$30-32$ (all)}& TT &$0.773$& $0.710$& $0.573$& $\mathbf{3.969}$& $0.889$\\
& & NAT &$\mathbf{0.718}$& $\mathbf{0.720}$& $\mathbf{0.577}$& $3.893$& $\mathbf{0.904}$\\
\end{tabular}
}
		\vspace{0.03in}
		\resizebox{0.8\columnwidth}{!}{\small\centerline{(a) ViNET on LEDOV, $d = $ KLD}}
	\end{minipage}%
	\hfill
	\begin{minipage}{0.49\columnwidth}
		\centering
		\vspace{-0.01in}
		\resizebox{1.0\columnwidth}{!}{\scriptsize
\centering
\begin{tabular}[b]{c|c|c|c|c|c|c|c}
\textbf{train videos $V$} & \textbf{train obs. $N$} & \textbf{loss} & \textbf{KLD}$\downarrow$&\textbf{CC}$\uparrow$ &\textbf{SIM}$\uparrow$ & \textbf{NSS}$\uparrow$ &\textbf{AUC-J}$\uparrow$ \\
\hline
\multirow{6}{*}{$30$} & \multirow{2}{*}{$5$} & TT &$1.538$& $0.541$& $0.426$& $3.261$& $0.713$\\
& & NAT &$\mathbf{1.264}$& $\mathbf{0.593}$& $\mathbf{0.460}$& $\mathbf{3.412}$& $\mathbf{0.773}$\\
\cline{2-8}
& \multirow{2}{*}{$10$} & TT &$1.779$& $0.514$& $0.399$& $3.130$& $0.633$\\
& & NAT &$\mathbf{1.220}$& $\mathbf{0.620}$& $\mathbf{0.488}$& $\mathbf{3.670}$& $\mathbf{0.764}$\\
\cline{2-8}
& \multirow{2}{*}{$15$} & TT &$\mathbf{1.218}$& $0.602$& $\mathbf{0.473}$& $\mathbf{3.542}$& $\mathbf{0.794}$\\
& & NAT &$1.257$& $\mathbf{0.605}$& $0.469$& $3.527$& $0.773$\\
\hline
\multirow{2}{*}{$100$} & \multirow{2}{*}{$5$} & TT &$1.263$& $0.600$& $0.473$& $3.609$& $0.775$\\
& & NAT &$\mathbf{1.149}$& $\mathbf{0.623}$& $\mathbf{0.485}$& $\mathbf{3.620}$& $\mathbf{0.798}$\\
\hline
\multirow{2}{*}{$200$} & \multirow{2}{*}{$5$} & TT &$1.134$& $0.629$& $\mathbf{0.494}$& $\mathbf{3.750}$& $0.804$\\
& & NAT &$\mathbf{0.982}$& $\mathbf{0.641}$& $0.489$& $3.704$& $\mathbf{0.882}$\\
\hline
\multirow{2}{*}{$379$} & \multirow{2}{*}{$5$} & TT &$\mathbf{0.994}$& $\mathbf{0.645}$& $\mathbf{0.495}$& $\mathbf{3.697}$& $0.860$\\
& & NAT &$1.026$& $0.625$& $0.438$& $3.505$& $\mathbf{0.918}$\\
\end{tabular}
}
		\vspace{0.07in}
		\resizebox{0.8\columnwidth}{!}{\small\centerline{(b) ViNET on ForGED, $d = $ KLD}}
		\resizebox{1.0\columnwidth}{!}{\renewcommand{\arraystretch}{1.05}
\scriptsize
\centering
\begin{tabular}[b]{c|c|c|c|c|c}
\textbf{training dataset} & \textbf{KLD}$\downarrow$ &\textbf{CC}$\uparrow$ &\textbf{SIM}$\uparrow$ & \textbf{NSS}$\uparrow$ &\textbf{AUC-J}$\uparrow$ \\
\hline
DHF1K& $2.038$ & $0.262$ & $0.228$ & $1.336$ & $0.805$\\
LEDOV&$1.573$& $0.436$& $0.345$& $2.583$& $0.818$\\
\end{tabular}
}
		\vspace{0.07in}
		\resizebox{0.8\columnwidth}{!}{\small\centerline{(c) pretrained ViNET tested on ForGED}}
	\end{minipage}

	\caption{\addtext{NAT vs. TT on (a) LEDOV and (b) ForGED with ViNet architecture trained on different training dataset sizes, using $d= $KLD as discrepancy. Best metrics between NAT and TT are in bold. The last two rows in (a) show the training on the \textit{entire} LEDOV dataset. (c) Training on existing large-scale video-saliency datasets shows poor generalization to ForGED since the videogame presents a very unique visual domain.}}
	\label{tab:vinet-ledov-forged}
\end{table}

\begin{table}[h!]
	\begin{minipage}{0.49\columnwidth}
		\centering
		\resizebox{1.0\columnwidth}{!}{\renewcommand{\arraystretch}{1.05}
\scriptsize
\centering
\begin{tabular}[b]{c|c|c|c|c|c|c|c}
\textbf{train videos $V$} & \textbf{train obs. $N$} & \textbf{loss} & \textbf{KLD}$\downarrow$&\textbf{CC}$\uparrow$ &\textbf{SIM}$\uparrow$ & \textbf{NSS}$\uparrow$ &\textbf{AUC-J}$\uparrow$ \\	
\hline
\multirow{6}{*}{$30$} & \multirow{2}{*}{$2$} & TT &$1.385$& $0.546$& $0.370$& $2.992$& $0.877$\\
& & NAT &$\mathbf{1.298}$& $\mathbf{0.558}$& $\mathbf{0.385}$& $\mathbf{3.161}$& $\mathbf{0.903}$\\
\cline{2-8}
& \multirow{2}{*}{$5$} & TT &$1.419$& $0.536$& $0.370$& $3.042$& $0.877$\\
& & NAT &$\mathbf{1.172}$& $\mathbf{0.590}$& $\mathbf{0.428}$& $\mathbf{3.372}$& $\mathbf{0.908}$\\
\cline{2-8}
& \multirow{2}{*}{$15$} & TT &$1.080$& $0.615$& $\mathbf{0.481}$& $3.598$& $0.897$\\
& & NAT &$\mathbf{0.995}$& $\mathbf{0.634}$& $0.478$& $\mathbf{3.750}$& $\mathbf{0.924}$\\
\hline
\multirow{4}{*}{$100$} & \multirow{2}{*}{$2$} & TT &$1.323$& $0.565$& $0.365$& $3.034$& $0.890$\\
& & NAT &$\mathbf{1.056}$& $\mathbf{0.610}$& $\mathbf{0.447}$& $\mathbf{3.386}$& $\mathbf{0.922}$\\
\cline{2-8}
& \multirow{2}{*}{$5$} & TT &$1.065$& $0.623$& $0.473$& $3.627$& $0.917$\\
& & NAT &$\mathbf{0.969}$& $\mathbf{0.643}$& $\mathbf{0.494}$& $\mathbf{3.749}$& $\mathbf{0.923}$\\
\hline
\multirow{4}{*}{$379$} & \multirow{2}{*}{$2$} & TT &$0.986$& $0.628$& $\mathbf{0.475}$& $3.434$& $0.925$\\
& & NAT &$\mathbf{0.974}$& $\mathbf{0.632}$& $0.470$& $\mathbf{3.497}$& $\mathbf{0.932}$\\
\cline{2-8}
& \multirow{2}{*}{$5$} & TT &$0.963$& $0.631$& $0.461$& $3.376$& $\mathbf{0.936}$\\
& & NAT &$\mathbf{0.888}$& $\mathbf{0.664}$& $\mathbf{0.508}$& $\mathbf{3.813}$& $0.934$\\
\end{tabular}
}
		\vspace{0.03in}
		\resizebox{0.8\columnwidth}{!}{\small\centerline{(a) TASED-Net on ForGED, $d = $ KLD}}
	\end{minipage}%
	\hfill
	\begin{minipage}{0.49\columnwidth}
		\centering
		\vspace{0.1in}
		\resizebox{1.0\columnwidth}{!}{\renewcommand{\arraystretch}{1.05}
\scriptsize
\centering
\begin{tabular}[b]{c|c|c|c|c|c|c|c}
\textbf{train videos $V$} & \textbf{train obs. $N$} & \textbf{loss} & \textbf{KLD}$\downarrow$&\textbf{CC}$\uparrow$ &\textbf{SIM}$\uparrow$ & \textbf{NSS}$\uparrow$ &\textbf{AUC-J}$\uparrow$ \\
\hline
\multirow{4}{*}{$30$} & \multirow{2}{*}{$5$} & TT &$1.155$& $0.612$& $0.440$& $3.600$& $0.904$\\
& & NAT &$\mathbf{1.061}$& $\mathbf{0.618}$& $\mathbf{0.466}$& $\mathbf{3.656}$& $\mathbf{0.912}$\\
\cline{2-8}
& \multirow{2}{*}{$15$} & TT &$1.095$& $0.612$& $0.448$& $3.574$& $0.919$\\
& & NAT &$\mathbf{0.993}$& $\mathbf{0.639}$& $\mathbf{0.475}$& $\mathbf{3.802}$& $\mathbf{0.928}$\\
\hline
\multirow{4}{*}{$100$} & \multirow{2}{*}{$2$} & TT &$1.138$& $\mathbf{0.601}$& $0.429$& $\mathbf{3.406}$& $0.911$\\
& & NAT &$\mathbf{1.099}$& $0.600$& $\mathbf{0.434}$& $3.356$& $\mathbf{0.920}$\\
\cline{2-8}
& \multirow{2}{*}{$5$} & TT &$1.097$& $0.623$& $0.425$& $3.533$& $0.921$\\
& & NAT &$\mathbf{1.016}$& $\mathbf{0.631}$& $\mathbf{0.468}$& $\mathbf{3.644}$& $\mathbf{0.924}$\\
\hline
\multirow{4}{*}{$379$} & \multirow{2}{*}{$2$} & TT &$1.069$& $0.618$& $0.436$& $3.456$& $0.920$\\
& & NAT &$\mathbf{1.011}$& $\mathbf{0.626}$& $\mathbf{0.450}$& $\mathbf{3.459}$& $\mathbf{0.931}$\\
\cline{2-8}
& \multirow{2}{*}{$5$} & TT &$0.958$& $0.655$& $0.467$& $3.652$& $\mathbf{0.934}$\\
& & NAT &$\mathbf{0.905}$& $\mathbf{0.669}$& $\mathbf{0.496}$& $\mathbf{3.946}$& $0.933$\\
\end{tabular}}
		\vspace{0.07in}
		\resizebox{0.8\columnwidth}{!}{\small\centerline{(b) TASED-Net on ForGED, $d = $ KLD - 0.1CC - 0.1NSS}}
	\end{minipage}
	\caption{NAT vs. TT on ForGED with TASED-Net architecture and different values of $N$,$V$, trained to minimize the discrepancy $d =$ KLD in (a), and $d = $ KLD - 0.1CC - 0.1NSS in (b).}
	\label{tab:tased-discrep}	
\end{table}

\begin{table}[h!]
	\begin{minipage}{0.49\columnwidth}
		\centering
		\resizebox{1.0\columnwidth}{!}{\renewcommand{\arraystretch}{1.05}
\scriptsize
\centering
\begin{tabular}[b]{c|c|c|c|c|c|c|c}
\textbf{train videos $V$} & \textbf{train obs. $N$} & \textbf{loss} & \textbf{KLD}$\downarrow$&\textbf{CC}$\uparrow$ &\textbf{SIM}$\uparrow$ & \textbf{NSS}$\uparrow$ &\textbf{AUC-J}$\uparrow$ \\
\hline
\multirow{6}{*}{$30$} & \multirow{2}{*}{$2$} & TT &$2.155$& $0.195$& $0.198$& $1.007$& $0.793$\\
& & NAT &$\mathbf{1.431}$& $\mathbf{0.428}$& $\mathbf{0.378}$& $\mathbf{2.082}$& $\mathbf{0.884}$\\
\cline{2-8}
& \multirow{2}{*}{$5$} & TT &$1.744$& $0.371$& $0.265$& $1.763$& $0.861$\\
& & NAT &$\mathbf{1.189}$& $\mathbf{0.495}$& $\mathbf{0.409}$& $\mathbf{2.378}$& $\mathbf{0.902}$\\
\cline{2-8}
& \multirow{2}{*}{$30$} & TT &$1.360$& $0.457$& $0.383$& $2.225$& $0.886$\\
& & NAT &$\mathbf{1.120}$& $\mathbf{0.532}$& $\mathbf{0.433}$& $\mathbf{2.638}$& $\mathbf{0.909}$\\
\hline
\multirow{6}{*}{$100$} & \multirow{2}{*}{$2$} & TT &$1.882$& $0.315$& $0.275$& $1.621$& $0.787$\\
& & NAT &$\mathbf{1.449}$& $\mathbf{0.457}$& $\mathbf{0.367}$& $\mathbf{2.281}$& $\mathbf{0.869}$\\
\cline{2-8}
& \multirow{2}{*}{$5$} & TT &$1.351$& $0.460$& $0.382$& $2.331$& $0.890$\\
& & NAT &$\mathbf{1.098}$& $\mathbf{0.554}$& $\mathbf{0.443}$& $\mathbf{2.753}$& $\mathbf{0.902}$\\
\cline{2-8}
& \multirow{2}{*}{$30$} & TT &$1.170$& $0.524$& $0.424$& $2.687$& $0.904$\\
& & NAT &$\mathbf{0.872}$& $\mathbf{0.648}$& $\mathbf{0.493}$& $\mathbf{3.604}$& $\mathbf{0.932}$\\
\hline
\multirow{8}{*}{$461$} & \multirow{2}{*}{$2$} & TT &$1.231$& $0.532$& $0.459$& $2.784$& $0.880$\\
& & NAT &$\mathbf{0.975}$& $\mathbf{0.595}$& $\mathbf{0.499}$& $\mathbf{2.931}$& $\mathbf{0.921}$\\
\cline{2-8}
& \multirow{2}{*}{$5$} & TT &$\mathbf{0.805}$& $\mathbf{0.684}$& $\mathbf{0.552}$& $\mathbf{3.788}$& $0.921$\\
& & NAT &$0.828$& $0.667$& $0.531$& $3.530$& $\mathbf{0.929}$\\
\cline{2-8}
& \multirow{2}{*}{$30-32$ (all)} & TT &$0.754$& $0.724$& $0.572$& $\mathbf{4.227}$& $0.921$\\
& & NAT &$\mathbf{0.686}$& $\mathbf{0.727}$& $\mathbf{0.575}$& $4.128$& $\mathbf{0.937}$\\
\cline{2-8}
& \multirow{2}{*}{$2,5,15,30$} & TT &$0.836$& $0.666$& $\mathbf{0.551}$& $3.615$& $0.916$\\
& & NAT &$\mathbf{0.768}$& $\mathbf{0.692}$& $0.545$& $\mathbf{3.855}$& $\mathbf{0.933}$\\
\end{tabular}
}
		\vspace{0.03in}
		\resizebox{0.8\columnwidth}{!}{\small\centerline{(a) TASED-Net on LEDOV, $d = $ KLD}}
	\end{minipage}%
	\hfill
	\begin{minipage}{0.49\columnwidth}
		\centering
		\vspace{-0.02in}
		\resizebox{1.0\columnwidth}{!}{\renewcommand{\arraystretch}{1.05}
\scriptsize
\centering
\begin{tabular}[b]{c|c|c|c|c|c|c|c}
\textbf{train videos $V$} & \textbf{train obs. $N$} & \textbf{loss} & \textbf{KLD}$\downarrow$&\textbf{CC}$\uparrow$ &\textbf{SIM}$\uparrow$ & \textbf{NSS}$\uparrow$ &\textbf{AUC-J}$\uparrow$ \\
\hline
\multirow{6}{*}{$30$} & \multirow{2}{*}{$2$} & TT &$1.922$& $0.249$& $0.232$& $1.039$& $0.803$\\
& & NAT &$\mathbf{1.768}$& $\mathbf{0.286}$& $\mathbf{0.285}$& $\mathbf{1.263}$& $\mathbf{0.843}$\\
\cline{2-8}
& \multirow{2}{*}{$5$} & TT &$2.168$& $0.280$& $0.276$& $1.348$& $0.844$\\
& & NAT &$\mathbf{1.710}$& $\mathbf{0.327}$& $\mathbf{0.298}$& $\mathbf{1.476}$& $\mathbf{0.848}$\\
\cline{2-8}
& \multirow{2}{*}{$30$} & TT &$1.888$& $0.256$& $0.225$& $1.082$& $0.821$\\
& & NAT &$\mathbf{1.510}$& $\mathbf{0.404}$& $\mathbf{0.321}$& $\mathbf{1.969}$& $\mathbf{0.874}$\\
\hline
\multirow{6}{*}{$100$} & \multirow{2}{*}{$2$} & TT &$1.621$& $0.355$& $0.307$& $1.634$& $0.854$\\
& & NAT &$\mathbf{1.538}$& $\mathbf{0.385}$& $\mathbf{0.311}$& $\mathbf{1.733}$& $\mathbf{0.867}$\\
\cline{2-8}
& \multirow{2}{*}{$5$} & TT &$1.381$& $0.455$& $0.363$& $2.179$& $0.882$\\
& & NAT &$\mathbf{1.340}$& $\mathbf{0.470}$& $\mathbf{0.392}$& $\mathbf{2.368}$& $\mathbf{0.893}$\\
\cline{2-8}
& \multirow{2}{*}{$30$} & TT &$1.359$& $0.532$& $0.408$& $2.909$& $0.883$\\
& & NAT &$\mathbf{1.284}$& $\mathbf{0.559}$& $\mathbf{0.408}$& $\mathbf{3.272}$& $\mathbf{0.884}$\\
\hline
\multirow{6}{*}{$461$} & \multirow{2}{*}{$2$} & TT &$1.277$& $0.487$& $0.382$& $2.247$& $0.895$\\
& & NAT &$\mathbf{1.243}$& $\mathbf{0.490}$& $\mathbf{0.403}$& $\mathbf{2.365}$& $\mathbf{0.899}$\\
\cline{2-8}
& \multirow{2}{*}{$5$} & TT &$1.139$& $\mathbf{0.568}$& $0.444$& $2.825$& $0.903$\\
& & NAT &$\mathbf{1.136}$& $0.567$& $\mathbf{0.450}$& $\mathbf{3.117}$& $\mathbf{0.908}$\\
\cline{2-8}
& \multirow{2}{*}{$30$} & TT &$1.052$& $0.612$& $\mathbf{0.462}$& $3.237$& $\mathbf{0.912}$\\
& & NAT &$\mathbf{1.045}$& $\mathbf{0.633}$& $0.457$& $\mathbf{3.425}$& $0.910$\\
\end{tabular}
}
		\vspace{0.02in}
		\resizebox{0.8\columnwidth}{!}{\small\centerline{(b) SalEMA on LEDOV, $d = $ KLD}}
	\end{minipage}
	\vspace{0.03in}
	\caption{NAT vs. TT on LEDOV dataset for different DNN architectures -- TASED-Net in (a) and SalEMA in (b) -- with $d = \text{KLD}$ and different training data sizes. The last two rows in (a) show the case of an unbalanced dataset with $N$ chosen from ${2,5,15,30}$ in a video.}
	\label{tab:kld-arch}
	\vspace{-0.08in}
\end{table}

\vspace{-0.05in}
\paragraph{Discrepancy functions:}
NAT can be applied to any choice of discrepancy $d$.
To demonstrate this, a mix of density- and fixation-based discrepancies, $d = \text{KLD}-0.1\text{CC}-0.1\text{NSS}$, which has also been a popular choice in literature~\cite{Dro20,Wan18}, is used to train TASED-Net on ForGED (Tab.~\ref{tab:tased-discrep}b).
Comparing Tab.~\ref{tab:tased-discrep}a and Tab.~\ref{tab:tased-discrep}b, we note that NAT provides a performance gain over TT, independently of the training discrepancy.
We show more experiments in the Supplementary \addtext{(Sec.~5)}, with a fixation-based metric (NSS), and on different datasets.

\vspace{-0.05in}
\paragraph{DNN architectures:}
Tab.~\ref{tab:kld-arch}a-b compares NAT vs. TT when training two different DNNs (TASED-Net~\cite{Min19} and SalEMA~\cite{Lin19}) on LEDOV, with KLD.
As also observed earlier, NAT outperforms TT and the performance gap shrinks with increasing training data.
The Supplementary \addtext{(Sec.~5)} shows results with SalEMA on ForGED.
\vspace{-0.1in}
\section{Discussion and conclusions}
\label{sec:discussion}
\textbf{NAT for images:} Image-based saliency datasets (\eg CAT2000~\cite{Bor15}, SALICON~\cite{Jia17}) have many fixations per image resulting in high-quality of the reconstructed saliency maps, as the accuracy rapidly increases with number of fixations (e.g., $>90\%$ accuracy at $20$ fixations~\cite{Jud12}). 
It is nonetheless fair to ask if NAT is effective for image-saliency predictors.
\begin{wraptable}{r}{0.6\columnwidth}
	\vspace{-0.1in}
	\centering
	\resizebox{0.6\columnwidth}{!}{
		\renewcommand{\arraystretch}{1.05}
\scriptsize
\centering
\begin{tabular}[b]{c|c|c|c|c|c|c}
\textbf{no. of fixations} & \textbf{loss} & \textbf{KLD}$\downarrow$ &\textbf{CC}$\uparrow$ &\textbf{SIM}$\uparrow$ & \textbf{NSS}$\uparrow$ &\textbf{AUC-J}$\uparrow$ \\
\hline
\multirow{2}{*}{$5$} & TT &$3.986$& $0.578$& $0.537$& $1.477$& $0.764$\\
& NAT &$\mathbf{1.672}$& $\mathbf{0.660}$& $\mathbf{0.611}$& $\mathbf{1.549}$& $\mathbf{0.817}$\\
\hline
\multirow{2}{*}{$15$} & TT &$2.877$& $0.655$& $0.589$& $1.669$& $0.795$\\
& NAT &$\mathbf{1.437}$& $\mathbf{0.714}$& $\mathbf{0.640}$& $\mathbf{1.676}$& $\mathbf{0.831}$\\
\end{tabular}
}
	\caption{Evaluation on EML-Net.}
	\label{tab:imsal}
	\vspace{-0.1in}
\end{wraptable}
We simulate a high-noise, incomplete, dataset by sampling a subset of fixations for each SALICON image\footnote{Mouse clicks are used as proxy for gaze in SALICON.} and train a state-of-the-art method, EML-Net~\cite{Jia20}, with TT and NAT. 
Tab.~\ref{tab:imsal} shows the results on the official SALICON benchmark test set, and confirms the advantage of NAT.

\vspace{-0.1in}
\paragraph{Alternative methods to reconstruct \boldmath{$\tilde{x}$}:}
Although reconstructing $\tilde{x}$ by blurring a binary map of fixations is prevalent practice~\cite{Jia21,Ehi09,Wil11}, we experiment with another reconstruction strategy for $\tilde{x}$ using Gaussian KDE with a uniform regularization. 
The optimal KDE bandwidth and regularization weight is estimated by optimizing a gold-standard model~\cite{Kum15,Tan20} (see Supplementary, Sec.~7).
Experiments with TASED-Net on ForGED ($N=5$, $V=30$) comparing TT with $\tilde{x}$ estimated using a fixed-size blur or KDE-based reconstruction, and NAT, show that while KDE improves TT, NAT still yields the best results (Tab.~\ref{tab:kde}).
\begin{wraptable}{r}{0.6\columnwidth}
	\vspace{-0.1in}
	\centering
	\resizebox{0.6\columnwidth}{!}{
		\renewcommand{\arraystretch}{1.05}
\scriptsize
\centering
\begin{tabular}[b]{c|c|c|c|c|c|c}
	\textbf{$\mathbf{\tilde{x}_i}$} for training & \textbf{loss} & \textbf{KLD}$\downarrow$ & \textbf{CC}$\uparrow$ & \textbf{SIM}$\uparrow$ & \textbf{NSS}$\uparrow$ & \textbf{AUC-J}$\uparrow$ \\\hline
	$1^{\circ}$ blur & TT & 1.419 & 0.536 & 0.370 & 3.042  & 0.877 \\
	KDE & TT & 1.223 & 0.573 & 0.399 & 3.271 & 0.897 \\
	$1^{\circ}$ blur & NAT & \textbf{1.172} & \textbf{0.590} & \textbf{0.428} & \textbf{3.372} & \textbf{0.908}\\
	
\end{tabular}}
	\caption{Methods for estimating $\tilde{x}$.}
	\label{tab:kde}
	\vspace{-0.15in}
\end{wraptable}

\vspace{-0.2in}
\paragraph{Limitations and future work:}
\addtext{Although the test saliency maps of LEDOV, DIEM and ForGED are derived from several observers leading to converged IOC on \textit{average}, per-frame inaccuracies of saliency maps can still add uncertainty about the conclusions one can draw.
Adopting alternative strategies such as deriving metric-specific saliency from the probabilistic output of a saliency predictor~\cite{Byl18, Kum18}, can give a clearer understanding.
Nonetheless, in our experiments all the metrics are generally in agreement about the ranking between TT and NAT: a strong evidence in favor of NAT~\cite{Ric13}.
NAT design principles can also be applied to saliency evaluation (not only training), where variable importance is given to each frame depending on its noise level.}

\vspace{-0.1in}
\paragraph{Conclusion:}
\addtext{Video gaze data acquisition is time-consuming and can be inaccurate.
To reduce the impact of dataset size in the field of visual saliency prediction, we introduce NAT to account for the level of reliability of a saliency map.
We also introduce a new dataset which offers a unique video-game context.
We show consistent improvements for NAT over TT across a variety of experiments.
The adoption of NAT has important practical implications, since it allows acquiring new datasets (or training on old ones) with less data, both in terms of videos and number of observers, without loss of quality.}

\section*{Acknowledgments}
We thank Abhishek Badki and Shoaib Ahmed Siddiqui for help with distributed training, Jan Kautz, Ming-Yu Liu, and Arash Vahdat for technical discussions, Michael Stengel for help with gaze-recording software, Paul Weakliem and Fuzzy Rogers for UCSB computational infrastructure, and the Fortnite players and observers for their help with creating ForGED. This work was supported in part through computational facilities purchased using NSF grant OAC-1925717 and administered by the Center for Scientific Computing (CSC) at UCSB. The CSC is supported by the California NanoSystems Institute and the Materials Research Science and Engineering Center (MRSEC; NSF DMR 1720256).

\bibliographystyle{bmvc2k_natbib}
\bibliography{egbib}

\newpage
\clearpage
\setcounter{section}{0}
\setcounter{figure}{0}
\setcounter{table}{0}
\setcounter{footnote}{0}
\setcounter{equation}{0}

\newpage
\null	
{\noindent\textcolor{myblue}{\textbf{\Large{Noise-Aware Video Saliency Prediction\\
(Supplementary Material)}}}}

\vspace{0.3in}
\hrule
\vspace{0.4in}
\vspace{-0.2in}
\section{Derivation of NAT cost function (mentioned in Sec.~3)}

We interpret $d(x_i, \tilde{x}_i)$ as a random variable with Gaussian distribution, $d(x_i, \tilde{x}_i) \sim G(\mu_i, \sigma_i^2)$, where $\mu_i = E[d(x_i, \tilde{x}_i)]$ indicates its mean, whereas $\sigma_i^2 = \text{Var}[d(x_i, \tilde{x}_i)]$ is its variance.
When the predicted saliency map $\hat{x}_i$ is optimal, \ie when $\hat{x}_i = x_i$, $d(\hat{x}_i, \tilde{x}_i)$ has the same statistical distribution of $d(x_i, \tilde{x}_i)$.
Therefore, for a perfect saliency predictor, we can write  $d(\hat{x}_i, \tilde{x}_i) \sim G(\mu_i, \sigma_i^2)$.
Note that, for our proposed noise-aware training (NAT), $\mu_i$ and $\sigma_i$ are assumed to be known, and therefore, $\hat{x}_i$ is the only unknown.
The likelihood of $d(\hat{x}_i, \tilde{x}_i)$ is given by:

\begin{equation}
	p[d(\hat{x}_i, \tilde{x}_i)] = \frac{1}{\sqrt{2\pi}\sigma_i}e^{-\frac{[d(\hat{x}_i, \tilde{x}_i)-\mu_i]^2}{2\sigma_i^2}}.
	\label{eq:likelihood_single}
\end{equation}
Given our interpretation of $d(\hat{x}_i, \tilde{x}_i)$, for a dataset containing $N+1$ saliency maps, the negative log likelihood is:

\begin{eqnarray}
	J(\hat{x}_0, \hat{x}_1, ..., \hat{x}_N) = -\text{ln}\prod\nolimits_i \frac{1}{\sqrt{2\pi}\sigma_i}e^{-\frac{[d(\hat{x}_i, \tilde{x}_i)-\mu_i]^2}{2\sigma_i^2}} = \nonumber  \\
	\sum\nolimits_i -\text{ln}\{\frac{1}{\sqrt{2\pi}\sigma_i}e^{-\frac{[d(\hat{x}_i, \tilde{x}_i)-\mu_i]^2}{2\sigma_i^2}}\} = \nonumber \\
	\sum\nolimits_i \{ \text{ln}(\sqrt{2\pi}\sigma_i) +\frac{[d(\hat{x}_i, \tilde{x}_i)-\mu_i]^2}{2\sigma_i^2} \}.
	\label{eq:loglikelihood}
\end{eqnarray}%

We want to train the saliency models to predict all the $\{\hat{x}_i\}_{i=0... N}$ that maximize the likelihood.
Therefore, the optimization problem becomes:
\begin{eqnarray}
	(\hat{x}_0, \hat{x}_1, ..., \hat{x}_N) = \underset{(\hat{x}_0, \hat{x}_1, ..., \hat{x}_N)}{\mathrm{argmin}} J(\hat{x}_0, \hat{x}_1, ..., \hat{x}_N) = \nonumber \\
	\underset{(\hat{x}_0, \hat{x}_1, ..., \hat{x}_N)}{\mathrm{argmin}} \sum\nolimits_i \{\text{ln}(\sqrt{2\pi}\sigma_i) +\frac{[d(\hat{x}_i, \tilde{x}_i)-\mu_i]^2}{2\sigma_i^2} \}.
\end{eqnarray}
Upon simplification (removing the terms that do not depend on $(\hat{x}_0, \hat{x}_1, ..., \hat{x}_N)$, that are the only unknowns), we obtain: 
\begin{eqnarray}
	(\hat{x}_0, \hat{x}_1, ..., \hat{x}_N) = \underset{(\hat{x}_0, \hat{x}_1, ..., \hat{x}_N)}{\mathrm{argmin}} \sum\nolimits_i \{\frac{[d(\hat{x}_i, \tilde{x}_i)-\mu_i]^2}{\sigma_i^2} \}.
\end{eqnarray}
This leads to the formulation of the NAT cost function:
\begin{eqnarray}
	J_{\text{NAT}}^{\text{ideal}} = \sum\nolimits_i \{\frac{[d(\hat{x}_i, \tilde{x}_i)-\mu_i]^2}{\sigma_i^2} \} =
	\sum\nolimits_i \{\frac{[d(\hat{x}_i, \tilde{x}_i)-\E[d(x_i, \tilde{x}_i)]]^2}{\text{Var}[d(x_i, \tilde{x}_i)]} \}.
\end{eqnarray}

\section{A toy example to motivate NAT (mentioned in Sec. 3)}
\begin{figure*}[h!]
	\centering
	\includegraphics[width=0.8\textwidth, trim=0cm 0cm 0cm 0cm, clip=true]{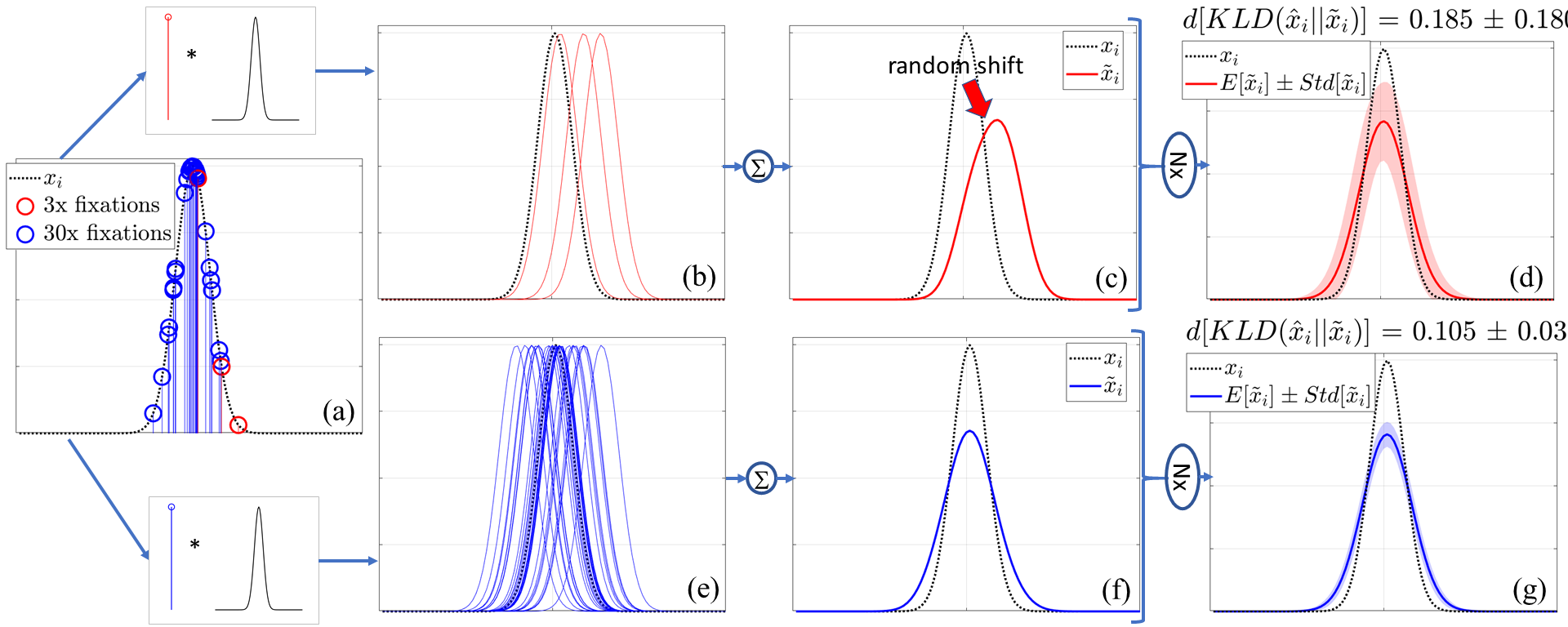}
	\includegraphics[width=0.8\textwidth, trim=0cm 0cm 0cm 0cm, clip=true]{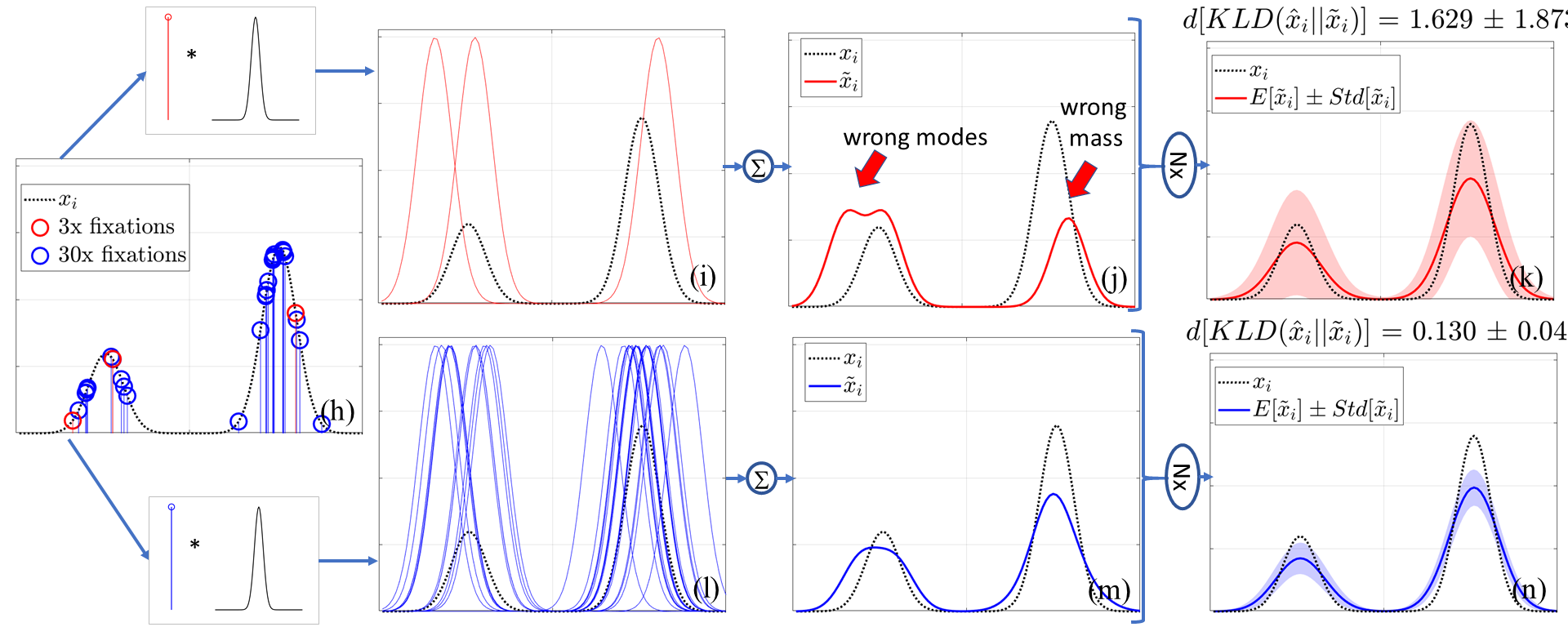}
	\caption{A toy example to motivate NAT. Plots in (a) and (h) show the unimodal and multimonal $1D$ pdfs 	$x_i$ in dashed black lines -- these are analogous to the true underlying $2D$ saliency maps for video frames / images. The measured (or training) saliency maps are reconstructed by first sampling ``fixations'' (red / blue circles in (a, h)) from $x_i$, then blurring (b, e, i, l), summing, and normalizing to obtain the resulting reconstructed saliency maps $\tilde{x}_i$ (d, g, j, m). When a limited number of observers is available (\eg 3, in the red plots in (c,j)), the resulting reconstructed $\tilde{x}_i$ may differ in shape from $x_i$, \eg due to random shifts, reconstruction errors, etc. Plots d, g, k, n show the expected value and standard deviation for multiple realizations of $\tilde{x}_i$, with respect to $x_i$. The deviation of $\tilde{x}_i$ from $x_i$ results in the statistics $\E[\text{KLD}(x_i,\tilde{x}_i)]$ and $\text{Var}[\text{KLD}(x_i,\tilde{x}_i)]$ to be non-zero (as shown in the titles of plots d, g, k, n). Furthermore, as is evident from these plots, these statistics are larger when few observers are available and when $x_i$ has a complex shape (\eg multimodal), which makes $x_i$ more susceptible to inaccurate approximation using $\tilde{x}_i$. The plots considered here for $x_i$ are: a Gaussian centered at $\mu=50$, $\sigma=5$; and a mixture with two components at $\mu=[25, 75]$, probabilities $P=[0.3, 0.7]$, and $\sigma=5$.}
	\label{fig:toy_example_new}
\end{figure*}

Assume that a method predicts the (unobservable) distribution $x_i$ exactly, that is $\hat{x}_i = x_i$.
Because of measurement noise and incomplete sampling in $\tilde{x}_i$ (which is the the saliency map estimated from insufficient gaze data, \ie the one typically used for training), $d(x_i,\tilde{x}_i) \neq 0$, even though the prediction is perfect.
In this scenario, it would suboptimal to train a saliency predictor to minimize $d(x_i,\tilde{x}_i)$. 

Let us consider a 1D toy example: Figs.~\ref{fig:toy_example_new}(a,h) show two 1D ground-truth ``saliency maps'' (or pdfs) $x_i$, one unimodal, and one bimodal.
We simulate the ``1D gaze-data acquisition'' by sampling 3 (red circles) or 30 (blue) spatial locations (or ``gaze fixations'') from $x_i$.
Following the \emph{de facto} standard to generate saliency maps from single gaze locations, we blur each fixation (Fig.~\ref{fig:toy_example_new}(b)), and accumulate the resulting curves (Fig.~\ref{fig:toy_example_new}(c)).
This results in approximations, $\tilde{x_i}$, of the ground-truth saliency maps.
The inaccurate positions of the modes in these estimated saliency maps mimics the measurement noise, while the finite number of 1D gaze fixations used to estimate these maps simulates incomplete sampling.

When few fixations are available, $\tilde{x}_i$ may be shifted with respect to $x_i$ (Fig.~\ref{fig:toy_example_new}(c)), and the number of its modes may not match $x_i$ (Fig.~\ref{fig:toy_example_new}(j)).
Furthermore, when $x_i$ is multimodal, the mass of each mode in $\tilde{x}_i$ may be imprecisely estimated compared to $x_i$ (Fig.~\ref{fig:toy_example_new}(j)).
The standard deviation of $1000$ random realizations of $\tilde{x}_i$ ($\text{Std}[\tilde{x}_i]$), which measures the uncertainty in $\tilde{x}_i$ (and therefore the quality of estimation of $x_i$ using $\tilde{x}_i$), decreases when a large number of fixations are used to reconstruct $\tilde{x}_i$ and remains high for a smaller number of fixations.
This is shown as the light-blue / light-red shaded regions in Figs.~\ref{fig:toy_example_new}(d, g, k, n), while the solid plot red / blue curve shows $\E[\tilde{x}_i]$.
Furthermore, the level of uncertainty is proportional the complexity of the ground-truth saliency map: \eg \ given $3$ fixations to reconstruct $\tilde{x_i}$, the uncertainty is lower when the underlying ground-truth $x_i$ map  is unimodal (Fig.~\ref{fig:toy_example_new}(d)), and higher when $x_i$ is bimodal Fig.~\ref{fig:toy_example_new}(k).
We note that in Figs.~\ref{fig:toy_example_new}(d, g, k, n), $\E[\tilde{x}_i]$ still differs from $x_i$ because of the blurring operation used in the reconstruction of $\tilde{x}_i$ from sampled 1D locations from $x_i$.
When the reconstruction process for $\tilde{x}_i$ is perfect (a topic of research beyond the scope of this work), such reconstruction errors would be eliminated.
For our experiments, we adopt this standard reconstruction process.

The uncertainty in $\tilde{x_i}$ due to measurement noise and incomplete sampling results in uncertainty in accurately estimating $d(x_i,\tilde{x}_i)$.
We now want to estimate the distribution $p[d(x_i,\tilde{x}_i)]$, where we model $d(x_i,\tilde{x}_i)$ as a Gaussian random variable.
We compute $\text{KLD}(x_i, \tilde{x}_i)$ for 1,000 random realizations of $\tilde{x}_i$ and estimate $\E[\text{KLD}(x_i, \tilde{x}_i)]$, $\text{Std}[\text{KLD}(x_i, \tilde{x}_i)]$.
These are reported in the titles of Figs.~\ref{fig:toy_example_new}(d, g, k, n).
We use $\text{KLD}$ as discrepancy function because of its wide adoption for saliency estimation, but the results presented here hold for other metrics as well.
We observe that:
\begin{itemize}
	\itemsep0em
	\item $\E[\text{KLD}(x_i,\tilde{x}_i)] > 0$, \ie $\text{KLD}(x_i,\tilde{x}_i)$ is biased. The source of the bias is twofold. First, $\text{KLD}(x_i,\tilde{x}_i) > 0$  because $\E[\tilde{x}_i]$ is a smoothed version of $x_i$ (bias due to the choice of the method used to reconstruct $\tilde{x}_i$), independently from the number of observers. Second, $\tilde{x}_i$ is noisy ($\text{Std}[\tilde{x}_i]>0$), which, especially for a limited number of observers, contributes with an additional bias to $\text{KLD}(x_i,\tilde{x}_i)$.
	\item $\text{Std}[\text{KLD}(x_i,\tilde{x}_i)] > 0$, and it tends to be smaller for a larger number of observers.
	\item For a given number of observers, $\E[\text{KLD}(x_i,\tilde{x}_i)]$ and $\text{Std}[\text{KLD}(x_i,\tilde{x}_i)]$ are larger for multimodal maps.
\end{itemize}

We conclude that, when $\tilde{x}_i$ is affected by measurement noise and incomplete sampling, the expected value and variance of the discrepancy $d(x_i,\tilde{x}_i)$ are not zero, depend on the number of observers, and are different for each frame.
These properties, which also hold for 2D saliency maps recorded from real human observers, form the basis for the development and interpretation of NAT.

\section{Gaze data analysis for ForGED (mentioned in Sec.~4)}

\begin{figure}[h!]
	\centering
	\includegraphics[width=0.45\textwidth,trim={0cm 0cm 0cm 0.5cm}, clip]{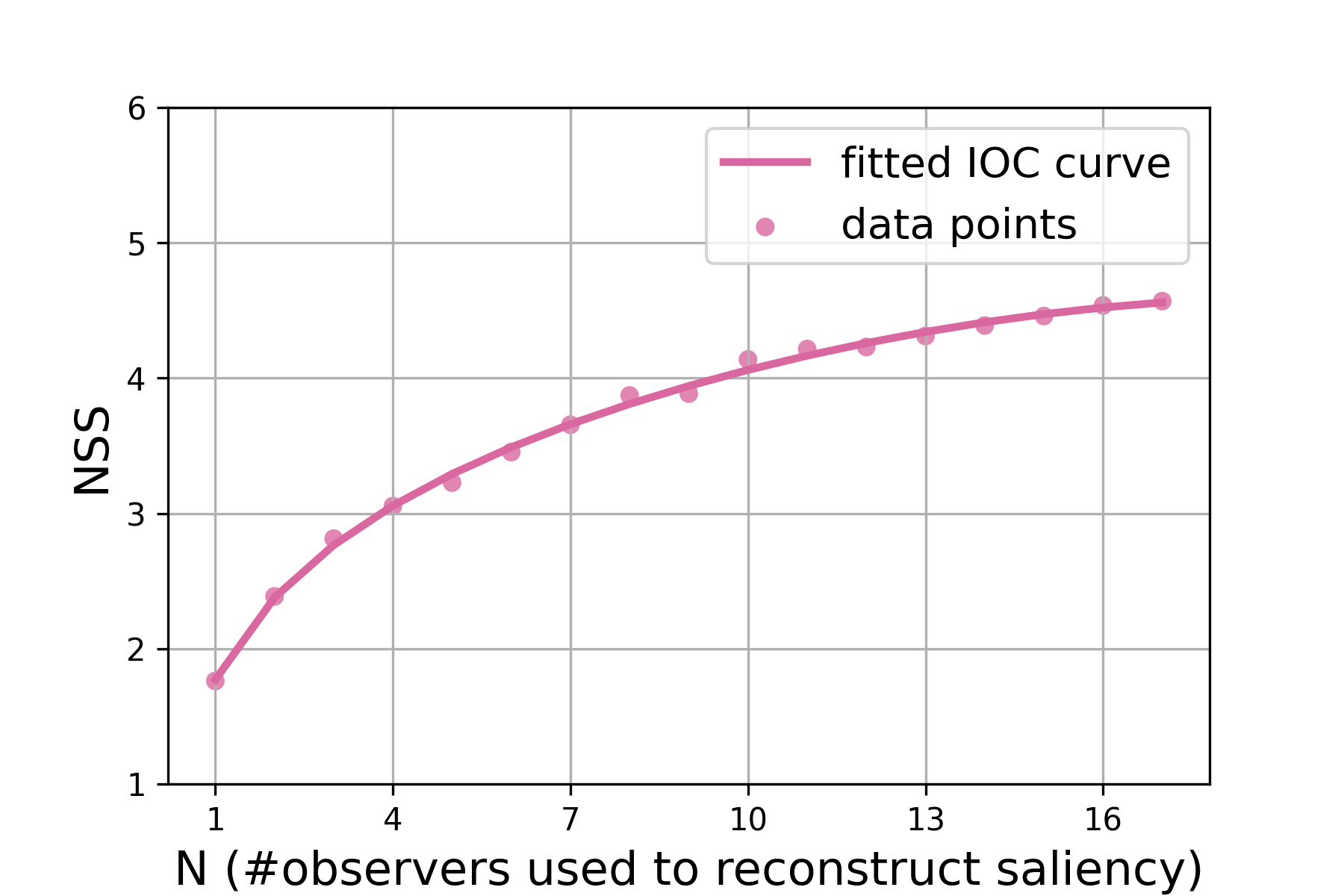}
	\caption{Inter-observer consistency (IOC) curve computed on test-set frames of ForGED containing at least $19$ observers. Each data point is an average of the IOC value for the given value of $N$, with the average computed over multiple realizations across the frames. The fitted curve is shown with a solid line and indicates the diminishing amount of new information that gaze data from additional observers imparts, when $N$ is sufficiently high.}
	\label{fig:forged-ioc-nss}
\end{figure}

\paragraph{Observer consistency and ForGED dataset split.} As discussed in Sec.~{1} of the main paper, IOC curve measures how well a saliency map reconstructed from gaze data of $N$ observers explains the gaze of a new observer as a function of $N$~\cite{Kum15, Wil11, Jia21}.
A converged IOC curve indicates that additional observers do not add significant new information to the reconstructed saliency map~\cite{Jia21,Jud12}.
A typical test of whether a dataset captures sufficient observers is to evaluate the level of convergence of the IOC curves \textit{on average across all frames} at maximum value for $N$ (sometimes by using curve-fitting and extrapolation~\cite{Jud12}). 
To obtain the average IOC for ForGED, we sample $1$ out of every $5$ frames from ForGED test videos containing at least $19$ observers -- for a total of $1500$ frames.
For each frame, we compute the per-frame IOC curve with $20$ random realizations for the subset of observers that constitute the $N$ observers and the subset that constitutes the new observer whose gaze data is to be explained by the $N$-observer saliency map. 
All realizations of the IOC curves across all sampled frames are averaged to obtain the IOC curve shown in Fig.~\ref{fig:forged-ioc-nss}.
As can be seen from Fig.~\ref{fig:forged-ioc-nss}, the gradient magnitude of the IOC curve is small at $N=17$ ($0.04$).
This further diminishes upon extrapolation to $N=21$ observers to $0.02$.
Our test set therefore contains gaze data from up to $21$ observers per video (median $17$).
As noted in the main paper (Limitations and Future Work in Sec.~6) - while on average the IOC curves across all evaluated datasets (LEDOV, DIEM, ForGED) show very small gradient at sufficiently high number of observers, the level of convergence for each frame may be different (content-dependent) and motivates the need for NAT. 
This also presents an interesting direction of future research to design noise-robust evaluation schemes.
Note that, while the ForGED test dataset contains gaze data from a large number of observers (that ensures small gradients in the IOC curves at maximum available $N$), the ForGED training dataset consists of a larger number of videos but with gaze data from only $5-15$ observers (the majority of the videos contain $5$ observers). 
This setting simulates the scenario where training data with limited number of observers is available (the setting most suitable for NAT) -- while the testing is always performed on more accurate saliency maps.
The training-validation-test split for ForGED videos is $379$ videos for training, $26$ for validation, and $75$ for testing.
As already discussed in the main paper (and also shown in Sec.~\ref{sec:sup-results}), for the different experiments enlisted in tables, the training dataset size is varied in terms of number of available training videos, $V$, and number of observers, $N$, used to reconstruct the saliency maps $\tilde{x_i}$ per video -- to demonstrate the performance gain of NAT for varying amount of training data.

\begin{figure}[h!]
	\centering
	\includegraphics[width=0.7\textwidth]{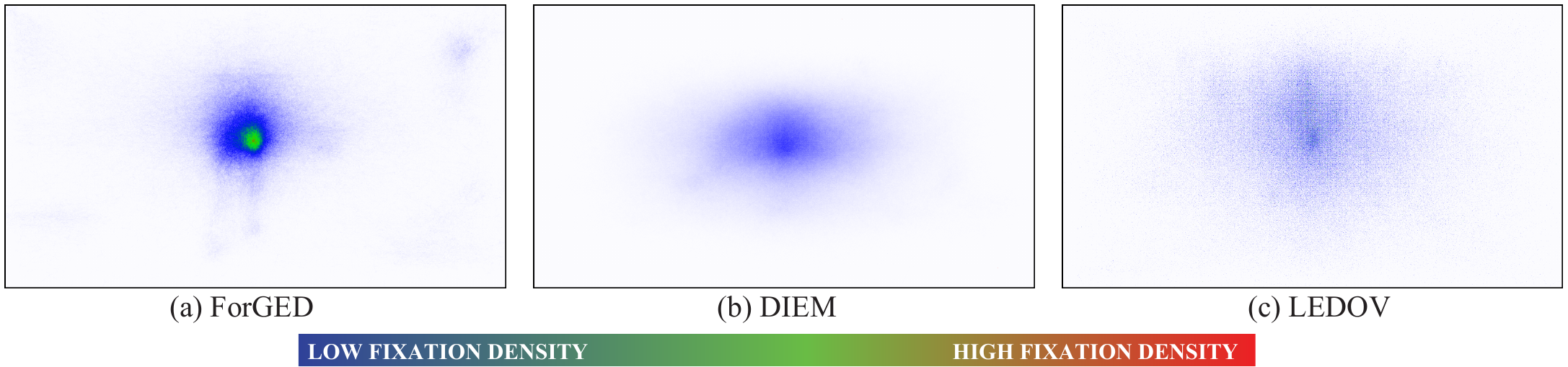}
	\caption{\addtext{Accumulated fixation density across gaze data from all observers across all frames in (a) ForGED (b) DIEM and (c) LEDOV.}}
	\label{fig:centerbias}
\end{figure}

\paragraph{Observer gaze behavior in ForGED.} Given that the main character is placed at the center of the screen in Fortnite game, we observe an affinity towards center in the gaze behavior. 
Events such as combat, focused motion towards a location such as the horizon, attempts to pick up resources such as ammunitions lead to observer gaze behavior that follows the narrative of the game-play (\eg viewers observe the opponent when the main character is in combat, viewers look towards the horizon when the main character is moving towards it).
On the other hand, when a scene becomes relatively uneventful, such as when the main central character has been running towards the horizon for a few seconds, the observers' gaze tends to become more exploratory -- scanning the surroundings, or simply browsing the evolving scenery. 
Examples of all such scenes can be found in the supplementary video, Fig.~3 of the main paper, and Fig.~\ref{fig:results} in this Supplementary document. 
\addtext{Lastly, we accumulate all of the gaze locations captured on ForGED into a fixation density map (Fig.~\ref{fig:centerbias}a) to assess the common viewing tendencies of observers. 
We also compare these for LEDOV and DIEM.
As discussed in Sec.~4 of the main paper, due to the main character near the center of the frame, the aiming reticle at the center of the frame, and a guiding mini-map on the top right, observers look at these regions frequently.
As compared to LEDOV and DIEM, such a behavior is uniquely representative of the observer behavior in third person shooting games such as Fortnite.
In case of LEDOV and DIEM, we also observe a bias towards the center -- but it tends to be more widespread as shown in Fig.~\ref{fig:centerbias}.}

\section{Analysis of approximation in Eq.~6 (mentioned in Sec.~3)}

To analyze the accuracy of Eq.~{6} in the main paper, we select a subset of the video frames from the DIEM dataset that contains gaze data from more than $200$ observers. 
Given the very large number of gaze fixations for these frames, we anticipate that the estimated human-saliency map $\tilde{x_i}$ is very close to ground-truth saliency $x_i$~\cite{Jud12} for every such frame $i$ (as also confirmed by converged IOC curves for these frames). 
We therefore analyze the accuracy of Eq.~{6} under the assumption that the $>200$-observer gaze maps of these frames represent $x_i$. 
From these $200$-observer gaze maps ($x_i$), we sample a certain number (denoted as $M$) of gaze fixation locations followed by blurring to compute $\tilde{x_i}$.
Therefore,  $\tilde{x}_i\ = SR(x_i; M)$.
Then, we compute $\tilde{\tilde{x}}$ by sampling $M$ spatial locations as per the pdf $\tilde{x}$ followed by blurring. 
That is, $\tilde{\tilde{x_i}}\ = SR(\tilde{x_i}; M)$.

Using multiple realizations of $\tilde{x}$ and $\tilde{\tilde{x}}$, we estimate $\mathbb{E}[d(x, \tilde{x})]$, $\mathbb{E}[d(\tilde{x}, \tilde{\tilde{x}})]$, $\text{Var}[d(x, \tilde{x})]$, $\text{Var}[d(\tilde{x}, \tilde{\tilde{x}})]$. 
We find that the mean absolute percentage error (MAPE) in the approximation of $\mathbb{E}[d(x, \tilde{x})]$ (Eq.~6 in main paper) goes from $21\%$ for $N=5$, to $13\%$ for $N=15$, and down to $10\%$ for $N=30$. 
Similarly, MAPE in the approximation of $\text{Var}[d(x, \tilde{x})]$ (Eq.~6 in main paper) goes from $13\%$ for $N=5$, to $6\%$ for $N=15$, and down to $5\%$ for $N=30$. 
Note that a large under/over-estimation of $\mathbb{E}[d(x,\tilde{x})]$ and $\text{Var}[d(x, \tilde{x})]$ in Eq.~{6} (main paper) may lead to overfitting to noisy data or sub-optimal convergence respectively using Eq.~{7} (main paper) for training. 
This would result in poor performance of NAT compared to traditional training -- which, as shown by the results, is not the case.

\section{Additional Results (mentioned in Sec.~5)}
\label{sec:sup-results}
We now report the additional experiments performed to compare NAT (Eq.~6 in main paper) to traditional training (abbreviated as TT in this section; Eq.~2 in main paper). Furthermore, we show typical gaze maps obtained through TT and NAT compared to the ground truth for TASED on the ForGED dataset in Fig.~\ref{fig:results}.

\subsection{Dataset type and size}
In this section, we continue reporting the results from Sec.~{5} of the main paper, where we compared NAT vs. TT for different dataset types and sizes. Table~\ref{tab:tased-kld-diem} compares the performance of TT to NAT on an additional dataset, the DIEM dataset~\cite{Mit11}, for the TASED architecture~\cite{Min19}, and using KLD as discrepancy for training.
As done throughout Sec.~{5} of the main paper, the evaluation on test set of DIEM is performed on videos with gaze data from \textit{all} of the available observers (in contrast to training, for which a subset of observers are used, see Table~{1} in main paper). 
In case of DIEM dataset, given that only $84$ videos are available, we use $30$ or $60$ videos for training and report the results on the remaining $24$ videos, which are also used as validation set. 
The number of observers for these videos ranges from $51$ to $219$, which makes DIEM a very low-noise evaluation set~\cite{Jud12}. 
Results on DIEM are consistent with those reported in the main paper, with NAT providing better metrics in evaluation when compared to TT when less training data (e.g., $30$ videos) is available.

\begin{table}[h!]
	\centering
	\resizebox{0.6\columnwidth}{!}{
		\renewcommand{\arraystretch}{1.05}
\scriptsize
\centering
\begin{tabular}[b]{c|c|c|c|c|c|c|c}
\textbf{train videos $V$} & \textbf{train obs. $N$} & \textbf{loss} & \textbf{KLD}$\downarrow$&\textbf{CC}$\uparrow$ &\textbf{SIM}$\uparrow$ & \textbf{NSS}$\uparrow$ &\textbf{AUC-J}$\uparrow$ \\	
\hline
\multirow{6}{*}{$30$} & \multirow{2}{*}{$5$} & TT &$0.641$& $0.698$& $0.591$& $\mathbf{3.517}$& $0.922$\\
& & NAT &$\mathbf{0.599}$& $\mathbf{0.708}$& $\mathbf{0.592}$& $3.513$& $\mathbf{0.934}$\\
\cline{2-8}
& \multirow{2}{*}{$15$} & TT &$0.597$& $0.710$& $0.602$& $3.582$& $0.930$\\
& & NAT &$\mathbf{0.583}$& $\mathbf{0.718}$& $\mathbf{0.607}$& $\mathbf{3.627}$& $\mathbf{0.932}$\\
\cline{2-8}
& \multirow{2}{*}{$31$} & TT &$0.576$& $0.724$& $0.614$& $3.663$& $0.925$\\
& & NAT &$\mathbf{0.559}$& $\mathbf{0.731}$& $\mathbf{0.618}$& $\mathbf{3.694}$& $\mathbf{0.928}$\\
\hline
\multirow{6}{*}{$60$} & \multirow{2}{*}{$5$} & TT &$0.528$& $0.735$& $\mathbf{0.619}$& $\mathbf{3.709}$& $0.933$\\
& & NAT &$\mathbf{0.518}$& $\mathbf{0.737}$& $0.616$& $3.639$& $\mathbf{0.940}$\\
\cline{2-8}
& \multirow{2}{*}{$15$} & TT &$\mathbf{0.485}$& $\mathbf{0.757}$& $\mathbf{0.639}$& $\mathbf{3.795}$& $0.933$\\
& & NAT &$0.493$& $0.754$& $0.635$& $3.792$& $\mathbf{0.936}$\\
\cline{2-8}
& \multirow{2}{*}{$31$} & TT &$0.476$& $0.759$& $0.641$& $3.821$& $\mathbf{0.938}$\\
& & NAT &$\mathbf{0.467}$& $\mathbf{0.766}$& $\mathbf{0.654}$& $\mathbf{3.864}$& $0.935$\\
\end{tabular}
}
	\caption{Saliency metrics on DIEM, for TASED Net, training with KLD as discrepancy, and various number of training videos and observers. The best metrics between TT (Eq.~{2} in main paper) and NAT are in bold.}		
	\label{tab:tased-kld-diem}
\end{table}

\begin{figure*}[h!]
	\centering
	\begingroup
	\setlength{\tabcolsep}{0pt} 
	\renewcommand{\arraystretch}{0.1}
	\begin{tabular}{ccccc}
		\rotatebox{90}{\scriptsize{\hspace{2em}measured $\tilde{x}_i$}}  &
		\includegraphics[width=0.22\textwidth,trim={1.5cm 0.5cm 1.5cm 0.5cm},clip]{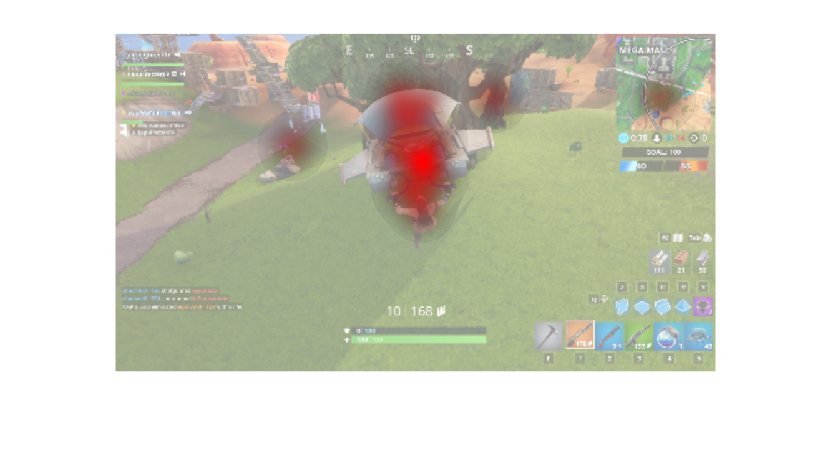}  &
		\includegraphics[width=0.22\textwidth,trim={1.5cm 0.5cm 1.5cm 0.5cm},clip]{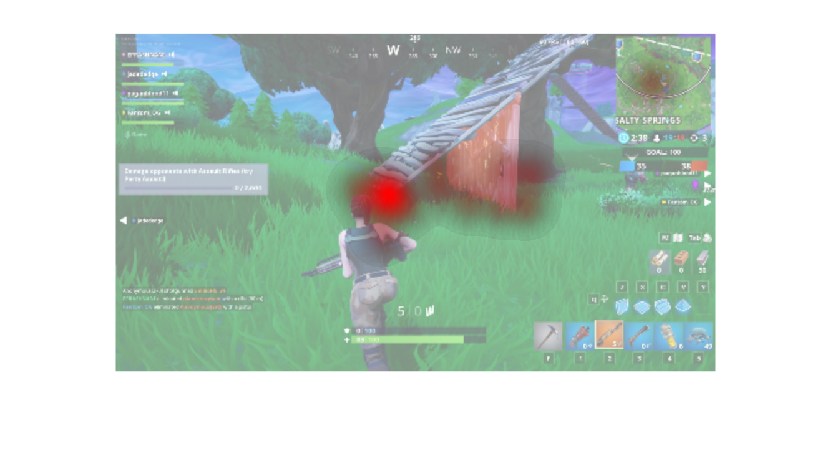}   &
		\includegraphics[width=0.22\textwidth,trim={1.5cm 0.5cm 1.5cm 0.5cm},clip]{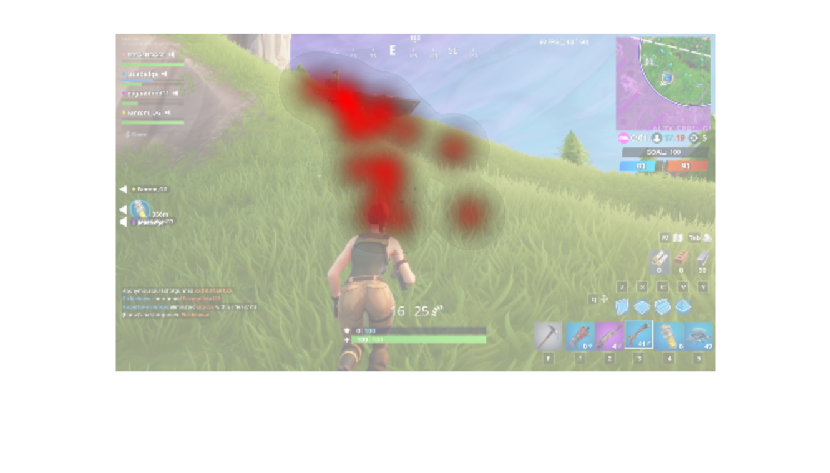}   &
		\includegraphics[width=0.22\textwidth,trim={1.5cm 0.5cm 1.5cm 0.5cm},clip]{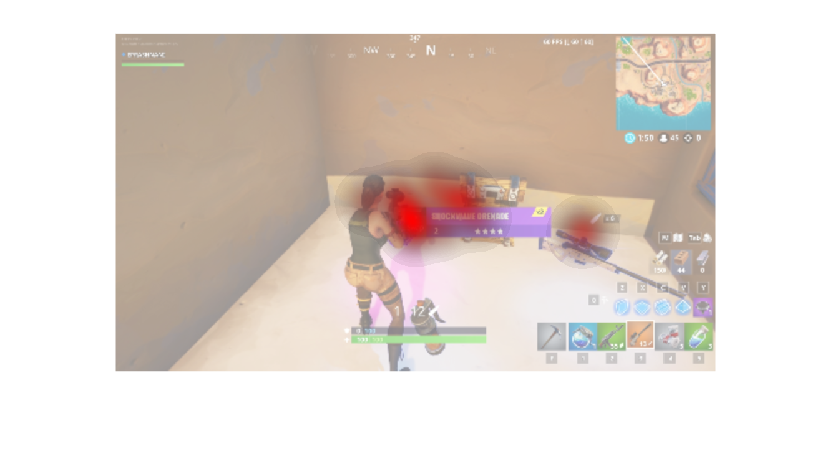}                                                                                   \\
		& \scriptsize{KLD=1.48, CC=0.52} & \scriptsize{KLD=0.88, CC=0.78} & \scriptsize{KLD=1.40, CC=0.47} & \scriptsize{KLD=1.22, CC=0.52}\\		                                                               

		\rotatebox{90}{\hspace{1.5em}\scriptsize{TT}}&
		\includegraphics[width=0.22\textwidth,trim={1.5cm 0.5cm 1.5cm 0.5cm},clip]{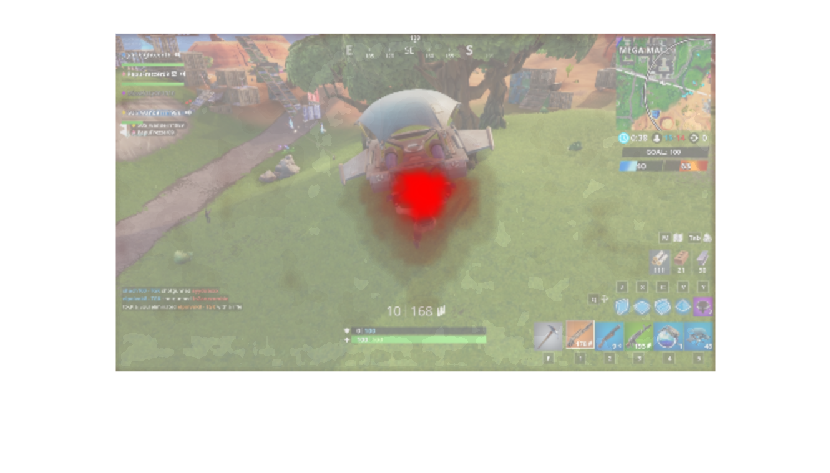} &
		\includegraphics[width=0.22\textwidth,trim={1.5cm 0.5cm 1.5cm 0.5cm},clip]{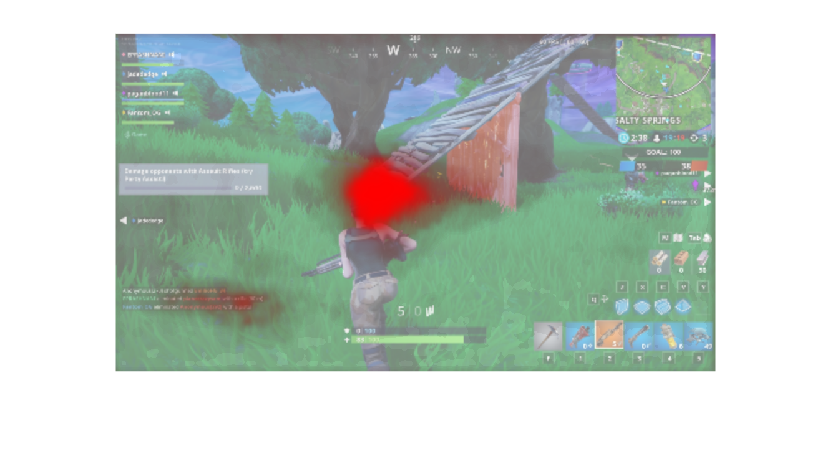}  &
		\includegraphics[width=0.22\textwidth,trim={1.5cm 0.5cm 1.5cm 0.5cm},clip]{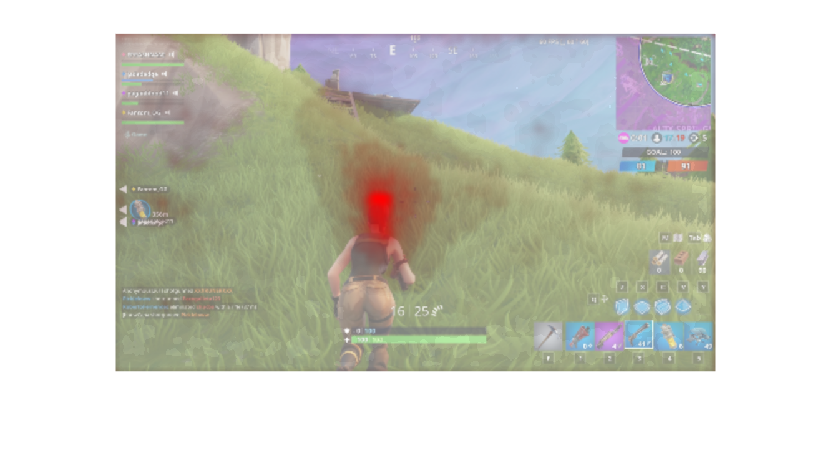}  &
		\includegraphics[width=0.22\textwidth,trim={1.5cm 0.5cm 1.5cm 0.5cm},clip]{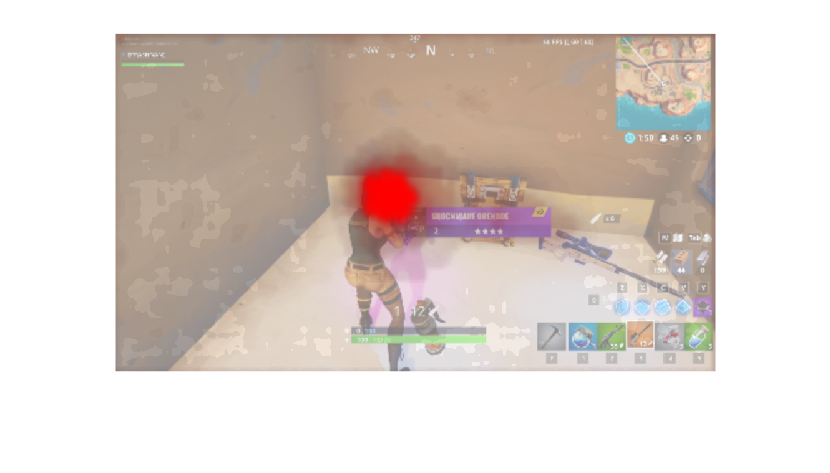}\\
		&\scriptsize{KLD=0.82, CC=0.77} & \scriptsize{KLD=0.59, CC=0.69} & \scriptsize{KLD=0.54, CC=0.75} & \scriptsize{KLD=1.28, CC=0.51}\\	
		\rotatebox{90}{\vspace{3em}\scriptsize{\hspace{3em}NAT}}&
		\includegraphics[width=0.22\textwidth,trim={1.5cm 0.5cm 1.5cm 0.5cm},clip]{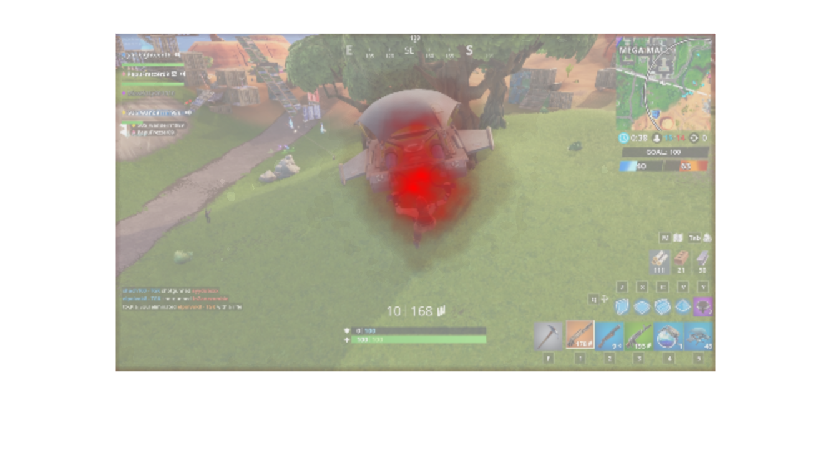} &
		\includegraphics[width=0.22\textwidth,trim={1.5cm 0.5cm 1.5cm 0.5cm},clip]{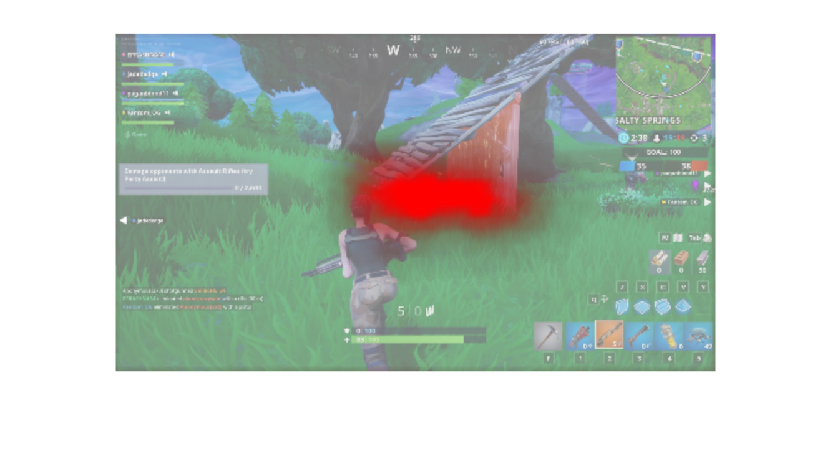}  &
		\includegraphics[width=0.22\textwidth,trim={1.5cm 0.5cm 1.5cm 0.5cm},clip]{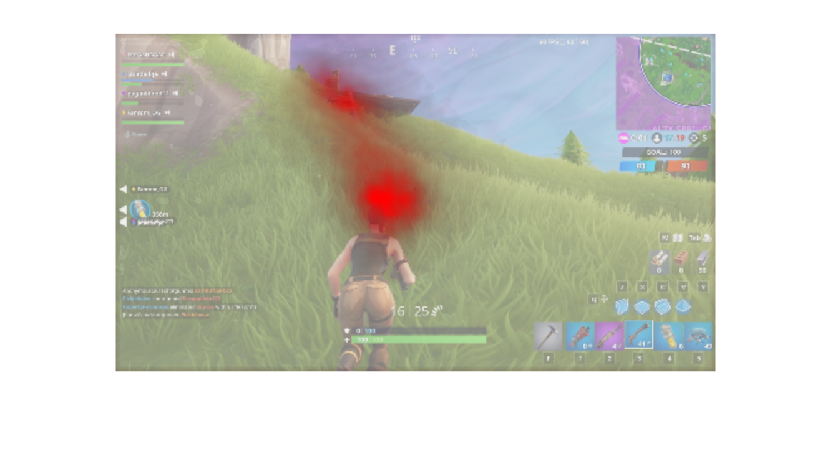}  &
		\includegraphics[width=0.22\textwidth,trim={1.5cm 0.5cm 1.5cm 0.5cm},clip]{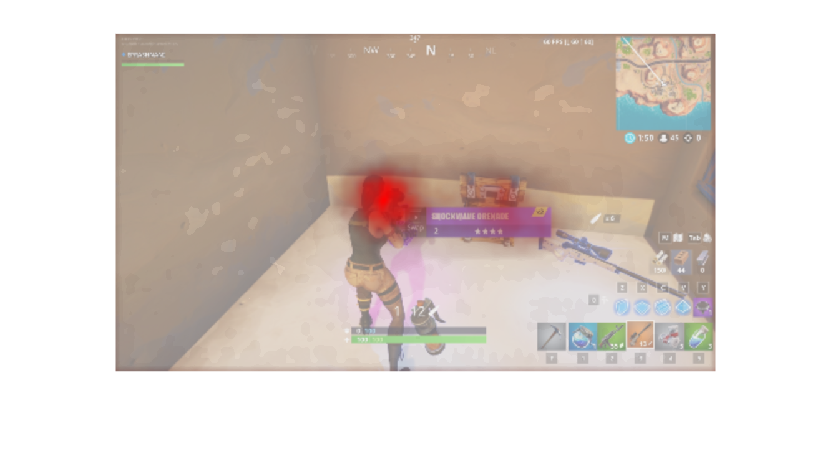}\\

	\end{tabular}
	\endgroup
	\caption{Typical gaze maps obtained through TT (second row -- Eq.~{2} in main paper) and NAT (third row) compared to the ground truth (first row) for TASED on the ForGED dataset, training with KLD loss, 30 training videos and 5 observers per frame (see Table~4a in main paper). Each panel reports in the title the corresponding KLD and CC values. The last column shows a failure case where the metrics KLD and CC indicate that NAT is worse than TT, although a visual inspection might indicate otherwise. Furthermore, the saliency maps predicted with TT indicate more centralized unimodal predictions -- while NAT accurately predicts decentralized, multi-modal saliency maps even when trained with less data.
	The visualization of saliency map overlays follows the scheme in Fig.~3 of main paper.\small{\textit{ ForGED images have been published with permission of Epic Games.}}}
	\label{fig:results}
\end{figure*}

\subsection{Discrepancy functions}
Table~\ref{tab:tased-nss-fortnite} shows  NAT vs. TT using $d=-\text{NSS}$ on ForGED dataset.
In Table~\ref{tab:tased-nss-fortnite}, we notice that NAT overcomes TT in terms of NSS only for 2 or 5 observers, and 30 training videos.
Recall that, by design, NSS optimizes the predicted saliency map only at the measured fixation locations. Consequently, when \textit{few} fixations per frame are available for training, a high NSS score may not generalize well to other evaluation metrics that evaluate different aspects of the quality of a predicted saliency map. 
This can be alleviated by additional regularization (such as using additional metrics as we do with  $d = \text{KLD}-0.1\text{CC}-0.1\text{NSS}$ in Table~4b of the main paper and observe that high NSS scores generalize to good performance in terms of other metrics).
In other words, for few-observer training, optimizing for NSS alone may not constrain the predicted saliency map sufficiently --- which shows up as poor generalization to other metrics.
This is what we observe in Table~\ref{tab:tased-nss-fortnite}, where the regularizing effect of NAT leads to worse NSS values compared to TT; but, \textit{all} of the other evaluation metrics indicate NAT to be better. 

\begin{table}[h!]
	\centering
	\resizebox{0.6\columnwidth}{!}{
		\renewcommand{\arraystretch}{1.05}
\scriptsize
\centering
\begin{tabular}[b]{c|c|c|c|c|c|c|c}
\textbf{train videos $V$} & \textbf{train obs. $N$} & \textbf{loss} & \textbf{KLD}$\downarrow$&\textbf{CC}$\uparrow$ &\textbf{SIM}$\uparrow$ & \textbf{NSS}$\uparrow$ &\textbf{AUC-J}$\uparrow$ \\	
\hline
\multirow{6}{*}{$30$} & \multirow{2}{*}{$2$} & TT &$2.005$& $0.362$& $0.302$& $2.677$& $0.788$\\
& & NAT &$\mathbf{1.408}$& $\mathbf{0.528}$& $\mathbf{0.371}$& $\mathbf{3.163}$& $\mathbf{0.887}$\\
\cline{2-8}
& \multirow{2}{*}{$5$} & TT &$1.642$& $0.489$& $0.28$& $3.212$& $0.898$\\
& & NAT &$\mathbf{1.254}$& $\mathbf{0.566}$& $\mathbf{0.417}$& $\mathbf{3.39}$& $\mathbf{0.906}$\\
\cline{2-8}
& \multirow{2}{*}{$15$} & TT &$1.518$& $0.506$& $0.403$& $\mathbf{3.672}$& $0.826$\\
& & NAT &$\mathbf{1.155}$& $\mathbf{0.608}$& $\mathbf{0.435}$& $3.552$& $\mathbf{0.91}$\\
\hline
\multirow{4}{*}{$100$} & \multirow{2}{*}{$2$} & TT &$1.328$& $0.563$& $0.383$& $\mathbf{3.783}$& $0.899$\\
& & NAT &$\mathbf{1.206}$& $\mathbf{0.584}$& $\mathbf{0.426}$& $3.44$& $\mathbf{0.906}$\\
\cline{2-8}
& \multirow{2}{*}{$5$} & TT &$1.312$& $0.578$& $0.452$& $\mathbf{3.983}$& $0.835$\\
& & NAT &$\mathbf{1.165}$& $\mathbf{0.61}$& $\mathbf{0.475}$& $3.747$& $\mathbf{0.879}$\\
\hline
\multirow{4}{*}{$379$} & \multirow{2}{*}{$2$} & TT &$1.163$& $0.614$& $0.475$& $\mathbf{4.161}$& $0.857$\\
& & NAT &$\mathbf{1.028}$& $\mathbf{0.642}$& $\mathbf{0.495}$& $3.858$& $\mathbf{0.901}$\\
\cline{2-8}
& \multirow{2}{*}{$5$} & TT &$1.093$& $0.633$& $0.491$& $\mathbf{4.381}$& $0.875$\\
& & NAT &$\mathbf{1.006}$& $\mathbf{0.658}$& $\mathbf{0.512}$& $3.928$& $\mathbf{0.892}$\\
\end{tabular}
}
	\caption{NAT vs. TT on ForGED for TASED, $d = -\text{NSS}$ (a fixation-based discrepancy), various number of training videos and observers. Best metrics for each pair of experiments in bold.}
	\label{tab:tased-nss-fortnite}
\end{table}

To further verify that NAT generalizes to different discrepancy functions, we train and test TASED on LEDOV~\cite{Jia18} with the fixation-based discrepancy function, $d=-\text{NSS}$, and the combination of fixation and density-based discrepancy functions, $d=\text{KLD}-0.1\text{CC}-0.1\text{NSS}$ (which is a popular discrepancy function used in video-saliency research~\cite{Dro20,Wan18}). 
The test set for LEDOV is used for all reported evaluations on LEDOV dataset, which contains gaze data from $32$ observers per video.

Table~\ref{tab:tased-nss-ledov} shows  NAT vs. TT (Eq.~{2} in main paper) using $d=-\text{NSS}$. 
For this specific experiment, with TT we observe that adopting RMSprop as the optimizer (as done for all experiments in the paper) shows very fast convergence to very high NSS values. 
While this property of fast and optimal convergence of discrepancy function has proven useful for all experiments in the paper (see Sec.~\ref{sec:hyperparam} for details), for this specific experiment the solution provided by RMSprop optimization shows poor generalization to all other saliency metrics.
This behavior is alleviated to some extent by switching RMSProp with Stochastic Gradient Descent (SGD) for TT -- but at the cost of poor convergence in terms of NSS. 
To show this, in Table~\ref{tab:tased-nss-ledov}, we report two sets of experiments for TT for each size of training dataset (one with SGD and another with RMSprop).
With NAT, however, we observe a consistent optimal convergence due to the regularizing effect of the NAT formulation that prevents overfitting to dataset noise.

We further observe that using additional terms with NSS in the discrepancy function, such as with $d=\text{KLD}-0.1\text{CC}-0.1\text{NSS}$ overcomes some of the issues of training with NSS alone. 
Table~\ref{tab:tased-kldccnss-ledov},~\ref{tab:tased-kldccnss-diem} show the comparison of TT vs. NAT for this combined discrepancy function. 
A high NSS performance in this case is well-correlated with good performance in terms of other metrics.
Furthermore we note that the performance of NAT is superior to TT when less gaze data is available, with the gap between the two approaches closing in with more gaze data. 
Given our analyses of all of the experiments with various discrepancy functions and dataset types, our conclusion is that the performance of models trained with density-based discrepancy functions (e.g., KLD) is better for TT as well as NAT, with NAT showing consistent superior performance compared to TT.

\begin{table}[h!]
	\centering
	\resizebox{0.6\columnwidth}{!}{
		\renewcommand{\arraystretch}{1.05}
\scriptsize
\centering
\begin{tabular}[b]{c|c|c|c|c|c|c|c}
\textbf{train videos $V$} & \textbf{train obs. $N$} & \textbf{loss} & \textbf{KLD}$\downarrow$&\textbf{CC}$\uparrow$ &\textbf{SIM}$\uparrow$ & \textbf{NSS}$\uparrow$ &\textbf{AUC-J}$\uparrow$ \\	

\hline
\multirow{6}{*}{$100$} & \multirow{3}{*}{$5$} & TT, SGD &$\mathit{2.352}$& $\mathit{0.244}$& $\mathbf{0.267}$& $2.272$& $\mathit{0.761}$\\
& & TT, RMSprop &$4.139$& $0.192$& $0.056$& $\mathbf{9.92}$& $0.178$\\
& & NAT &$\mathbf{1.746}$& $\mathbf{0.428}$& $\mathit{0.230}$& $\mathit{2.358}$& $\mathbf{0.916}$\\

\cline{2-8}
& \multirow{3}{*}{$30$} & TT, SGD &$\mathit{2.302}$& $\mathit{0.258}$& $\mathbf{0.275}$& $\mathit{2.661}$& $\mathit{0.775}$\\
& & TT, RMSprop &$3.593$& $0.247$& $0.111$& $\mathbf{13.628}$& $0.423$\\
& & NAT &$\mathbf{1.903}$& $\mathbf{0.398}$& $\mathit{0.198}$& $2.370$& $\mathbf{0.919}$\\

\hline
\multirow{6}{*}{$461$} & \multirow{3}{*}{$5$} & TT, SGD &$\mathit{2.777}$&$	\mathit{0.317}$&$	\mathit{0.232}$&$	\mathit{4.464}$&$	\mathit{0.612}$\\
& & TT, RMSprop &$4.00$&$	0.241$&$	0.062$&$	\mathbf{14.617}$&$	0.206$\\
& & NAT &$\mathbf{1.305}$&$	\mathbf{0.575}$&$	\mathbf{0.354}$&$	3.29$&$	\mathbf{0.929}$\\

\cline{2-8}
& \multirow{3}{*}{$30$} & TT, SGD &$\mathit{2.252}$&$	\mathit{0.470}$&$	\mathbf{0.355}$&$	2.463$&$	\mathit{0.593}$\\
& & TT, RMSprop &$3.526$&$	0.292$&$	0.127$&$	\mathbf{14.048}$&$	0.381$\\
& & NAT &$\mathbf{1.402}$&$	\mathbf{0.571}$&$	\mathit{0.310}$&$	\mathit{2.933}$&$	\mathbf{0.927}$\\

\end{tabular}
}
	\caption{Comparison of TT (Eq.~{2} in main paper) vs. NAT on LEDOV testing set, for TASED Net, trained with $-0.1\text{NSS}$ as discrepancy, and various number of training videos and observers. The best metric between each set of 3 experiments for a given dataset size (videos and observers) is in bold and the second-best is italicized. Given the strong overfitting behavior of NSS with TT using RMSprop for this particular set of experiments, we report TT optimized with SGD as well.}%
	\label{tab:tased-nss-ledov}
\end{table}

\begin{table}[h!]
	\centering
	\resizebox{0.6\columnwidth}{!}{
		\renewcommand{\arraystretch}{1.05}
\scriptsize
\centering
\begin{tabular}[b]{c|c|c|c|c|c|c|c}
\textbf{train videos $V$} & \textbf{train obs. $N$} & \textbf{loss} & \textbf{KLD}$\downarrow$&\textbf{CC}$\uparrow$ &\textbf{SIM}$\uparrow$ & \textbf{NSS}$\uparrow$ &\textbf{AUC-J}$\uparrow$ \\	
\hline
\multirow{2}{*}{$30$} & \multirow{2}{*}{$30$} & TT &$1.652$& $0.446$& $0.261$& $2.269$& $0.871$\\
& & NAT &$\mathbf{1.243}$& $\mathbf{0.494}$& $\mathbf{0.394}$& $\mathbf{2.491}$& $\mathbf{0.900}$\\
\hline
\multirow{4}{*}{$100$} & \multirow{2}{*}{$5$} & TT &$1.368$& $0.496$& $0.395$& $2.430$& $0.863$\\
& & NAT &$\mathbf{1.149}$& $\mathbf{0.540}$& $\mathbf{0.423}$& $\mathbf{2.782}$& $\mathbf{0.905}$\\
\cline{2-8}
& \multirow{2}{*}{$30$} & TT &$1.261$& $0.534$& $0.368$& $2.658$& $0.903$\\
& & NAT &$\mathbf{1.034}$& $\mathbf{0.574}$& $\mathbf{0.432}$& $\mathbf{3.250}$& $\mathbf{0.928}$\\
\hline
\multirow{4}{*}{$461$} & \multirow{2}{*}{$5$} & TT &$1.159$& $0.577$& $0.485$& $\mathbf{3.912}$& $0.864$\\
& & NAT &$\mathbf{0.852}$& $\mathbf{0.626}$& $\mathbf{0.513}$& $3.451$& $\mathbf{0.931}$\\
\cline{2-8}
& \multirow{2}{*}{$30$} & TT &$0.913$& $0.626$& $0.513$& $\mathbf{5.743}$& $0.910$\\
& & NAT &$\mathbf{0.755}$& $\mathbf{0.688}$& $\mathbf{0.554}$& $3.559$& $\mathbf{0.930}$\\
\end{tabular}
	}
	\caption{Saliency quality metrics on LEDOV testing set, for TASED Net, training with KLD-0.1CC-0.1NSS as discrepancy, and various number of training videos and observers. The best metrics between TT (Eq.~{2} in main paper) and NAT are in bold.}
	\label{tab:tased-kldccnss-ledov}
\end{table}
\begin{table}[h!]
	\centering
	\resizebox{0.6\columnwidth}{!}{
		\renewcommand{\arraystretch}{1.05}
\scriptsize
\centering
\begin{tabular}[b]{c|c|c|c|c|c|c|c}
\textbf{train videos $V$} & \textbf{train obs. $N$} & \textbf{loss} & \textbf{KLD}$\downarrow$&\textbf{CC}$\uparrow$ &\textbf{SIM}$\uparrow$ & \textbf{NSS}$\uparrow$ &\textbf{AUC-J}$\uparrow$ \\	
\hline
\multirow{4}{*}{$30$}& \multirow{2}{*}{$15$} & TT &$0.687$& $0.696$& $0.590$& $3.618$& $0.900$\\
& & NAT &$\mathbf{0.588}$& $\mathbf{0.718}$& $\mathbf{0.601}$& $\mathbf{3.629}$& $\mathbf{0.932}$\\
\cline{2-8}
& \multirow{2}{*}{$31$} & TT &$\mathbf{0.555}$& $0.727$& $0.609$& $3.605$& $\mathbf{0.935}$\\
& & NAT &$0.560$& $\mathbf{0.730}$& $\mathbf{0.612}$& $\mathbf{3.666}$& $0.933$\\
\hline
\multirow{6}{*}{$60$} & \multirow{2}{*}{$5$} & TT &$0.555$& $0.728$& $\mathbf{0.615}$& $3.660$& $0.930$\\
& & NAT &$\mathbf{0.535}$& $\mathbf{0.736}$& $0.612$& $\mathbf{3.681}$& $\mathbf{0.937}$\\
\cline{2-8}
& \multirow{2}{*}{$15$} & TT &$0.514$& $0.743$& $0.631$& $3.804$& $0.931$\\
& & NAT &$\mathbf{0.488}$& $\mathbf{0.755}$& $\mathbf{0.636}$& $\mathbf{3.814}$& $\mathbf{0.939}$\\
\cline{2-8}
& \multirow{2}{*}{$31$} & TT &$\mathbf{0.502}$& $0.748$& $\mathbf{0.639}$& $\mathbf{3.887}$& $0.931$\\
& & NAT &$0.503$& $\mathbf{0.750}$& $0.632$& $3.774$& $\mathbf{0.938}$\\
\end{tabular}}
	\caption{Saliency quality metrics on DIEM testing set, for TASED Net, training with KLD-0.1CC-0.1NSS as discrepancy, and various number of training videos and observers. The best metrics between TT (Eq.~{2} in main paper) and NAT are in bold.}
	\label{tab:tased-kldccnss-diem}
\end{table}

\subsection{DNN architectures}
To further verify that NAT works effectively on different DNN architectures, independently from the adopted dataset, we train SalEMA~\cite{Lin19} on the ForGED dataset.
We use KLD as the discrepancy function, with RMSprop as the optimizer with a learning rate equal to $1e^{-5}$ rather than Adam with learning rate $1e^{-7}$ and binary cross entropy as discrepancy function, as suggested by the authors (an analysis of this hyperparameter choice is discussed later).
Consistently with the other cases analyzed here, NAT outperforms TT, notably when the number of observers or videos is limited (Table~\ref{tab:salema-kld-fortnite}).
\begin{table}[h!]
	\centering
	\resizebox{0.6\columnwidth}{!}{
		\scriptsize
\centering
\begin{tabular}[b]{c|c|c|c|c|c|c|c}
\textbf{train videos $V$} & \textbf{train obs. $N$} & \textbf{loss} & \textbf{KLD}$\downarrow$&\textbf{CC}$\uparrow$ &\textbf{SIM}$\uparrow$ & \textbf{NSS}$\uparrow$ &\textbf{AUC-J}$\uparrow$ \\	
\hline
\multirow{4}{*}{$30$} & \multirow{2}{*}{$5$} & TT &$1.229$& $0.546$& $0.412$& $2.911$& $0.912$\\
& & NAT &$\mathbf{1.187}$& $\mathbf{0.559}$& $\mathbf{0.428}$& $\mathbf{3.050}$& $\mathbf{0.915}$\\
\cline{2-8}
& \multirow{2}{*}{$15$} & TT &$1.214$& $0.544$& $0.420$& $2.972$& $0.916$\\
& & NAT &$\mathbf{1.184}$& $\mathbf{0.563}$& $\mathbf{0.426}$& $\mathbf{3.152}$& $\mathbf{0.916}$\\
\hline
\multirow{2}{*}{$100$} & \multirow{2}{*}{$5$} & TT &$1.077$& $\mathbf{0.600}$& $0.444$& $3.273$& $0.923$\\
& & NAT &$\mathbf{1.071}$& $0.599$& $\mathbf{0.447}$& $\mathbf{3.274}$& $\mathbf{0.926}$\\
\hline
\multirow{4}{*}{$379$} & \multirow{2}{*}{$2$} & TT &$\mathbf{1.054}$& $\mathbf{0.601}$& $\mathbf{0.447}$& $3.248$& $0.926$\\
& & NAT &$1.076$& $0.600$& $0.440$& $\mathbf{3.284}$& $\mathbf{0.930}$\\
\cline{2-8}
& \multirow{2}{*}{$5$} & TT &$\mathbf{1.014}$& $0.623$& $\mathbf{0.482}$& $\mathbf{3.533}$& $0.929$\\
& & NAT &$1.019$& $\mathbf{0.623}$& $0.471$& $3.526$& $\mathbf{0.930}$\\
\end{tabular}
}
	\caption{Saliency quality metrics on ForGED testing set, for SalEMA, training with KLD as discrepancy, and various number of training videos and observers. The best metrics between TT (Eq.~{2} in main paper) and NAT are in bold.}
	\label{tab:salema-kld-fortnite}
\end{table}

\section{Additional training details (mentioned in Sec.~5)}

\label{sec:hyperparam}
\begin{figure}[h!]
	\centering
	\includegraphics[width=0.45\textwidth,trim={0cm 0cm 0cm 0.5cm}, clip]{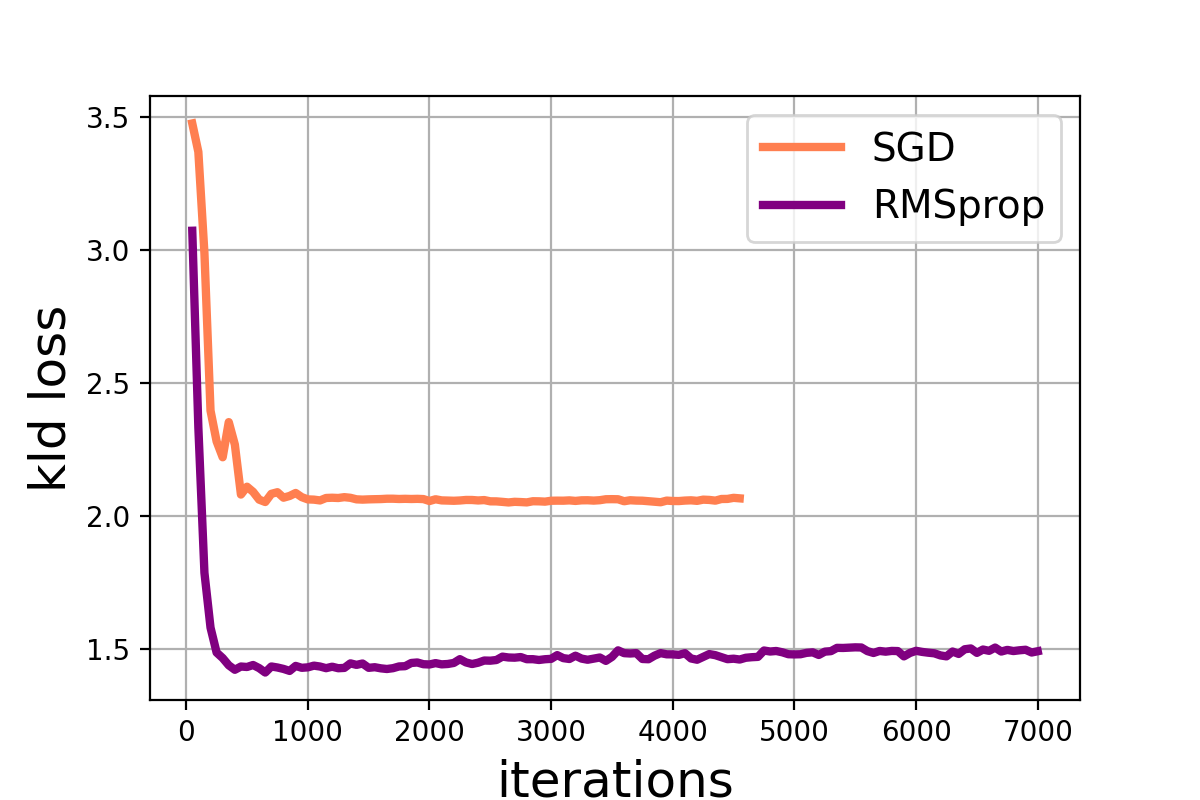}
	\caption{Validation-set performance plots (KLD vs. training iterations) for the LEDOV dataset during training of TASED with KLD as loss function and LEDOV dataset using: SGD, initial learning rate $0.1$ for decoder and $0.001$ for encoder, momentum $0.9$ (default used by authors); and RMSprop, learning rate $0.001$. Based on this experiment, we choose RMSprop with a learning rate of $0.001$ for our experiments.}
	\label{fig:tased-hyperparam}
\end{figure}

\paragraph{Hyperparameters for TASED training on LEDOV.} To ensure a fair comparison against traditional training and guarantee that the best performance are achieved for the given architecture and dataset, we first perform some hyperparameter tuning of TASED on LEDOV with traditional training (Eq.~2 in main paper).
We found that using RMSprop with a learning rate of $0.001$ for KLD optimization gives better performance than the default settings originally proposed for training on DHF1K (\ie, SGD with momentum $0.9$ and learning rate $0.1$ for decoder stepping down by a factor of $0.1$ at iteration $750$ and $950$, and $0.001$ for encoder), as shown in Table~\ref{tab:tased-hyperparam} and in Fig.~\ref{fig:tased-hyperparam}.
Thus, we adopt RMSprop with a learning rate of $0.001$ to train TASED for both traditional training and NAT in all the experiments. 
An exception to this rule is the traditional training with SGD reported in Table~\ref{tab:tased-nss-ledov}, where we adopt SGD with a learning rate of $0.0001$ (any higher leads to training instabilities due to data noise) and momentum $0.9$.

\begin{table}[h!]
	\centering
	\resizebox{0.6\columnwidth}{!}{
		\renewcommand{\arraystretch}{1.05}
\scriptsize
\centering
\begin{tabular}[b]{c|c|c|c|c|c}
\textbf{hyperparameter settings} & \textbf{KLD}$\downarrow$ &\textbf{CC}$\uparrow$ &\textbf{SIM}$\uparrow$ & \textbf{NSS}$\uparrow$ &\textbf{AUC-J}$\uparrow$ \\
\hline
TASED-Net, SGD, learning rate schedule (default)& $1.104$& $0.554$& $0.452$& $2.536$& $0.828$\\
TASED-Net, RMSprop, $0.001$, KLD (improved)&$\mathbf{0.754}$& $\mathbf{0.724}$& $\mathbf{0.572}$& $\mathbf{4.227}$& $\mathbf{0.921}$\\
\end{tabular}
}
	\caption{Performance on LEDOV for TASED trained traditionally using KLD with original settings, and those used in the main paper (RMSprop, learning rate $0.001$) on the full LEDOV training set. We adopted the best hyperparameter setting (best metrics in bold) for all experiments. *Original settings: SGD, initial learning rate $0.1$ for decoder and $0.001$ for encoder, momentum $0.9$.}
	\label{tab:tased-hyperparam}
\end{table}

\paragraph{Hyperparameters for SalEMA training on LEDOV.}
We train SalEMA~\cite{Lin19} on the full LEDOV dataset with the default choice for loss function and optimizer (Adam optimizer, binary cross entropy, with learning rate $1e^{-7}$), and compare against the adoption of the RMSprop optimizer with KLD as the loss function and $2$ learnings rates: $1e^{-5}$ and $1e^{-7}$ (see Table.~\ref{tab:salema-hyperparam}). 
We train with LEDOV training set and we choose the best hyperparameter setting based on the LEDOV test-set performance for all of the experiments in the paper.

\begin{table}[h!]
	\centering
	\resizebox{0.6\columnwidth}{!}{
		\renewcommand{\arraystretch}{1.05}
\scriptsize
\centering
\begin{tabular}[b]{c|c|c|c|c|c}
\textbf{hyperparameter settings} & \textbf{KLD}$\downarrow$ &\textbf{CC}$\uparrow$ &\textbf{SIM}$\uparrow$ & \textbf{NSS}$\uparrow$ &\textbf{AUC-J}$\uparrow$ \\
\hline
Adam, $1e^{-7}$, BCE (original)& $1.238$& $0.511$& $0.412$& $2.426$& $0.894$\\
RMSprop, $1e^{-7}$, KLD &$1.206$& $0.532$& $0.418$& $2.602$& $0.900$\\
RMSprop, $1e^{-5}$, KLD&$\mathbf{1.052}$& $\mathbf{0.612}$& $\mathbf{0.463}$& $\mathbf{3.237}$& $\mathbf{0.912}$\\
\end{tabular}
}
	\caption{Performance comparisons on LEDOV test set for SalEMA trained with the original hyperaprameter settings and the ones used in this paper (RMPprop optimizer with $1e^{-5}$ learning rate) after training on LEDOV training set. Best metrics are in bold.}
	\label{tab:salema-hyperparam}
\end{table}

\paragraph{Details of training EML-Net.} 
We train EML-Net~\cite{Jia20} for image-saliency on our noisy version of SALICON train set~\cite{Jia17} (generated by randomly selecting a subset $5$ or $15$ fixations per image, see Sec.~6 in main paper). 
To do so, we select the ResNet50 backbone~\cite{He16}. 
Consistent with recommendations from authors, we train two versions of the encoder: first, we finetune starting from ImageNet-pretrained weights~\cite{Rus15}, and second, we finetune from Places365-pretrained weights~\cite{Zho17}. 
The two saliency models obtained from the encoder-training stage are input to the decoder-training pipeline to give the final image-saliency predictor for EML-Net approach. 
We adopt the EML discreapncy (which is a combination of KLD, CC and NSS losses described by authors) for training both traditionally (Eq.~{2} in main paper) and using NAT. 
After searching through learning rates and optimizers, we find the author-specified choices to be most optimal: SGD with momentum with a learning rate of $0.01$ at the beginning and multiplied by $0.1$ after every epoch. 
We train both encoder and decoder for $10$ epochs. 
After training, the best model for each experiment in Table~6 of the main paper is selected based on validation-set performance (using \textit{all} available fixations on images in validation set), and submitted to SALICON benchmark for evaluation on the test set~\cite{Jia17}. 
Note that even though the training is performed on few-fixation images to simulate a noisy version of the SALICON dataset, the evaluation on test set and validation set contains \textit{all} of the available fixations.

\section{Alternative methods to estimate \boldmath{$\tilde{x}$} (mentioned in Sec.~6)}
In Sec.~{6} of main paper, we discuss an alternative strategy using Gaussian kernel density estimation (KDE) with uniform regularizer to estimate $\tilde{x}$ for training, instead of the common practice of blurring human gaze fixation locations using a Gaussian blur kernel of size approximately $1^{\circ}$ viewing angle.
We provide further details here.
We estimate the optimal KDE bandwidth for \textit{each} video frame, mixed with a uniform regularizer whose coefficient is also a parameter to be estimated.
We do a per-frame estimation of optimal KDE bandwidth and mixing coefficient, to account for the general case where each frame can have a different variety of points of interest to attract gaze which cannot be explained with the optimal KDE bandwidth of another frame.
The alternative to this is to estimate an optimal KDE bandwidth independent of the video frames, which amounts to the case of obtaining a universal Gaussian-blur kernel of a different size. 
In this case, the treatment of the underlying gaze data for obtaining the measured saliency maps, $\tilde{\tilde{x_i}}$, remains the same, in principle, as our experiments with $\sim1^{\circ}$ viewing-angle Gaussian-blur kernel (which amounts to $36$ pixels and $1920\times1080$ resolution for ForGED).
To demonstrate this for completeness, in Table~\ref{tab:tased-forged-kld-smallblur}, we show some of the results for TASED trained with ForGED and KLD as discrepancy.
For this experiment, the training gaze maps are estimated using a Gaussian-blur kernel of size $27$ pixels (at resolution $1920\times1080$), which amounts to $\sim 0.75^{\circ}$ viewing angle. 
We note in Table~\ref{tab:tased-forged-kld-smallblur} that NAT outperforms traditional training, consistent with our experiments with $\sim1^{\circ}$ viewing-angle Gaussian-blur kernel reported in the main paper.
\begin{table}[h!]
	\centering
	\resizebox{0.6\columnwidth}{!}{
		\renewcommand{\arraystretch}{1.05}
\scriptsize
\centering
\begin{tabular}[b]{c|c|c|c|c|c|c|c}
\textbf{train videos $V$} & \textbf{train obs. $N$} & \textbf{loss} & \textbf{KLD}$\downarrow$&\textbf{CC}$\uparrow$ &\textbf{SIM}$\uparrow$ & \textbf{NSS}$\uparrow$ &\textbf{AUC-J}$\uparrow$ \\	
\hline
\multirow{6}{*}{$30$} & \multirow{2}{*}{$2$} & TT &$1.586$& $0.471$& $0.329$& $2.832$& $0.878$\\
& & NAT &$\mathbf{1.387}$& $\mathbf{0.546}$& $\mathbf{0.378}$& $\mathbf{3.153}$& $\mathbf{0.879}$\\
\cline{2-8}
& \multirow{2}{*}{$5$} & TT &$1.358$& $0.563$& $0.345$& $3.184$& $0.903$\\
& & NAT &$\mathbf{1.239}$& $\mathbf{0.565}$& $\mathbf{0.406}$& $\mathbf{3.272}$& $\mathbf{0.905}$\\
\cline{2-8}
& \multirow{2}{*}{$15$} & TT &$1.056$& $\mathbf{0.622}$& $\mathbf{0.483}$& $3.682$& $0.902$\\
& & NAT &$\mathbf{1.035}$& $0.616$& $0.476$& $\mathbf{3.757}$& $\mathbf{0.917}$\\
\hline
\multirow{2}{*}{$100$} & \multirow{2}{*}{$5$} & TT &$1.085$& $0.634$& $0.464$& $\mathbf{3.770}$& $0.903$\\
& & NAT &$\mathbf{1.018}$& $\mathbf{0.636}$& $\mathbf{0.474}$& $3.633$& $\mathbf{0.926}$\\
\hline
\multirow{2}{*}{$379$} & \multirow{2}{*}{$5$} & TT &$0.959$& $0.651$& $0.480$& $3.652$& $\mathbf{0.931}$\\
& & NAT &$\mathbf{0.888}$& $\mathbf{0.670}$& $\mathbf{0.517}$& $\mathbf{4.091}$& $0.924$\\
\end{tabular}
}
	\caption{Performance comparisons on ForGED test set for TASED trained with KLD as discrepancy. Instead of computing gaze maps for train set with Gaussian blur kernel of size approximately $1^{\circ}$ viewing angle (which amounts of $36$ pixels at $1920\times1080$ resolution), we use a Gaussian blur kernel of size approximately $0.75^{\circ}$ viewing angle ($27$ pixels). As we can see, the conclusion regarding the superior performance of NAT compared to traditional training applies independent of blur kernel size.}
	\label{tab:tased-forged-kld-smallblur}
\end{table}

To estimate the optimal bandwidth using KDE, we optimize a gold-standard model for saliency prediction, which predicts the probability of fixation for one observer, given the gaze data from the remaining observers for the video frame (leave-one-out cross-validation)~\cite{Kum15,Tan20}.
We observe that, when gaze fixation locations are sparsely distributed across a frame, the optimal bandwidth for KDE is high, which would result is high-spread, almost-uniform saliency maps.
Independent of the estimation strategy for $\tilde{x}$, we posit that there is an underlying uncertainty / noise in the measured saliency map --  which is accounted for during training using NAT, to obtain improved performance over traditional training.

\begin{figure*}[h!]
	\centering
	\begingroup
	\setlength{\tabcolsep}{0pt} 
	\renewcommand{\arraystretch}{2}
	\begin{tabular}{ccc}
		\includegraphics[width=0.3\textwidth,trim={0cm 0cm 0cm 0.5cm}, clip]{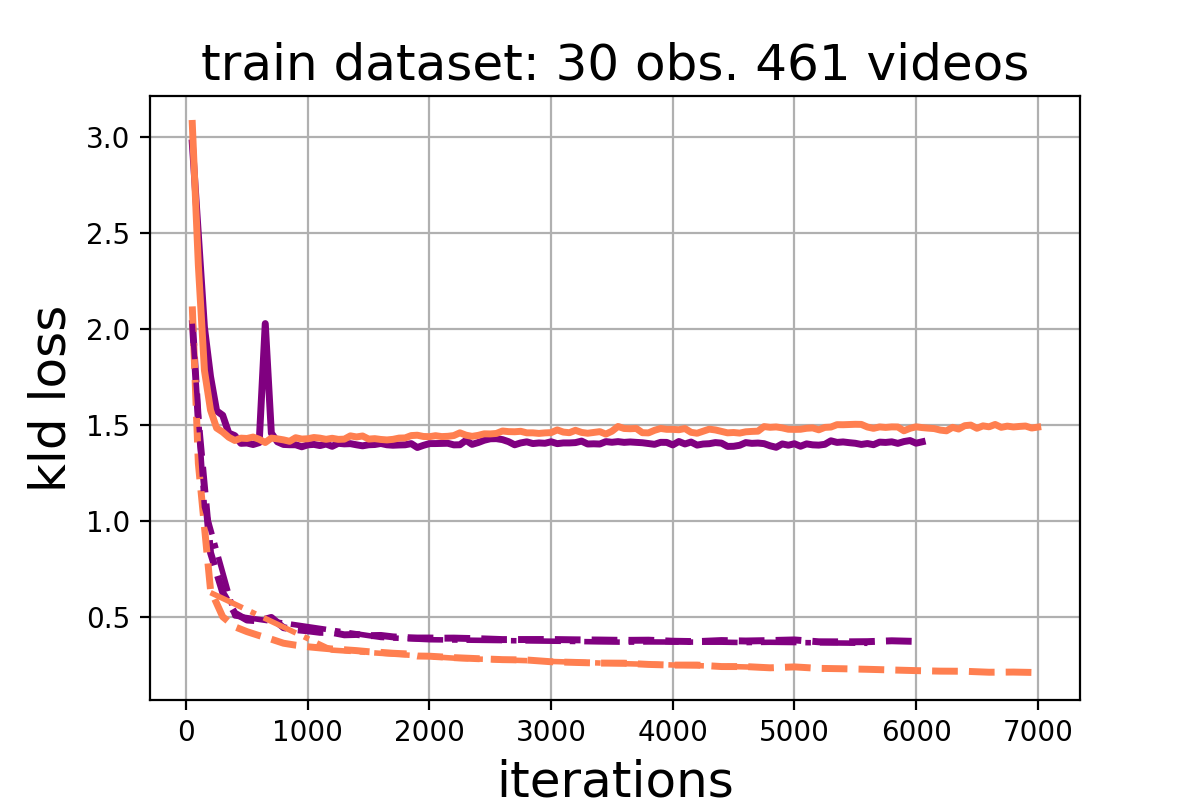} &
		\includegraphics[width=0.3\textwidth,trim={0cm 0cm 0cm 0.5cm}, clip]{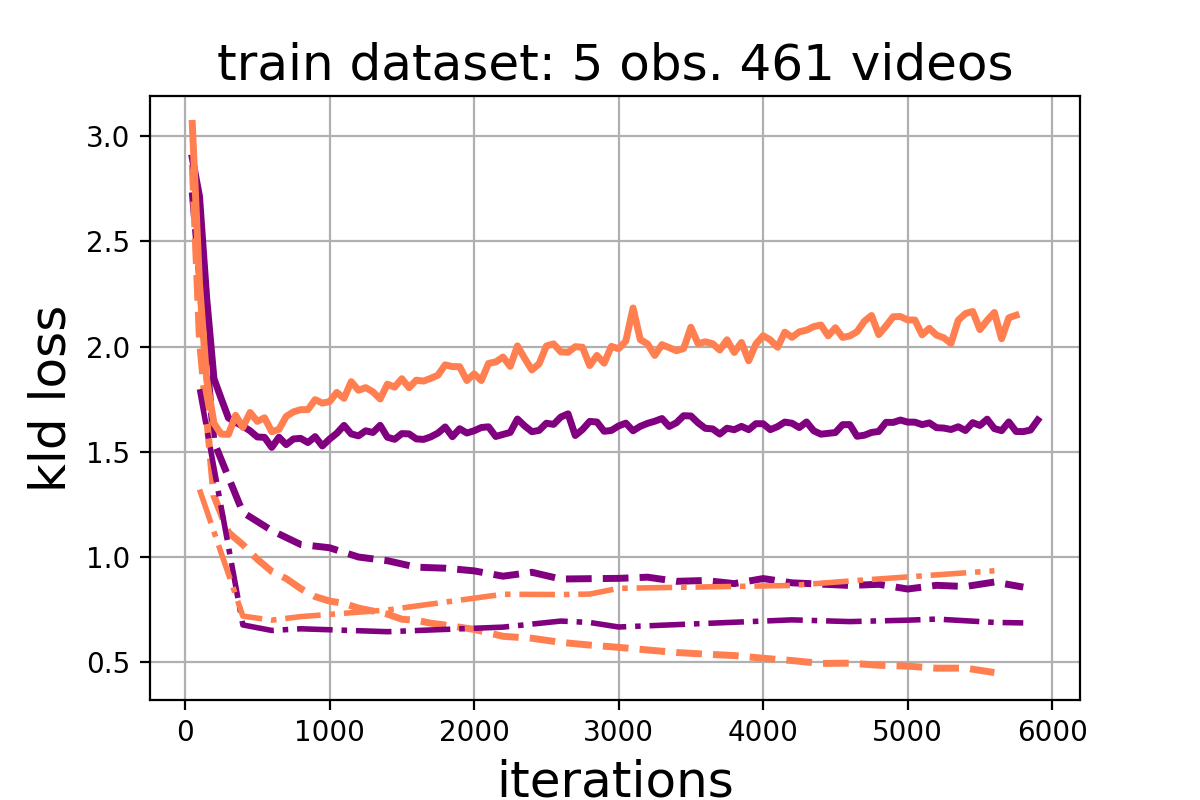}  &
		\includegraphics[width=0.3\textwidth,trim={0cm 0cm 0cm 0.5cm}, clip]{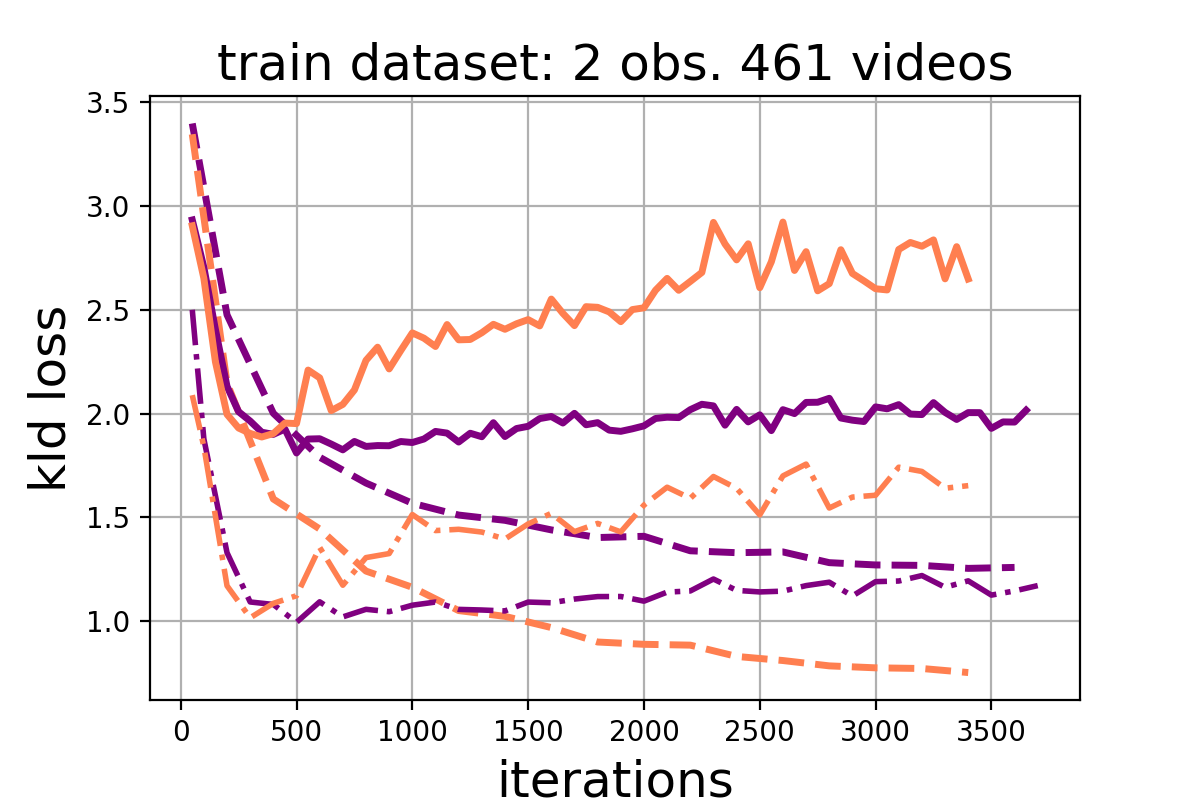}    \\
		\includegraphics[width=0.3\textwidth,trim={0cm 0cm 0cm 0.5cm}, clip]{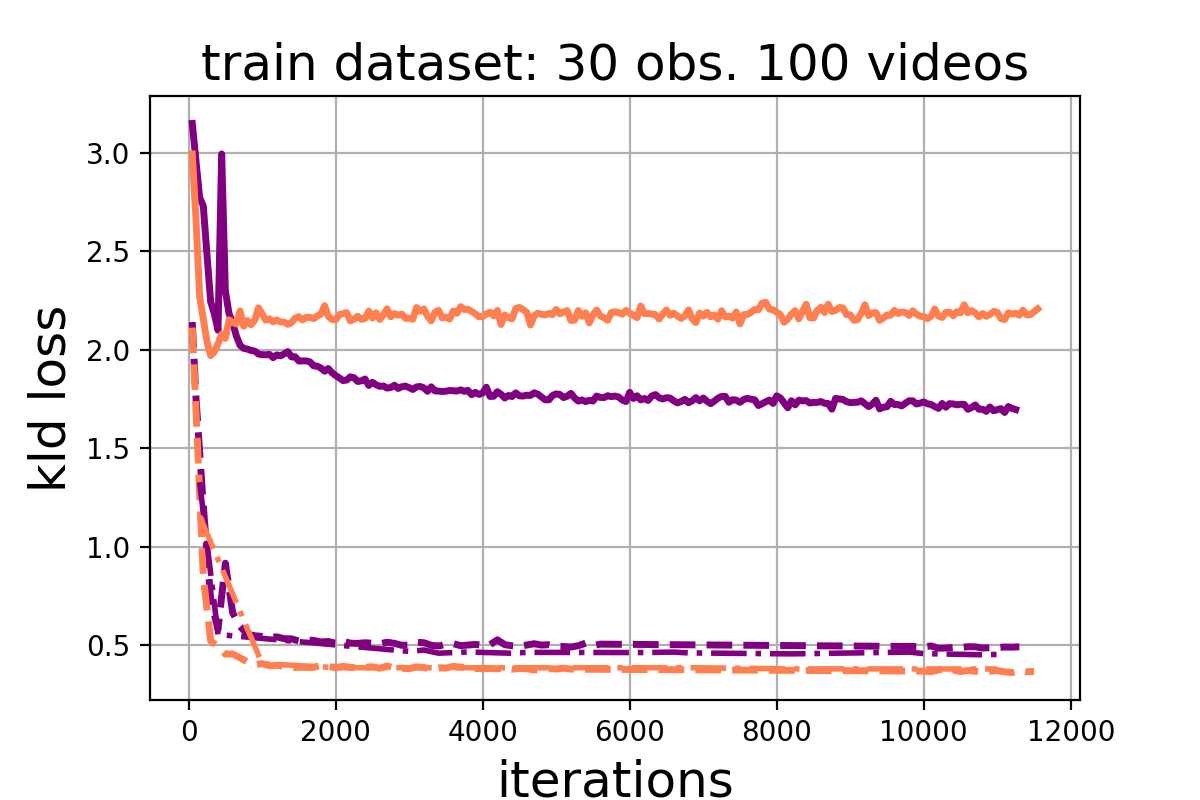} &
		\includegraphics[width=0.3\textwidth,trim={0cm 0cm 0cm 0.5cm}, clip]{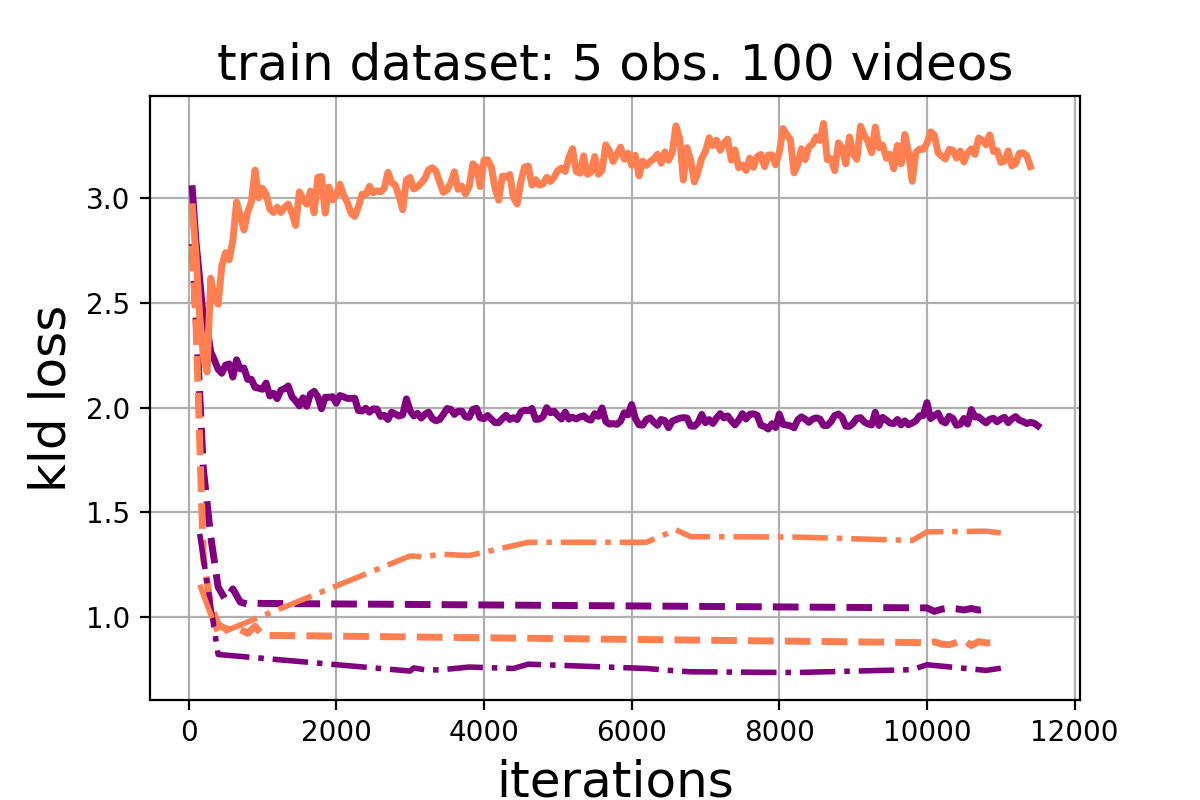}  &
		\includegraphics[width=0.3\textwidth,trim={0cm 0cm 0cm 0.5cm}, clip]{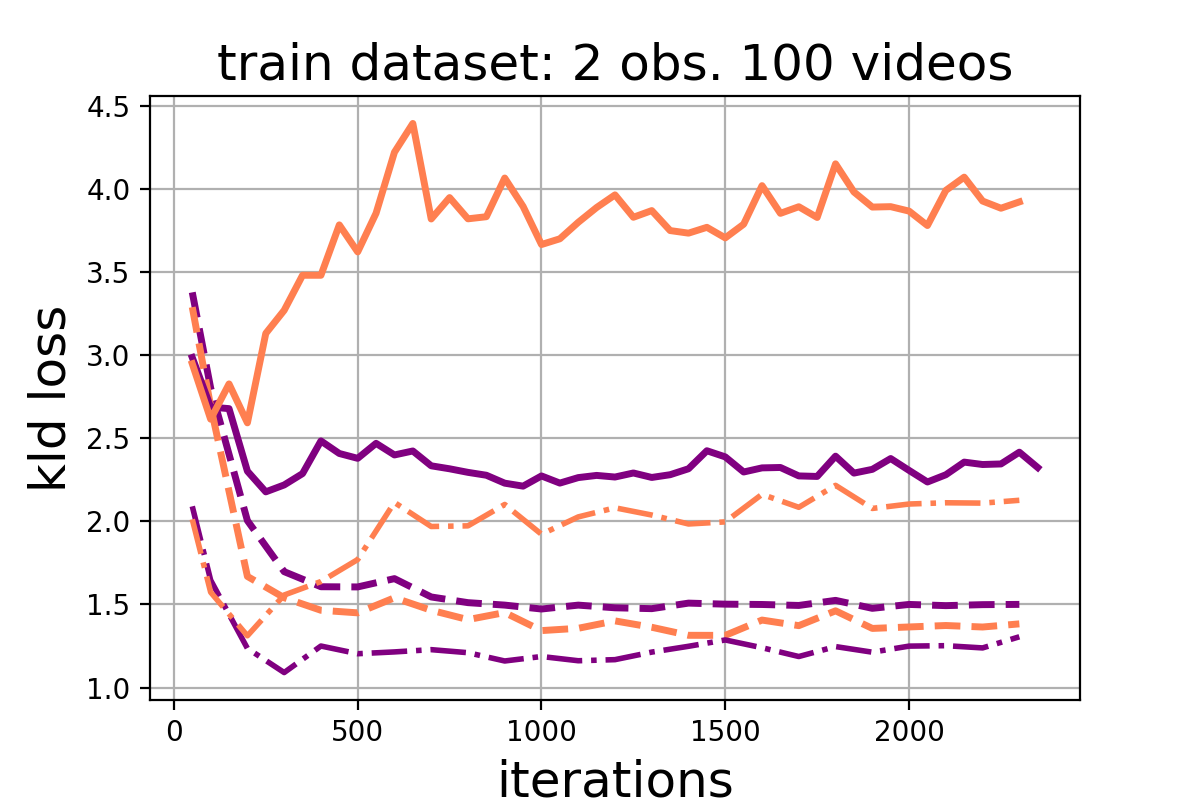}    \\
		\includegraphics[width=0.3\textwidth,trim={0cm 0cm 0cm 0.5cm}, clip]{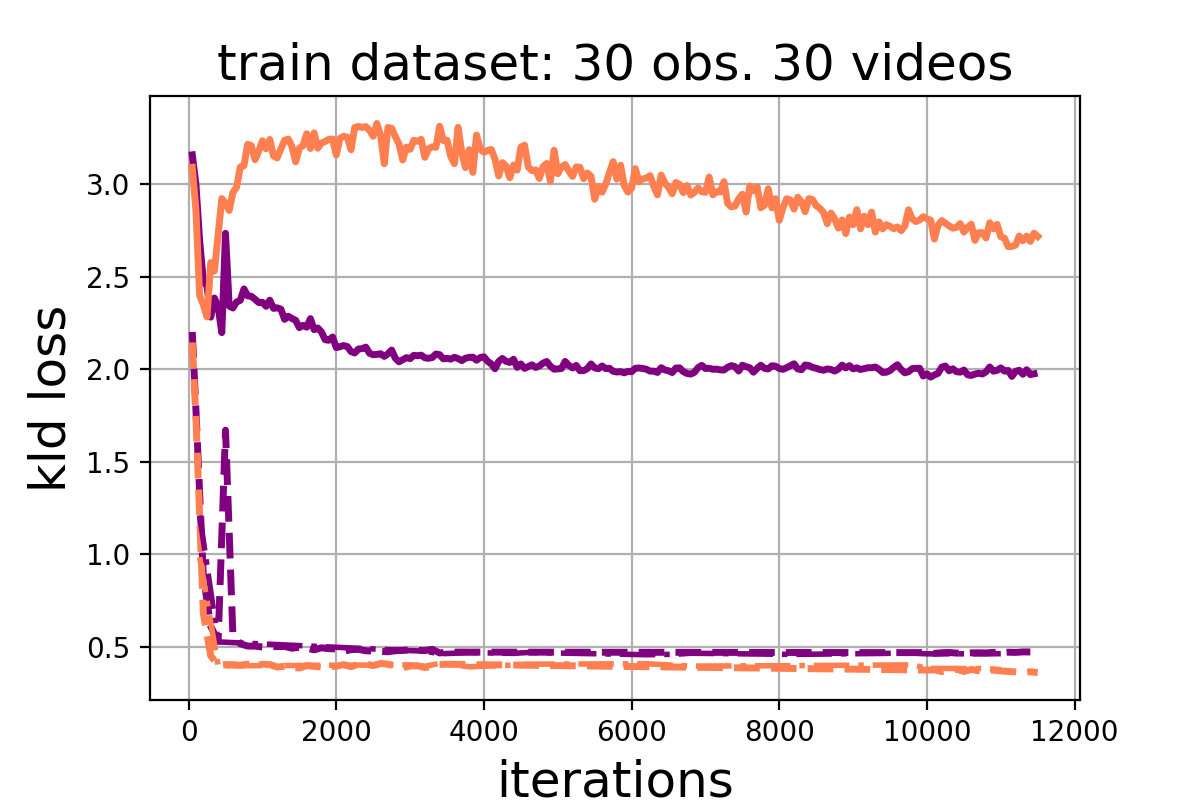}  &
		\includegraphics[width=0.3\textwidth,trim={0cm 0cm 0cm 0.5cm}, clip]{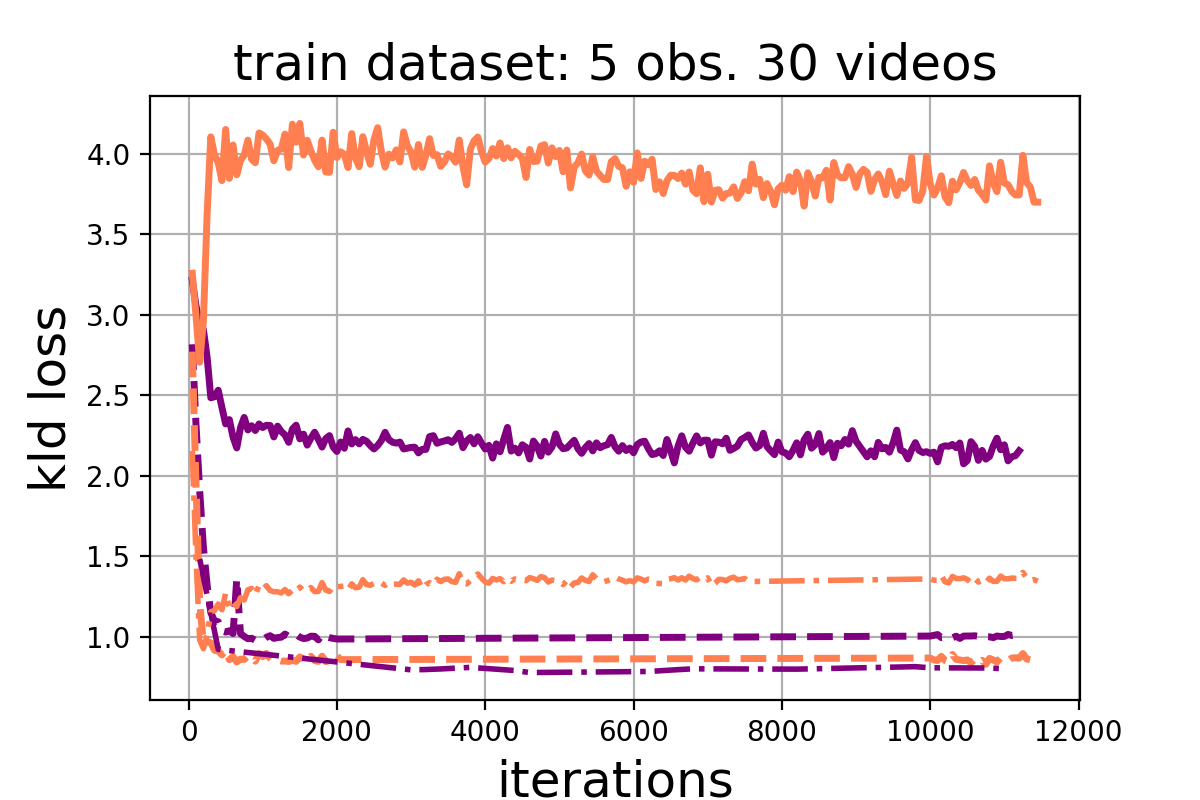}   &
		\includegraphics[width=0.3\textwidth,trim={0cm 0cm 0cm 0.5cm}, clip]{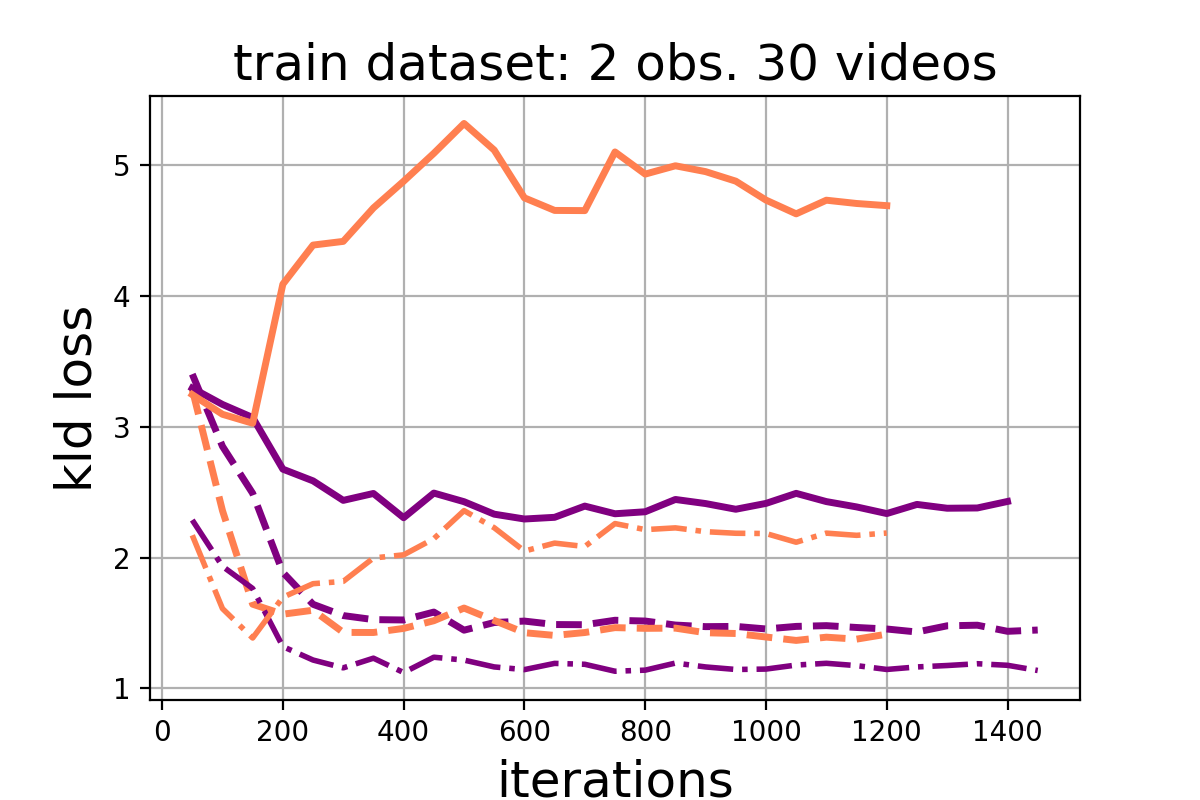}     \\
		\multicolumn{3}{c}{\includegraphics[width=0.8\textwidth,trim={0cm 0cm 0cm 0.5cm}, clip]{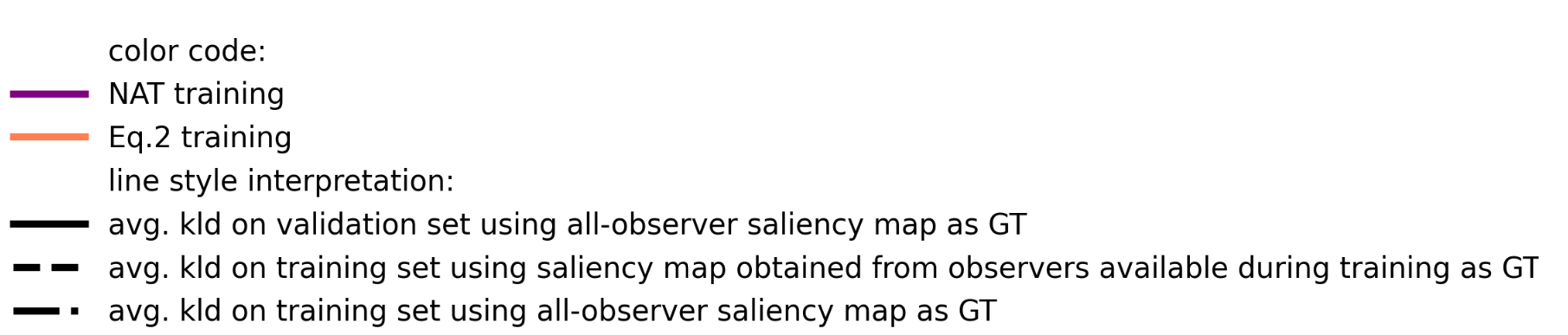}} \\
	\end{tabular}
	\endgroup
	\caption{Training-set and validation-set KLD as a function of training iterations for TASED trained on LEDOV (``GT'' in the legend indicates ``ground-truth''). In contrast to the traditional training (Eq.2 in main paper), NAT does not overfit. }
	\label{fig:training-behavior}
\end{figure*}

\section{Overfitting behavior with NAT (mentioned in Sec.~3)}
Figure~\ref{fig:training-behavior} shows the training and validation set performance (in terms of KLD) as a function of the training iteration when training TASED on LEDOV dataset with KLD discrepancy, for different number of observers and videos in the training set.
For both the traditional approach (dashed orange line) and NAT (dashed purple line), the training-set curves decrease regularly, as expected in a smooth optimization process. 
However, the validation-set curves for traditional training (continuous orange line) quickly reach a minimum and then start diverging towards a higher asymptotic value, which is a clear sign of overfitting.
On the other hand, the validation curves for NAT (continuous purple line) are always lower (suggesting better performance) and tend to stabilize around asymptotic values without growing anymore --- a clear sign, in this case, that overfitting is avoided.
Note that for the training-set curves (dashed lines), the human saliency map used for KLD computation is derived using the limited number of observers available during the specific training experiment.
As an additional check for the overfitting behavior of traditional training, we plot the performance of training set when compared against human saliency maps obtained from \textit{all} the observers available in the training videos ($32$). 
These are indicated with dash-dotted lines.
For few-observer experiments, the performance of traditional training on all-observer evaluations gets worse with increasing iterations.
On the contrary, the performance on validation set, training set, and all-observer training set do not generally show signs of overfitting for NAT.
Only in few cases, NAT plots are unstable at the beginning of the training (see the peaks in the validation curves in the left most panels for $30$ observers trainings), but then the curves stabilize to an asymptotic value.
The only exception to this is represented by the upper right panel in the figure ($2$-observer training with $461$ videos), where we believe that the slight increase in the validation-set performance value is due to the approximation introduced in NAT to make it computable in practice.
We observed a similar behavior when training on other datasets.

\end{document}